\documentclass[10pt,twocolumn,letterpaper]{article}

\usepackage{cvpr}              

\usepackage{graphicx}
\usepackage{amsmath}
\usepackage{amssymb}
\usepackage{booktabs}
\usepackage{times}
\usepackage{epsfig}
\usepackage{bm}
\usepackage{comment}
\usepackage{color}
\usepackage{mathtools}
\usepackage{xfrac}
\usepackage{multirow}
\usepackage{nicefrac}
\usepackage{caption}
\usepackage{stmaryrd}
\usepackage[pagebackref,breaklinks,colorlinks]{hyperref}

\usepackage[capitalize]{cleveref}
\crefname{section}{Sec.}{Secs.}
\Crefname{section}{Section}{Sections}
\Crefname{table}{Table}{Tables}
\crefname{table}{Tab.}{Tabs.}

\def\cvprPaperID{1343} 
\def\confName{CVPR}
\def\confYear{2022}

\begin{document}

\title{
Multi-Frame Self-Supervised Depth with Transformers
}

\author{
Vitor Guizilini \quad Rareș Ambruș \quad Dian Chen \quad Sergey Zakharov \quad Adrien Gaidon
\vspace{1mm}
\\ 
Toyota Research Institute (TRI), Los Altos, CA
\\
{\tt\small \{first.lastname\}@tri.global}
}

\twocolumn[{%
\renewcommand\twocolumn[1][]{#1}%
\vspace{-15mm}
\maketitle
\begin{center}
    \vspace{-3mm}
    \centering
    \captionsetup{type=figure}
    \includegraphics[width=.85\textwidth,height=4.4cm]{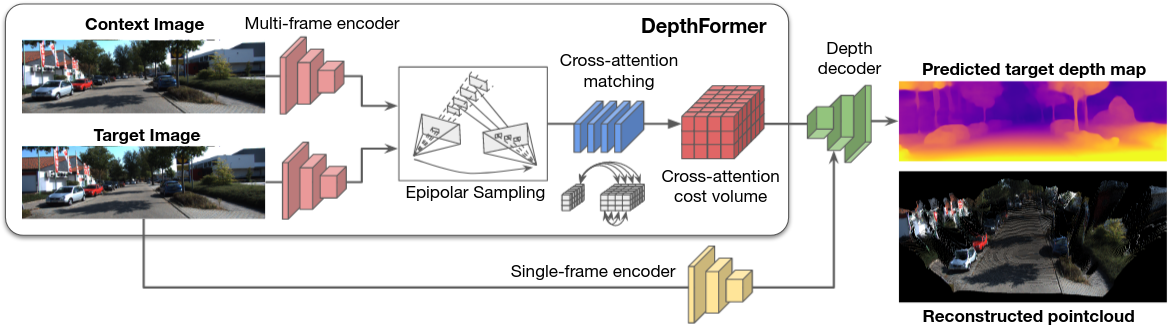}
    \captionof{figure}{\textbf{Our DepthFormer architecture} achieves state-of-the-art multi-frame self-supervised monocular depth estimation by \textbf{improving feature matching} across images during cost volume generation.}
    \label{fig:teaser}
\end{center}%
}]


\begin{abstract}

Multi-frame depth estimation improves over single-frame approaches by also leveraging geometric relationships between images via feature matching, in addition to learning appearance-based features.
In this paper we revisit feature matching for self-supervised monocular depth estimation, and propose a novel transformer architecture for cost volume generation. 
We use depth-discretized epipolar sampling to select matching candidates, and refine predictions through a series of self- and cross-attention layers. These layers sharpen the matching probability between pixel features, improving over standard similarity metrics prone to ambiguities and local minima.   
The refined cost volume is decoded into depth estimates, and the whole pipeline is trained end-to-end from videos using only a photometric objective.
Experiments on the KITTI and DDAD datasets show that our DepthFormer architecture establishes a new state of the art in self-supervised monocular depth estimation, and is even competitive with highly specialized supervised single-frame architectures.
We also show that our learned cross-attention network yields representations transferable across datasets, increasing  the effectiveness of pre-training strategies. Project page: \url{https://sites.google.com/view/tri-depthformer}.

\end{abstract}

\vspace{-3mm}
\section{Introduction}


Feature matching is a fundamental component of Structure-from-Motion (SfM). By establishing correspondences between points across frames, a wide range of tasks can be performed, including depth estimation~\cite{godard2017unsupervised,packnet,monodepth2,casser2018depth}, ego-motion estimation~\cite{tang2020kp3d,kendall2015posenet,kendall2017geometric}, keypoint extraction~\cite{tang2020kp3d,tang2020neural}, calibration~\cite{vasiljevic2020neural,gordon2019depth}, optical flow~\cite{competi_colab,yin2018geonet,jonschkowski2020matters}, and scene flow~\cite{selfsceneflow,multi-scene-flow}. Within these tasks, self-supervision enables learning without explicit ground-truth~\cite{zhou2017unsupervised,godard2017unsupervised}, by using view synthesis losses obtained via the warping of information from one image onto another, obtained from multiple cameras or a single moving camera. While more challenging from a training perspective~\cite{monodepth2,manydepth,packnet}, self-supervised methods can leverage arbitrarily large amounts of unlabeled data, which has been shown to achieve performance comparable to supervised methods~\cite{packnet,manydepth}, while enabling new applications such as test-time refinement~\cite{shu2020featdepth,manydepth,gordon2019depth} and unsupervised domain adaptation~\cite{guda}.

Single-frame self-supervised methods use multi-view information only at training time, as part of the loss calculation~\cite{zhou2017unsupervised,godard2017unsupervised,monodepth2,packnet,shu2020featdepth}. In contrast, multi-frame methods use multi-view information at \emph{inference time}, traditionally by building cost volumes~\cite{manydepth,monorec,deep_mv_depth,pwcnet} or correlation layers~\cite{deepv2d,raft,selfsceneflow}. These methods learn geometric features in addition to appearance-based ones, which leads to better performance relative to single-frame methods~\cite{manydepth,monorec,deepv2d}. However, multi-frame calculation relies heavily on feature matching to establish correspondences between frames, using only image information. Because of that, correspondences will be noisy and often inaccurate~\cite{monodepth2,packnet,monorec} due to ambiguities and local minima caused by lack of texture, repetitions, luminosity changes, dynamic objects, and so forth. 

In this paper we introduce a novel architecture designed to improve self-supervised feature matching (Figure \ref{fig:teaser}), focusing on the task of monocular depth estimation. We build a cost volume between target and context image features using differentiable depth-discretized epipolar sampling, and propose a novel attention-based mechanism to refine per-pixel matching probabilities. We show that the refined probabilities are sharper and more representative of the underlying 3D structure than traditional similarity metrics~\cite{wang2004image}. The resulting multi-frame cost volume is converted into depth estimates directly, via high-response window filtering, and in combination with single-frame features from a separate network, to account for failure cases in cost volume generation. Through extensive experiments, we show that our feature matching refinement module leads to a new state of the art in self-supervised depth estimation, and that it can be directly transferred between datasets with minimal degradation thanks to its strong geometric grounding. 
Our main contributions are:

\begin{itemize}
\item We introduce a novel architecture, the \emph{DepthFormer}, that \textbf{improves multi-view feature matching via cross- and self-attention combined with depth-discretized epipolar sampling}. 

\item Our architecture leads to \textbf{state-of-the-art depth estimation results}. It outperforms other self-supervised multi-frame methods by a large margin, and even \textbf{surpasses supervised single-frame architectures}. 

\item Our learned attention-based matching function is \textbf{transferable across datasets}, which can significantly improve convergence speed while decreasing memory.

\end{itemize}

\section{Related Work}

\subsection{Self-Supervised Depth Estimation}

%
The work of Godard \etal~\cite{godard2017unsupervised} introduced self-supervision to the task of depth estimation by framing it as a view synthesis problem, and minimizing an image reconstruction objective~\cite{wang2004image}. Originally proposed for stereo pairs, the same self-supervised framework was later extended to the monocular setting~\cite{zhou2017unsupervised}, with the addition of a pose network to estimate camera motion between frames. Although more challenging and restrictive, due to limitations such as scale ambiguity~\cite{packnet} and inability to model dynamic objects~\cite{monodepth2}, monocular self-supervision enables learning from raw videos, which makes it much more scalable to large amounts of data from different sources. Further improvements in the past few years, in terms of view synthesis~\cite{monodepth2,shu2020featdepth}, camera geometry modeling~\cite{gordon2019depth,vasiljevic2020neural}, network architectures~\cite{packnet}, domain adaptation~\cite{guda,virtualworld,gasda,sharingan}, scale disambiguation~\cite{packnet}, and other sources of supervision~\cite{gordon2019depth,packnet-semguided}, have led to performance comparable to or even surpassing supervised approaches~\cite{manydepth,packnet, klingner2020self}.  

\subsection{Multi-Frame Depth Estimation}

Depth estimation from a single image is inherently an ill-posed problem, since an infinite number of 3D scenes could result in the same 2D projection~\cite{hartley2003multiple}. Single-frame networks learn appearance-based cues that are suitable for depth estimation (e.g., vanishing point distance, location relative to the ground plane), however these cues are usually based on strong assumptions and will fail with the right adversarial attacks~\cite{how_do_networks}. Multi-frame depth estimation methods circumvent this limitation by using multiple images at test time, which enables the learning of additional geometric cues from feature matching across frames. Although other frameworks for multi-view depth estimation are available, e.g., test-time refinement~\cite{casser2018depth,Luo-VideoDepth-2020,shu2020featdepth} and recurrent neural networks~\cite{forget_past,depthnet,zhang2019temporal}, here we focus on methods that explicitly reason about geometry during inference.


\begin{figure*}[t!]
\vspace{-3mm}
    \centering
    \hspace{-5mm}
    \subfloat[\textbf{Cross-attention cost volume generation} from two images.]{
        \includegraphics[width=0.55\textwidth]{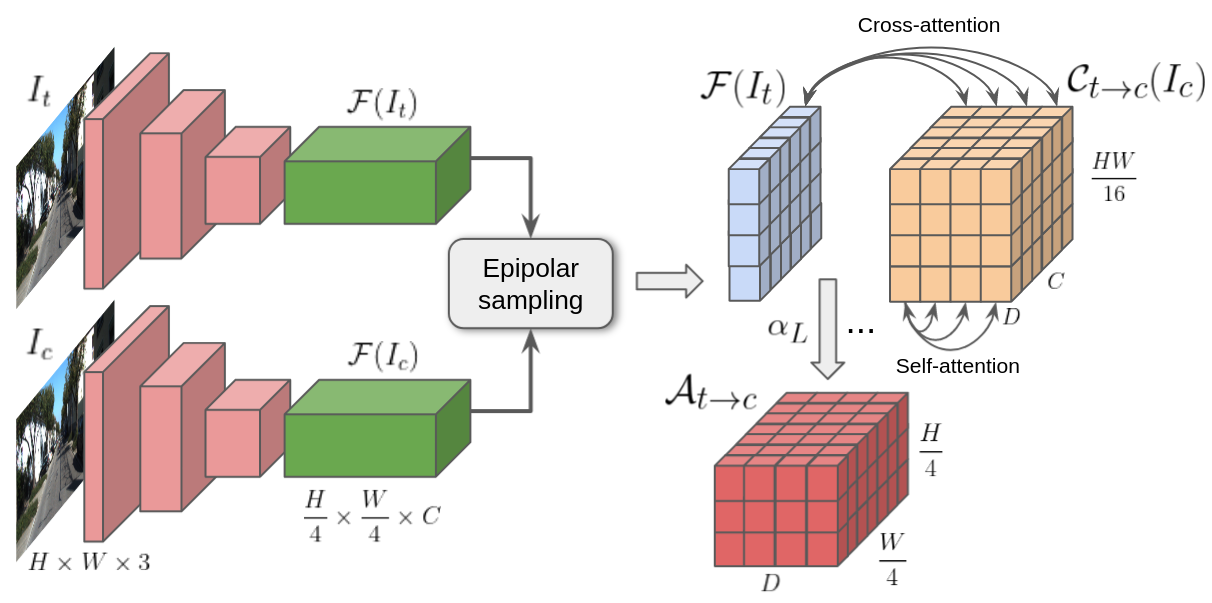}
    \label{fig:cross-attention}
    }
    \hspace{-3mm}
    \subfloat[\textbf{Depth-discretized epipolar sampling} for cost volume generation.]{
        \includegraphics[width=0.45\textwidth,height=5.2cm]{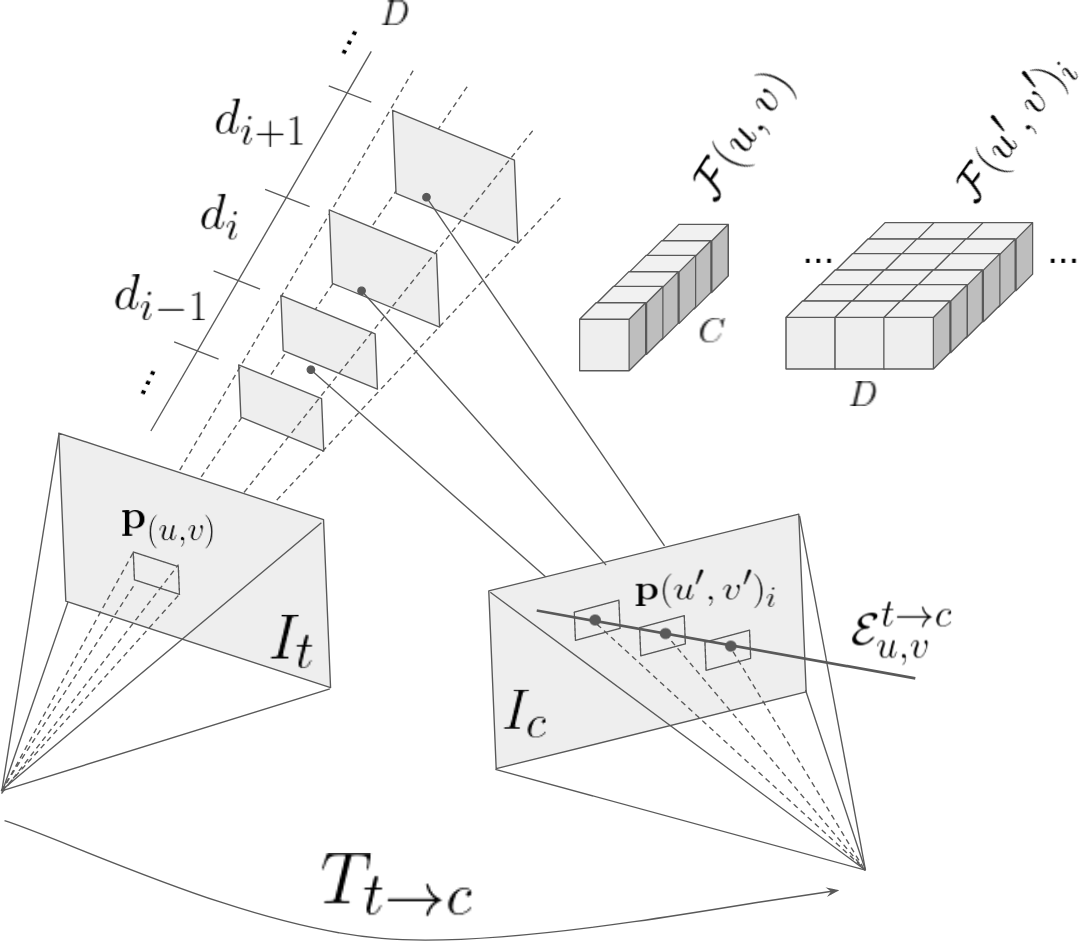}
    \label{fig:epipolar-sampling}
    }
    \caption{
        \textbf{Diagram for our proposed cross-attention cost volume generation.} Two images (target $I_t$ and context $I_c$) are processed by a feature extraction network $\mathcal{F}$ and, for each target feature $\mathcal{F}(u,v)$, $D$ matching candidates $\mathcal{F}(u'_i,v'_i)$ are sampled from the depth-discretized epipolar line $\mathcal{E}_{u,v}^{t \rightarrow c}$. A series of $L$ self- and cross-attention layers is then used to refine this initial matching distribution. The output is a \emph{cross-attention cost volume} $\mathcal{A}_{t \rightarrow c}$, containing the matching probability of each target feature relative to its epipolar candidates, given by the corresponding estimated cross-attention value $\alpha_L(u_i,v_i)$. 
    }
    \label{fig:diagram}
\vspace{-4mm}
\end{figure*}

Stereo methods simplify this feature matching process by considering fronto-parallel rectified image pairs with known baseline~\cite{large_dataset,Liang2018Learning,deep_stero_regression,aleotti2020reversing,Wu_2019_ICCV}.
Multi-view stereo (MVS) is a generalization of the rectified setting, that operates on images with arbitrary overlaps~\cite{huang2018deepmvs, im2019dpsnet, kar2017learning, luo2020attention, xue2019mvscrf}.
Most MVS approaches, however, are supervised and assume known camera poses (either as ground-truth or obtained through COLMAP~\cite{schonberger2016structure}). 
Similarly, recently implicit representation methods have also enabled multi-view self-supervised learning~\cite{nerf++,nerf--,scnerf,pixelnerf}, including extensions to depth estimation~\cite{dsnerf,nerfingmvs}. However, such methods focus on over-fitting to simple scenes with static objects and surrounding high-overlapping views, which limits their generalization  to large-scale  datasets~\cite{geiger2013vision,caesar2020nuscenes,cordts2015cityscapes,packnet}. 

Importantly, the use of known camera poses, stereo pairs, supervision and/or static scenes, side-steps some of the main limitations of monocular self-supervised learning. A few methods~\cite{manydepth,monorec} have recently enabled depth and ego-motion estimation in this setting by combining a multi-frame cost volume with single-frame features. However, they still rely on hand-crafted similarity metrics: ManyDepth~\cite{manydepth} uses sum of absolute differences (SAD); and MonoREC~\cite{monorec} uses structural similarity (SSIM). As we shown in our experiments, these metrics are prone to ambiguity and local minima, leading to sub-optimal correspondences. Our attention-based mechanism is designed to improve multi-frame matching for cost volume generation. 

\subsection{Attention for Depth Estimation}

After transforming the field of natural language processing~\cite{attention_all}, attention-based architectures are becoming increasingly popular in computer vision~\cite{dpt,sttr,dosovitskiy2020vit,liu2021swin}. 
In~\cite{Huynh2020GuidingMD}, a depth-attention volume is used to guide the learning of indoor planar surfaces, while~\cite{Sadek2020SelfSupervisedAL} uses attention for depth decoding. 
Similarly,~\cite{Lee2021PatchWiseAN} uses patch-wise attention over convolutional features, and~\cite{dpt} eliminates convolutional encoding by proposing a fully attention-based 
backbone. 
In ~\cite{johnston2020selfsupervised} a self-attention mechanism is used to process a convolutional feature embedding, and depth is decoded via integration over a discretized disparity cost volume.
More related to our work,~\cite{sttr} proposes self- and cross-attention over rectified images, followed by cost volume decoding into depth estimates. 
Their approach, however, is supervised and operates on the simpler stereo setting. A self-supervised monocular attention-based method is proposed in ~\cite{ruhkamp2021attention}, using a spatio-temporal module to leverage both geometric and appearance information. 
However, by focusing on 3D points for attention, they forego the epipolar constraints we use to determine matching candidates.

\section{Self-Supervised Depth with Transformers}


\subsection{Monocular Depth Estimation}

The standard self-supervised monocular depth and ego-motion architecture consists of (i) a depth network $f_D(I_t;\theta_D)$, that produces depth maps $\hat{D}_{t}$ for a target image $I_t$; and (ii) a pose network $f_T(I_t,I_c;\theta_T)$, that predicts
the relative transformation for pairs of target $I_t$ and context $I_c$ images.  This pose prediction is a rigid transformation $\hat{\mathbf{T}}_{t \shortrightarrow c} = \begin{psmallmatrix}\mathbf{\hat{R}_{t\shortrightarrow c}} & \mathbf{\hat{t}_{t\shortrightarrow c}}\\ \mathbf{0} & \mathbf{1}\end{psmallmatrix} \in \text{SE(3)}$.
We train these two networks jointly by minimizing a photometric reprojection error~\cite{godard2017unsupervised, zhou2017unsupervised} between the original target image $I_t$ and the synthesized target image $\hat{I}_t$, obtained by projecting pixels from $I_c$ onto $I_t$ using predicted depth and pose.
The synthesized image is obtained via grid sampling with bilinear interpolation~\cite{zhou2017unsupervised}, and is thus differentiable, which enables gradient back-propagation for end-to-end training. 




\subsection{Cross-Attention Cost Volumes}
\label{sec:cross_attention}

\subsubsection{Monocular Epipolar Sampling}

A diagram of our proposed cross-attention cost volume generation procedure is shown in Figure \ref{fig:cross-attention}. %
Two $H \times W \times 3$ input images, target $I_t$ and context $I_c$, are encoded to produce $C$-dimensional features $\mathcal{F}_t$ and $\mathcal{F}_c$ at 1/4 the original resolution. 
%
%
For each feature $\textit{\textbf{f}}_{t}^{\,uv} \in \mathcal{F}_t$, corresponding to pixel $\textbf{p}_t = \{ u,v \}$, matching candidates are sampled from $\mathcal{F}_c$ along its epipolar line $\mathcal{E}^{uv}_{t \rightarrow c}$, as shown in Figure \ref{fig:epipolar-sampling}.
We use spatial-increasing discretization (SID)~\cite{fu2018deep} to uniformly sample depth values in \emph{log} space. Assuming $D$ bins ranging from $d_{min}$ to $d_{max}$, each depth value $d_i$ is given by:
\begin{equation}
    \log(d_i) = {\log(d_{min}) + \frac{\log(d_{max} / d_{min}) * i}{D}}
\end{equation}

A $H/4 \times W/4 \times D \times C$ \emph{feature volume} $\mathcal{C}_{t \rightarrow c}$ is generated from these matching candidates. Each $(u,v,i)$ cell receives sampled features $\mathcal{F}_{t \rightarrow c}^{uv} = \mathcal{F}_c \langle u'_i,v'_i \rangle \, , \, \text{for} \, i \in \left[ 0, \cdots, D \right]$, where $\langle \rangle$ is the bilinear sampling operator and $(u',v')$ are projected pixel coordinates such that:
\begin{equation}
\small
    z'_i \left[\begin{matrix}u'_i \\ v'_i \\ 1 \\
    \end{matrix}\right] = 
    \mathbf{K} \mathbf{R}_{t \rightarrow c} \left( \mathbf{K}^{-1} 
    \left[\begin{matrix}u \\ v \\ 1 \\ 
    \end{matrix}\right] d_i 
    + \mathbf{t}_{t \rightarrow c}\right)
\end{equation}
where $\mathbf{R}_{t \rightarrow c}$ and $\mathbf{t}_{t \rightarrow c}$ are relative rotation and translation between frames, and $\mathbf{K} \in \mathbb{R}^{3 \times 3}$ are pinhole camera intrinsics. In practice, relative rotation and translation are predicted by the pose network, and $\mathbf{K}$ is assumed known and constant, although this assumption can be relaxed~\cite{gordon2019depth,vasiljevic2020neural}.

\begin{figure*}[t!]
    \centering
    \subfloat[SSIM]{
        \includegraphics[width=0.48\textwidth,height=3.0cm]{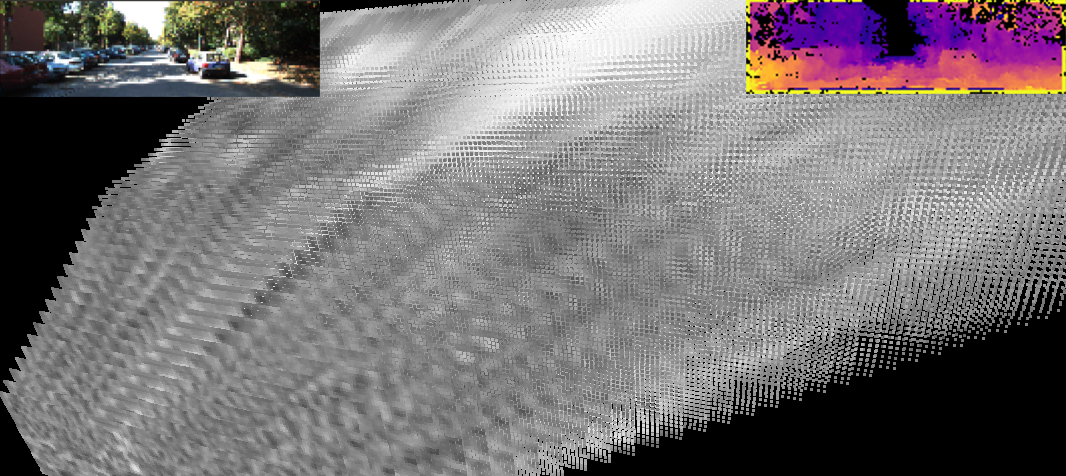}
        \label{fig:histogram_ssim}
    }
    \subfloat[Cross-Attention]{
        \includegraphics[width=0.48\textwidth,height=3.0cm]{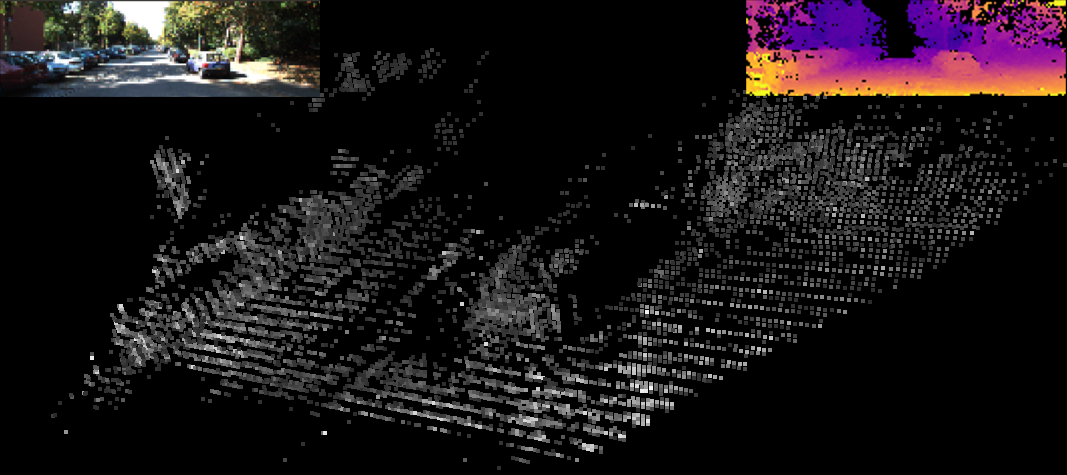}
        \label{fig:histogram_cross}
    }    
    \\
    \subfloat[Matching probability distribution along depth bins for different pixels relative to their depth-discretized epipolar candidates. The blue line shows SSIM values (Equation \ref{eq:photo_mono}), the red line shows cross-attention values $\alpha_L$ after refinement (Section \ref{sec:cross_attention}), and the green dot marks the corresponding ground-truth depth value (used only for comparison).]{
        \includegraphics[width=0.24\textwidth, trim={5mm 0mm 5mm 0mm},clip]{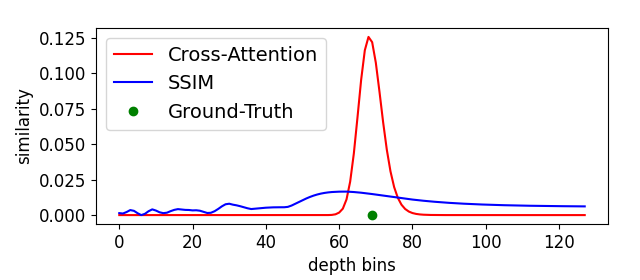}
        \includegraphics[width=0.24\textwidth, trim={5mm 0mm 5mm 0mm},clip]{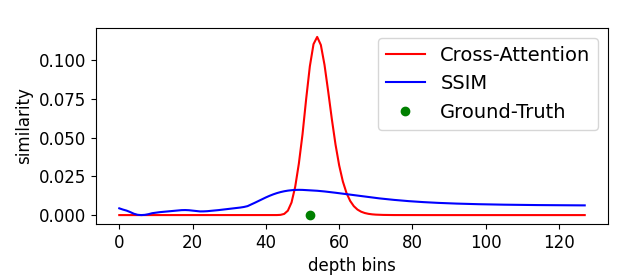}
        \includegraphics[width=0.24\textwidth, trim={6mm 0mm 5mm 0mm},clip]{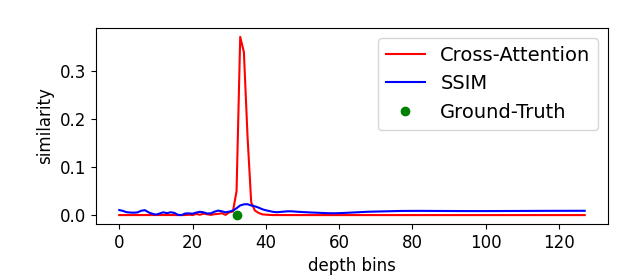}
        \includegraphics[width=0.24\textwidth, trim={6mm 0mm 5mm 0mm},clip]{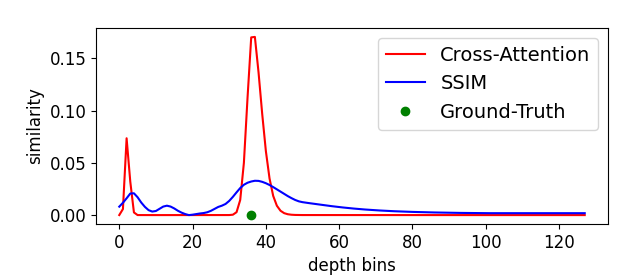}
        \label{fig:histogram_distrib}
    }    
    \caption{
    \textbf{Cost volume visualization.} In (a) and (b), each of the $H \times W \times D$ cells is colored based on its corresponding normalized SSIM or cross-attention value. Even though the decoded depth maps (top right) look similar, our proposed cross-attention cost volume produces sharper distributions, as further evidenced in (c) where we present various per-pixel matching distributions over depth bins. Our proposed attention-based similarity significantly increases the sharpness of these distributions, eliminating ambiguities and local minima.
    }
    \label{fig:histogram}
    \vspace{-5mm}
\end{figure*}

\subsubsection{Cross-Attention Matching}

An \emph{attention} module~\cite{attention_all} is then used to compute the similarity between $\mathcal{F}_t$ and $\mathcal{C}_{t \rightarrow c}$. More specifically, we use $L$ multi-head attention layers, splitting the $C$ feature channel dimensions into $N_h$ groups such that $C_h = C / N_h$. Feature updates are computed per head and each may have different representations, which increases expressiveness. For each attention head $h$, a set of linear projections are used to compute queries $\mathcal{Q}_h$ from the target features $\mathcal{F}_t$, and keys $\mathcal{K}_h$ and values $\mathcal{V}_h$ from the feature volume $\mathcal{C}_{t \rightarrow c}$:
\begin{align}
\small
    \mathcal{Q}_h &= \mathcal{F}_t W_{\mathcal{Q}_h}  + b_{\mathcal{Q}_h} \nonumber \\
    \mathcal{K}_h &= \mathcal{C}_{t \rightarrow c} W_{\mathcal{K}_h}  + b_{\mathcal{K}_h} \\
    \mathcal{V}_h &= \mathcal{C}_{t \rightarrow c} W_{\mathcal{V}_h} + b_{\mathcal{V}_h} \nonumber
\end{align}
with $W_{\mathcal{Q}_h}$,$W_{\mathcal{K}_h}$,$W_{\mathcal{V}_h}$ $\in \mathbb{R}^{C_h \times C_h}$, and $b_{\mathcal{Q}_h}$,$b_{\mathcal{K}_h}$,$b_{\mathcal{V}_h}$ $\in \mathbb{R}^{C_h}$. Similarities are normalized per-bin using \emph{softmax} to obtain the attention values $\alpha_h \in \mathbb{R}^{N_h \times D}$:
\vspace{-2mm}
\begin{equation}
\small
\alpha_h = \emph{softmax} \Big( \frac{\mathcal{Q}_h^T \mathcal{K}_h}{\sqrt{C_h}} \Big) \Bigg|_D
\end{equation}

The output values $\mathcal{V} \in \mathbb{R}^{C}$ are obtained as a weighted concatenation of per-head output values:
\begin{equation}
\small
    \mathcal{V} = \big( \alpha_1 \mathcal{V}_1 \oplus \dots \oplus  \alpha_{N_h} \mathcal{V}_{N_h} \big) W_{\mathcal{O}} + b_{\mathcal{O}}
\end{equation}
where $W_\mathcal{O} \in \mathbb{R}^{C \times C}$ and $b_\mathcal{O} \in \mathbb{R}^{C}$, and $\oplus$ is the concatenation operation. Similarly, per-bin attention values $\alpha = \frac{1}{N_h} \sum_h \alpha_h$ are obtained by averaging over the number of heads.
This process is repeated $L$ times, each using the output values to update the feature volume for key and value calculation, such that $\mathcal{C}^{l+1}_{t \rightarrow c} = \mathcal{V}^{l}$. The final attention values are used to populate a \emph{cross-attention cost volume} $\mathcal{A}_{t \rightarrow c}$, a $H/4 \times W/4 \times D$ structure encoding the similarity between each feature in $\mathcal{F}_t$ and its matching candidates in $\mathcal{C}_{t \rightarrow c}$. Each $(u,v,i)$ cell of $\mathcal{A}_{t \rightarrow c}$ receives the corresponding attention value $\alpha(u'_i,v'_i)$ from the last cross-attention layer L as the similarity metric for feature matching.

In Figure \ref{fig:histogram} we show the impact of our proposed cross-attention matching refinement procedure. In Figure \ref{fig:histogram_ssim} the input features are used directly to build a similarity cost volume using SSIM (Equation \ref{eq:photo_mono}), similar to \cite{manydepth,monorec}, and in Figure \ref{fig:histogram_cross} we use the refined cross-attention weights generated from the same features. After refinement the matching distributions are sharper (see Figure \ref{fig:histogram_distrib} for per-pixel examples), resulting in a more robust cost volume without the ambiguities and local minima found in other non-learned appearance-based similarity metrics. 

\subsubsection{Self-Attention Refinement}

Similar to~\cite{sarlin20superglue}, we alternate \emph{cross-attention} between target $\mathcal{F}_t$ and sampled context features $\mathcal{C}_{t \rightarrow c}$ with \emph{self-attention} among epipolar-sampled context features. In this setting, queries $\mathcal{Q}'_h$ are also calculated from $\mathcal{C}_{t \rightarrow c}$, such that:
\begin{equation}
\small
    \mathcal{Q}'_h = \mathcal{C}_{t \rightarrow c} W'_{\mathcal{Q}_h} + b_{\mathcal{Q}_h}'
\end{equation}
The self-attention refinement step takes place after each cross-attention layer, and is repeated $L-1$ times. It is omitted from the last iteration because cross-attention weights $\alpha$ from the last layer L are used to populate $\mathcal{A}_{t \rightarrow c}$, not output values $\mathcal{V}$, so self-attention updates are not required.


\subsection{Cost Volume Decoding}
\label{sec:decoding}


\subsubsection{High-Response Depth Decoding}
\label{sec:high_response_decoding}

\begin{figure}[t!]
\vspace{-3mm}
    \centering
\begin{tabular}{cc}
\hspace{-5mm}
\multirow{3}{*}{
\raisebox{0mm}[34mm]{
\subfloat[High-response window]{
\includegraphics[width=0.30\textwidth,height=4.5cm]{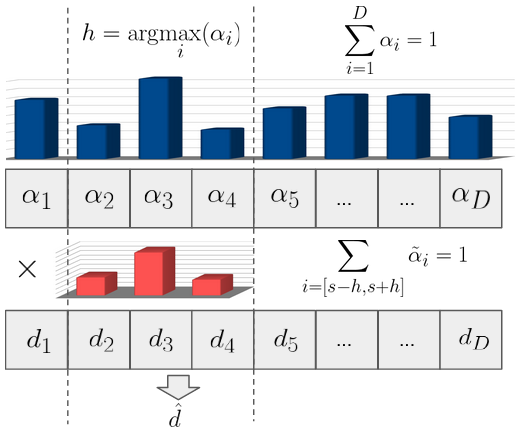}
\label{fig:high_response_a}
}
}
}
&
\hspace{-4mm}
    \subfloat[Input image]{
    \includegraphics[width=0.15\textwidth,height=1.2cm]{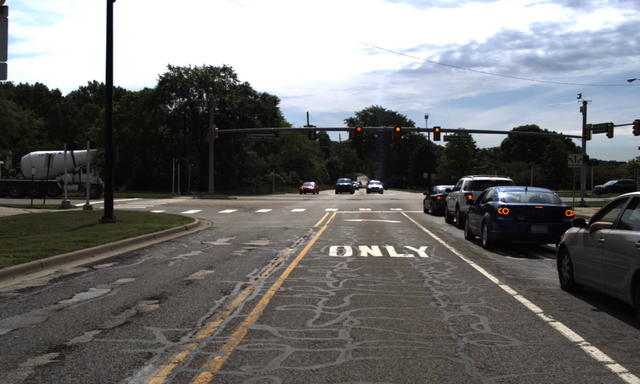}
    \label{fig:high_response_b}
    }
\\
&
\hspace{-4mm}
    \subfloat[Maximum attention]{
    \includegraphics[width=0.15\textwidth,height=1.2cm]{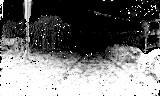}
    \label{fig:high_response_c}
    }
\\
&
\hspace{-4mm}
    \subfloat[High-response depth]{
    \includegraphics[width=0.15\textwidth,height=1.2cm]{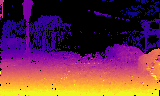}
    \label{fig:high_response_d}
    }
\\ 
    \end{tabular}
    \caption{\textbf{High response depth estimation} from a cross-attention cost volume. Instead of a weighted sum of all candidates, we use a window centered on the most probable candidate.}
\label{fig:high_response2}
\vspace{-4mm}
\end{figure}

We use a localized high-response window~\cite{high-response} to estimate continuous depth values from discretized bins, thus increasing robustness to multi-modal distributions~\cite{sttr}. A diagram is shown in Figure \ref{fig:high_response_a}, and below we describe each step. For each pixel $\textbf{p}_{uv}$, the $\arg\!\max$ operation is used to find the index $h_{uv}$ of the most probable $\alpha$ alongside its sampled epipolar line $\mathcal{E}_{t \rightarrow c}^{uv}$. A 1-dimensional $2s+1$ window is placed around $h_{uv}$, and a re-normalization step is applied:
\begin{equation}
\small
    \tilde{\alpha}_i = \frac{\alpha_i}{\sum_i a_i}, \quad \text{for} \quad i \in \left[h-s,h+s\right]
\end{equation}
such that its sum is 1. The depth value for $\textbf{p}_{uv}$ is calculated by multiplying this re-normalized distribution with the corresponding depth bins:
\begin{equation}
\small
    \hat{d}_H = \sum_{i \in \left[ h-s,h+s\right]} d_i \tilde{\alpha}_i
\end{equation}
The normalized attention values can also be used as a measure of matching confidence, as shown in Figure \ref{fig:high_response_c}. In particular, maximum attention values have a clear tendency to decrease at longer depth ranges and particularly towards the vanishing point, which is expected due to resolution degradation and small motion between frames. We leverage this novel matching confidence metric by masking out pixels with maximum attention value below a certain threshold $\lambda_{min}$, both from the high response loss calculation and the decoded  features (Figure \ref{fig:high_response_d}). Evaluation for these intermediate depth maps are provided in Table \ref{table:kitti_intermediate}.




\subsubsection{Context-Adjusted Depth Decoding}
\label{sec:cal_decoding}

Because our proposed cross-attention cost volume is regressed over epipolar lines, it lacks surrounding context information. To address this limitation, we use a context adjustment layer similar to~\cite{sttr}, where estimated depth values are adjusted via conditioning with input images. This adjustment is \emph{residual}, with the output being added to the normalized high-response depth map $\hat{D}_H$ before it is restored using the same statistics. For more details, including qualitative examples, please refer to the supplementary material.

\subsubsection{Multi-Scale Depth Decoding}
\label{sec:monocular_decoding}

Generating cost volumes from monocular information has two main limitations: (i) it requires ego-motion, and will fail if the camera is static between frames; (ii) it assumes a static world, and will fail in the presence of dynamic objects. To circumvent these limitations, recent methods~\cite{manydepth,monorec} have proposed combining multi-frame cost volumes with features from a single-frame depth network. These features are then decoded jointly, which makes predicted depth maps robust to multi-frame failure cases.

\begin{figure}[t!]
\vspace{-3mm}
    \centering
    \includegraphics[width=0.47\textwidth,height=5cm]{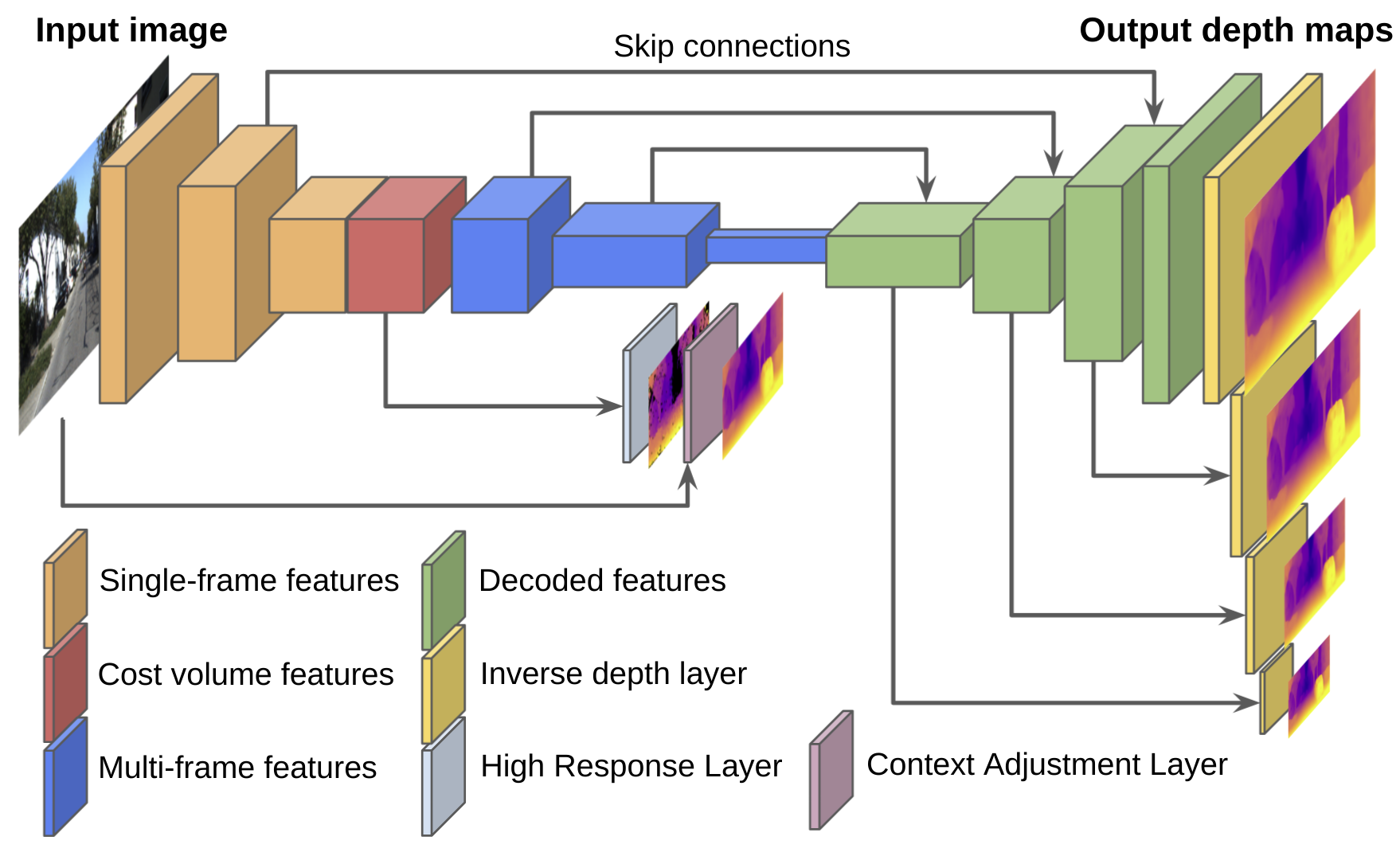}
    \caption{
        \textbf{Decoding architecture} used in our experiments. A single-frame depth network is augmented to include multi-frame cross-attention features  (Section \ref{sec:cross_attention}), generating depth maps at multiple scales in addition to those generated directly from multi-frame attention (Sections \ref{sec:high_response_decoding} and \ref{sec:cal_decoding}). 
    }
    \label{fig:decoder}
\vspace{-4mm}
\end{figure}

Our multi-scale decoding architecture is shown in Figure \ref{fig:decoder}. The cross-attention cost volume (Figure \ref{fig:cross-attention}) is first masked out, removing pixels with low matching confidence, and then concatenated with single-frame features from $I_t$ encoded by a separate network. A bottleneck convolutional layer is used to combine these two feature maps, and the output is decoded to produce $S$ depth estimates at multiple increasing resolutions.  Similar to~\cite{manydepth}, we use a teacher-student training procedure, improving the performance of multi-frame predictions via the supervision of a single-frame depth network in areas where cost volume generation fails. This single-frame depth network is trained jointly, sharing the same pose predictions, and discarded during evaluation.

\begin{figure*}[t!]
\vspace{-3mm}
\small
    \label{fig:cost_volume_depth}
    \centering
    \rotatebox{90}{\hspace{2mm} RGB} \hspace{-1.5mm}
    \subfloat{
    \includegraphics[width=0.19\textwidth,height=1.15cm]{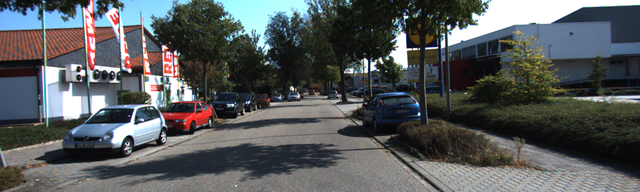}}
    \subfloat{
    \includegraphics[width=0.19\textwidth,height=1.15cm]{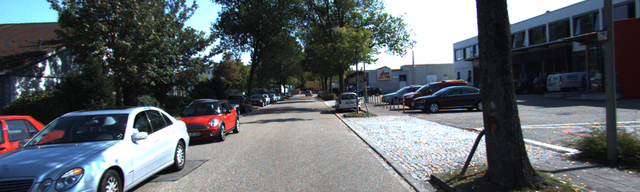}}
    \subfloat{
    \includegraphics[width=0.19\textwidth,height=1.15cm]{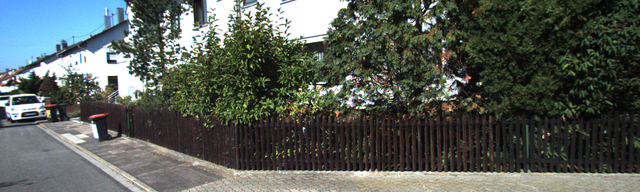}}
    \subfloat{
    \includegraphics[width=0.19\textwidth,height=1.15cm]{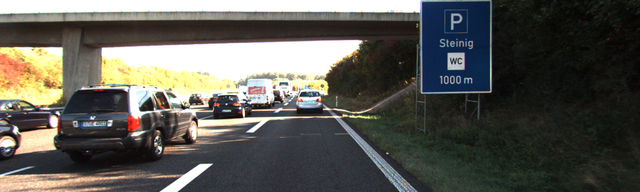}}
    \subfloat{
    \includegraphics[width=0.19\textwidth,height=1.15cm]{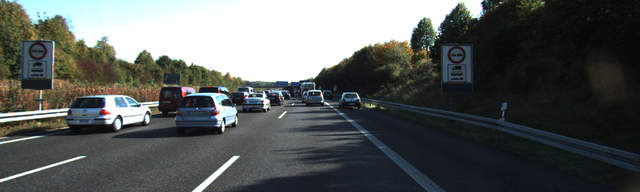}}
    \\
    \rotatebox{90}{\hspace{1mm} SSIM} \hspace{-1.5mm}
    \subfloat{
    \includegraphics[width=0.19\textwidth,height=1.15cm]{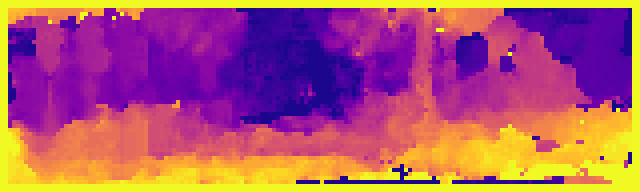}}
    \subfloat{
    \includegraphics[width=0.19\textwidth,height=1.15cm]{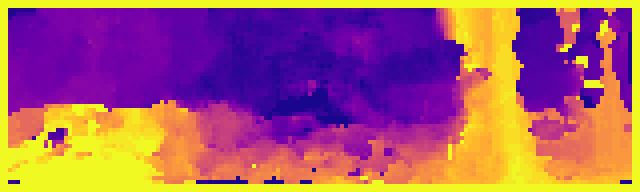}}
    \subfloat{
    \includegraphics[width=0.19\textwidth,height=1.15cm]{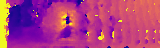}}
    \subfloat{
    \includegraphics[width=0.19\textwidth,height=1.15cm]{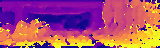}}
    \subfloat{
    \includegraphics[width=0.19\textwidth,height=1.15cm]{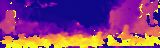}}
    \\
    \rotatebox{90}{\hspace{1mm} Cross} \hspace{-1.5mm}
    \subfloat{
    \includegraphics[width=0.19\textwidth,height=1.15cm]{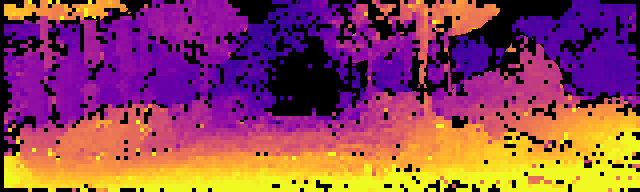}}
    \subfloat{
    \includegraphics[width=0.19\textwidth,height=1.15cm]{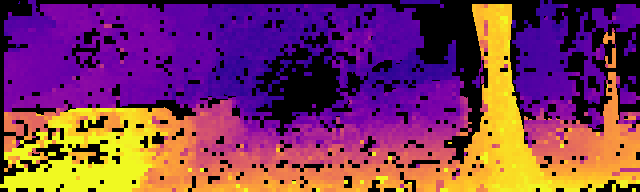}}
    \subfloat{
    \includegraphics[width=0.19\textwidth,height=1.15cm]{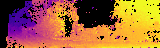}}
    \subfloat{
    \includegraphics[width=0.19\textwidth,height=1.15cm]{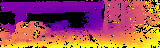}}
    \subfloat{
    \includegraphics[width=0.19\textwidth,height=1.15cm]{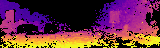}}
    \\
    \rotatebox{90}{\hspace{0mm}Decoded} \hspace{-1.5mm}
    \subfloat{
    \includegraphics[width=0.19\textwidth,height=1.15cm]{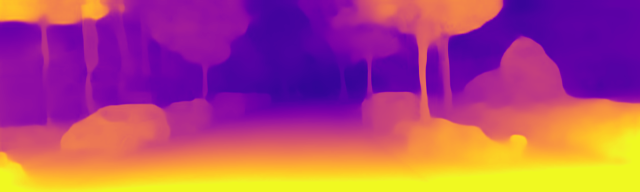}}
    \subfloat{
    \includegraphics[width=0.19\textwidth,height=1.15cm]{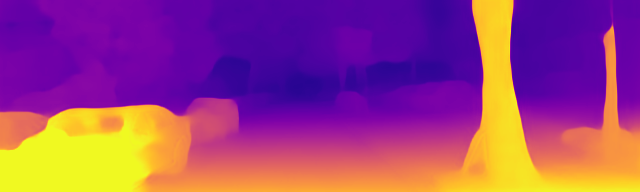}}
    \subfloat{
    \includegraphics[width=0.19\textwidth,height=1.15cm]{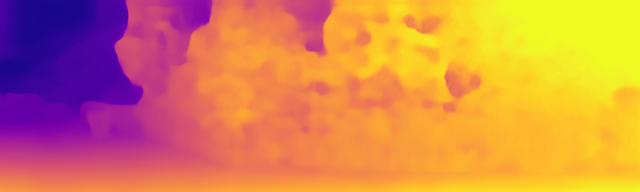}}
    \subfloat{
    \includegraphics[width=0.19\textwidth,height=1.15cm]{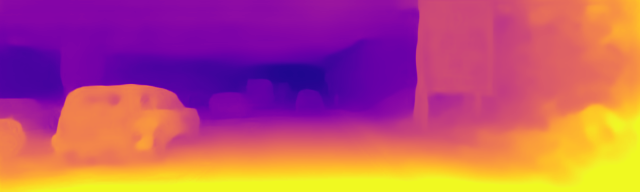}}
    \subfloat{
    \includegraphics[width=0.19\textwidth,height=1.15cm]{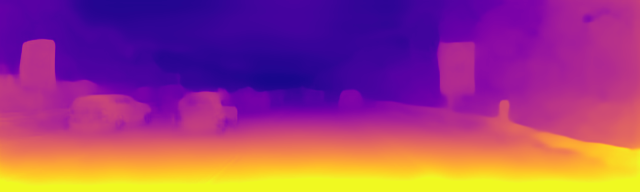}}
    \\
    \vspace{-1mm}
    \caption{
        \textbf{Predicted depth maps} on the KITTI dataset, obtained from SSIM ($\arg\!\min$) and cross-attention (high response) cost volumes, as well as the decoded depth estimation output (full resolution). Corresponding quantitative results are reported in Table \ref{table:kitti_intermediate}.
    \label{fig:kitti_intermediate}
    \vspace{-3mm}
}
\end{figure*}

\subsection{Training Loss}
\label{sec:losses}

We train our self-supervised depth and ego-motion architecture end-to-end using only the photometric reprojection loss, consisting of a weighted sum between a structure similarity (SSIM)~\cite{wang2004image} and absolute error (L1) terms:
\begin{equation}
\small
\mathcal{L}_{p} = \alpha~\frac{1 - \text{SSIM}(I_t,\hat{I_t})}{2} + (1-\alpha)~\| I_t - \hat{I_t} \|
\label{eq:photo_mono}
\end{equation}
Following standard procedure, we also use depth regularization~\cite{godard2017unsupervised} to enforce smoothness in low-textured regions:
\begin{equation}
\small
\mathcal{L}_{s} = 
\scriptsize
\frac{1}{HW} 
\small
\sum_{u,v} 
|\delta_u \hat{d}_{uv}| e^{-||\delta_u I_{uv}||} + 
|\delta_v \hat{d}_{uv}| e^{-||\delta_v I_{uv}||}    
\end{equation}
These two terms are combined to produce the final training loss $\mathcal{L} = \mathcal{L}_p + \lambda_s \mathcal{L}_s$, which is aggregated across all predicted depth maps: $\hat{D}_H$ (high response, Section \ref{sec:high_response_decoding}), $\hat{D}_C$ (context adjustment, Section \ref{sec:cal_decoding}), and $\hat{D}_{M}$ (multi-scale, Section \ref{sec:monocular_decoding}) as follows:
\begin{equation}
\mathcal{L} = \lambda_H \mathcal{L}_H + \lambda_C \mathcal{L}_C +\sum_{i=1}^S \frac{1}{2^i} \mathcal{L}_{M_i}
\end{equation}

\begin{figure}[t!]
    \vspace{-1mm}
    \centering
    \subfloat{
    \includegraphics[width=0.15\textwidth,height=1.15cm]{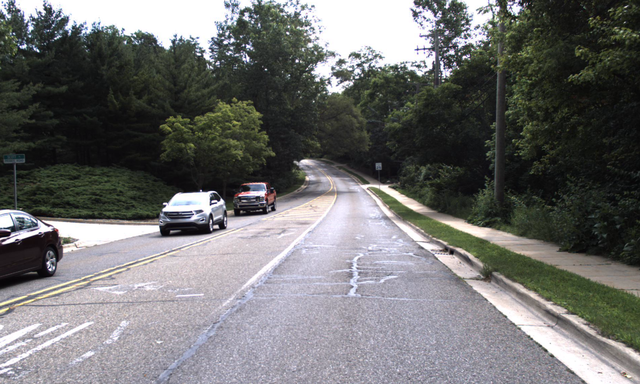}}
    \subfloat{
    \includegraphics[width=0.15\textwidth,height=1.15cm]{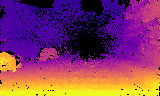}}
    \subfloat{
    \includegraphics[width=0.15\textwidth,height=1.15cm]{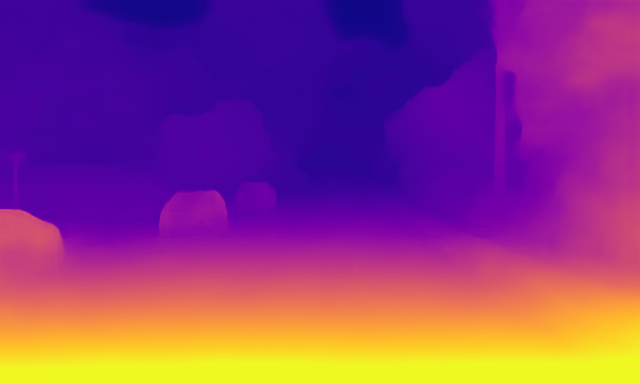}}
    \\ 
    \subfloat{
    \includegraphics[width=0.15\textwidth,height=1.15cm]{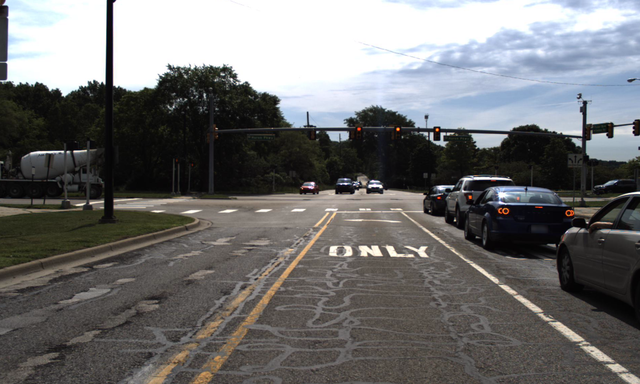}}
    \subfloat{
    \includegraphics[width=0.15\textwidth,height=1.15cm]{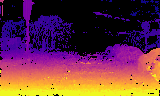}}
    \subfloat{
    \includegraphics[width=0.15\textwidth,height=1.15cm]{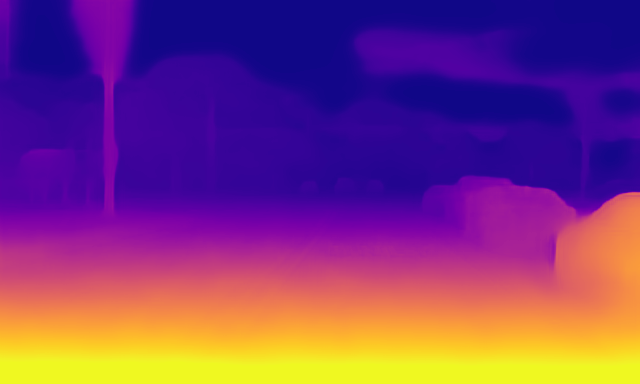}}
    \\ 
    \setcounter{subfigure}{0}
    \subfloat[Input target image]{
    \includegraphics[width=0.15\textwidth,height=1.15cm]{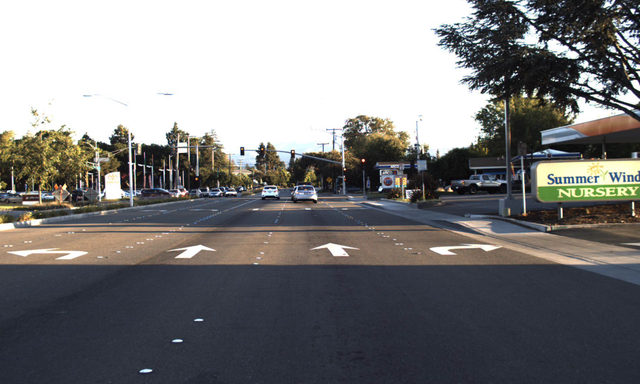}}
    \subfloat[High-response depth]{
    \includegraphics[width=0.15\textwidth,height=1.15cm]{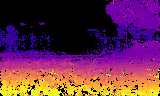}}
    \subfloat[Decoded depth]{
    \includegraphics[width=0.15\textwidth,height=1.15cm]{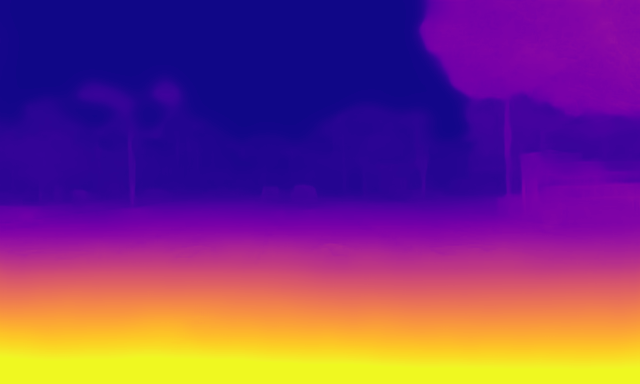}}
    \\    
    \vspace{-1mm}
    \caption{\textbf{Predicted depth maps} on the DDAD dataset. 
    \label{fig:ddad_depth}
    \vspace{-3mm}
}
\end{figure}

\section{Experiments}

\captionsetup[table]{skip=6pt}

\begin{table*}[t!]
\small
\renewcommand{\arraystretch}{0.9}
\centering
{
\small
\setlength{\tabcolsep}{0.3em}
\begin{tabular}{l|c|c|c|cccc|ccc}
\toprule
\multirow{2}[2]{*}{\textbf{Method}} & 
\multirow{2}[2]{*}{\rotatebox{90}{\scriptsize{Multi-Fr.}}} & 
\multirow{2}[2]{*}{\rotatebox{90}{\scriptsize{Synthetic}}} & 
\multirow{2}[2]{*}{\rotatebox{90}{\scriptsize{Semantic}}} & 
\multicolumn{4}{c|}{\textit{Lower is better}} &
\multicolumn{3}{c}{\textit{Higher is better}} 
\\
\cmidrule(lr){5-8} \cmidrule(lr){9-11}
& & & &
AbsRel &
SqRel &
RMSE &
RMSE$_{log}$ &
$\delta < {1.25}$ &
$\delta < {1.25}^2$ &
$\delta < {1.25}^3$
\\
\midrule
Struct2Depth \cite{casser2018depth}
& & & \checkmark
& 0.141 & 1.026 & 5.291 & 0.215 & 0.816 & 0.945 & 0.979 
\\
Gordon \emph{et al.} \cite{gordon2019depth}
& & & \checkmark
& 0.128 & 0.959 & 5.230 & 0.212 & 0.845 & 0.947 & 0.976 
\\
GASDA \cite{gasda}
& & \checkmark &
& 0.120 & 1.022 & 5.162 & 0.215 & 0.848 & 0.944 & 0.974
\\
SharinGAN \cite{sharingan}
& & \checkmark &
& 0.116 & 0.939 & 5.068 & 0.203 & 0.850 & 0.948 & 0.978
\\
Monodepth2 \cite{monodepth2}
& & & 
& 0.115 & 0.903 & 4.863 & 0.193 & 0.877 & 0.959 & 0.981 
\\
Patil \emph{et al.} \cite{forget_past}
& \checkmark & & 
& 0.111 & 0.821 & 4.650 & 0.187 & 0.883 & 0.961 & 0.982 
\\
PackNet-SFM \cite{packnet}
& & & 
& 0.111 & 0.785 & 4.601 & 0.189 & 0.878 & 0.960 & 0.982 
\\
GUDA \cite{guda}
& & \checkmark & 
& 0.107 & 0.714 & 4.421 & --- & 0.883 & --- & --- 
\\
Johnston \emph{et al.} \cite{johnston2020selfsupervised}
& & & 
& 0.106 & 0.861 & 4.699 & 0.185 & 0.889 & 0.962 & 0.982 
\\
Wang \emph{et al.} \cite{implicit_cues}
& \checkmark & & 
& 0.106 & 0.799 & 4.662 & 0.187 & 0.889 & 0.961 & 0.982 
\\
MonoDEVSNet \cite{virtualworld}
& & \checkmark &
& 0.104 & 0.721 & 4.396 & 0.185 & 0.880 & 0.962 & \underline{0.983}
\\
TC-Depth \cite{ruhkamp2021attention}
& \checkmark & &
& 0.103 & 0.746 & 4.483 & 0.185 & 0.894 & --- & \underline{0.983}
\\
Guizilini \emph{et al.} \cite{packnet-semguided}
& & & \checkmark
& 0.102 & \underline{0.698} & \underline{4.381} & 0.178 & 0.896 & 0.964 & \textbf{0.984} 
\\
ManyDepth \cite{manydepth} 
& \checkmark & & 
& \underline{0.098} & 0.770 & 4.459 & \underline{0.176} & \underline{0.900} & \underline{0.965} & \underline{0.983} 
\\
\midrule
\textbf{DepthFormer} 
& \checkmark & & 
& \textbf{0.090} & \textbf{0.661} & \textbf{4.149} & \textbf{0.175} & \textbf{0.905} & \textbf{0.967} & \textbf{0.984} 
\\
\bottomrule
\end{tabular}
}
\caption{
\textbf{Depth estimation results} on the KITTI \emph{Eigen} test split \cite{eigen2014depth}, for distances up to 80m with the \emph{Garg} crop~\cite{garg2016unsupervised} and half resolution ($640 \times 192$ or similar). \emph{Multi-Fr.} indicates the use of multiple frames at test time; \emph{Synthetic} the use of additional synthetic training data; and \emph{Semantic} the use of additional semantic information. 
}
\label{table:kitti_depth_selfsup}
\vspace{-4mm}
\end{table*}

\captionsetup[table]{skip=6pt}

\begin{table}[t!]
\small
\renewcommand{\arraystretch}{0.9}
\centering
{
\small
\setlength{\tabcolsep}{0.3em}
\begin{tabular}{l|ccc}
\toprule
\textbf{Method} & AbsRel$\downarrow$ & RMSE$\downarrow$ & $\delta < {1.25}$$\uparrow$ 
\\
\midrule
SAD Depth ($\arg\!\min$) & 
0.647 & 17.662 & 0.575 
\\
SSIM Depth ($\arg\!\min$) & 
0.632 & 17.124 & 0.598 
\\
\midrule
High-Response Depth & 
0.264 & 10.919 & 0.714 
\\
Context-Adjusted Depth &
0.167 & 6.367 & 0.808 
\\
\midrule
Decoded Depth (1/8) & 
0.095 & 4.336 & 0.892 
\\
Decoded Depth (1/4) & 
\underline{0.091} & 4.201 & 0.900 
\\
Decoded Depth (1/2) & 
\textbf{0.090} & \textbf{4.146} & \underline{0.904}
\\
Decoded Depth (Full) & 
\textbf{0.090} & \underline{4.149} & \textbf{0.905} 
\\
\bottomrule
\end{tabular}
}
\caption{
\textbf{Intermediate depth estimation results} of our architecture on the KITTI dataset (Table \ref{table:kitti_depth_selfsup}, Figure \ref{fig:kitti_intermediate}). 
}
\label{table:kitti_intermediate}
\vspace{-5mm}
\end{table}

\subsection{Datasets}

\noindent\textbf{KITTI~\cite{geiger2013vision}}
The KITTI dataset is the standard benchmark for depth evaluation. To compare with other methods, we adopt the training protocol from Eigen \emph{et al.}~\cite{eigen2014depth}, with Zhou \emph{et al.}'s~\cite{zhou2017unsupervised} filtering of static frames, resulting in $39810$/$4424$/$697$ training, validation, and test images.

\noindent\textbf{DDAD~\cite{packnet}}
The DDAD dataset is a novel benchmark for depth evaluation, with denser ground-truth and longer ranges, which is particularly challenging for multi-frame methods. Following \cite{packnet}, we use only the front camera, resulting in $12560$/$3950$ training and validation images. 

\noindent\textbf{Cityscapes~\cite{cordts2016cityscapes}} We use the Cityscapes dataset to test the generalization properties of our proposed cross-attention module. We use the $2975$ training images with their $30$-frame sequences, for a total of $89250$ images. 

\noindent\textbf{VKITTI2~\cite{cabon2020vkitti2}}
The Virtual KITTI 2 dataset contains reconstructions of five sequences from the KITTI odometry benchmark\cite{Geiger2012CVPR}, for a total of $12936$ samples in varying weather conditions and time of day.

\noindent\textbf{Parallel Domain~\cite{parallel_domain}}
The Parallel Domain dataset, recently introduced in \cite{guda}, contains procedurally-generated and fully annotated renderings of urban driving scenes. It contains $42000$/$8000$ training and validation samples. 

\noindent\textbf{TartanAir~\cite{wang2020tartanair}}
TartanAIR is a synthetic photo-realistic  dataset for visual SLAM. We train on monocular videos, and following~\cite{teed2021droid} select context images only if the average optical flow magnitude is between $8$ and $96$ pixels. Our total training set consists of $189696$ images.

\subsection{Implementation Details}

Our models are implemented using PyTorch~\cite{paszke2017automatic} and trained across 8 Titan V100 GPUs.
We use the Adam optimizer~\cite{kingma2014adam}, with $\beta_1=0.9$ and $\beta_2=0.999$, and a batch size of $1$ per GPU. Our networks are trained for $50$ epochs, with an initial learning rate of $2 \cdot 10^{-4}$ that is halved every 20 epochs. Following \cite{manydepth}, we freeze the pose and single-frame teacher network for the final 5 epochs. We use frame $t-1$ as context for cost volume calculation, and frames $t-1$ and $t+1$ for loss calculation. Our training and network parameters are: SSIM weight $\alpha=0.85$, smoothness weight $\lambda_s=10^{-4}$, high-response and context-adjusted weights $\lambda_H=\lambda_C=0.5$, minimum attention $\lambda_{min}=0.1$, high-response window size $s=1$, epipolar depth bins $D=128$, attention dimension $C=128$, attention heads $h=8$, attention layers $N=6$, number of output scales $S=4$. For more details, please refer to the supplementary material.

\subsection{Depth Evaluation} 

To validate our DepthFormer architecture, we conducted a thorough comparison of its performance relative to other published methods. Our findings targeting the KITTI dataset, considered the standard benchmark for this task, are summarized in Table \ref{table:kitti_depth_selfsup}. We consistently outperform all other considered methods by a large margin, including single-frame and multi-frame methods, and even those that leverage additional information in the form of semantic labels~\cite{packnet-semguided,gordon2019depth,casser2018depth} or synthetic data~\cite{gasda,sharingan,virtualworld,johnston2020selfsupervised,guda}. In particular, we significantly improve upon ManyDepth~\cite{manydepth}, that uses a similar depth decoding strategy but relies directly on the sum of absolute differences (SAD) as the similarity metric, without any feature matching refining strategy. Our architecture also compares favourably to single-frame supervised methods, outperforming the current state of the art (more details in the supplementary material).

In Table \ref{table:kitti_intermediate} we show intermediate depth estimation results from the various outputs of our architecture, with qualitative examples in Figure \ref{fig:kitti_intermediate}. By replacing SAD or SSIM cost volumes with our cross-attention cost volume with high-response depth self-supervision, we already significantly improve performance, from an Abs.Rel. of $0.647$ and $0.632$ to $0.264$. These results are further improved after context adjustment, to account for low confidence matches, occlusions and inaccuracies in epipolar projection, achieving $0.167$. Finally, by combining multi-frame cross-attention with single-frame features for joint decoding, to reason over multi-frame failure cases, we achieve the reported result of $0.090$. Interestingly, decoded depth maps at lower resolutions perform almost as well as the full resolution output. We attribute this behavior to the cross-attention cost volume, that is calculated at a lower resolution (1/4) and connected to the decoder via skip connections. Although high resolution decoding is beneficial, it is not necessary for our reported state-of-the-art performance.

\captionsetup[table]{skip=6pt}

\begin{table}[t!]
\renewcommand{\arraystretch}{0.9}
\centering
{
\small
\setlength{\tabcolsep}{0.3em}
\begin{tabular}{l|ccc|c}
\toprule
\textbf{Method} & 
AbsRel$\downarrow$ &
SqRel$\downarrow$ &
RMSE$\downarrow$ &
$\delta < {1.25}$$\uparrow$ 
\\
\midrule
Monodepth2 \cite{monodepth2}
& 0.213 & 4.975 & 18.051 & 0.761
\\
PackNet-SFM \cite{packnet}
& 0.162 & 3.917 & 13.452 & 0.823 
\\
GUDA$^\dagger$ \cite{guda}
& 0.147 & \textbf{2.922} & 14.452 & 0.809
\\
ManyDepth \cite{manydepth}
& 0.146 & \textbf{3.258} & 14.098 & 0.822
\\
\midrule
\textbf{DepthFormer} 
& \textbf{0.135} & \underline{2.953} & \textbf{12.477} & \textbf{0.836} 
\\
\bottomrule
\end{tabular}
}
\caption{
\textbf{Depth estimation results} on the DDAD validation split \cite{packnet}, for distances up to 200m without any cropping (Figure \ref{fig:ddad_depth}). The symbol $^\dagger$ indicates supervision from synthetic data. 
}
\label{table:ddad_depth}
\vspace{-4mm}
\end{table}

We also performed experiments on the DDAD dataset, which is a more challenging benchmark due to its longer depth ranges and larger number of dynamic objects. Even under these conditions, our DepthFormer architecture achieves state-of-the-art results, as shown in Table \ref{table:ddad_depth}, with qualitative examples in Figure \ref{fig:ddad_depth}.

\subsection{Ablation Analysis}

In Table \ref{table:ablation} we provide an analysis of the different components used in our DepthFormer architecture, including depth estimation results and memory requirements. Firstly, we analyze the impact of our proposed cross-attention module, showing that it is crucial for the reported state-of-the-art performance. We also show that optimizing the cross-attention cost volume itself, via self-supervision on the high-response and context-adjust depth maps, is key to our reported performance as well. This is expected, since without them the cross-attention features are only used in the context of joint single-frame decoding, rather than optimized to generate multi-frame-only depth estimates as well. Similarly, removing self-attention calculation from context features also degrades results. We also ablated different high-response and context-adjusted weights, achieving similar results between $\lambda_H = \lambda_C = \left[0.1, 1.0\right]$. 

\captionsetup[table]{skip=6pt}

\begin{table}[t!]
\vspace{-2mm}
\small
\renewcommand{\arraystretch}{0.9}
\centering
{
\small
\setlength{\tabcolsep}{0.3em}
\begin{tabular}{l|ccc|cc}
\toprule
\multirow{2}{*}{\textbf{Method}} & \multicolumn{3}{c}{Depth Evaluation} & \multicolumn{2}{|c}{GPU (GB)}
\\
\cmidrule(lr){2-4} \cmidrule(lr){5-6}
& AbsRel$\downarrow$ & RMSE$\downarrow$ & $\delta < {1.25}$ $\uparrow$ & train & test 
\\
\midrule
W/o cross-attn. & 
0.099 & 4.430 & 0.900 & 3.8 & 2.9
\\
W/o cross-attn. loss & 
0.103 & 4.581 & 0.892 & 12.1 & 5.3 
\\
W/o self-attn. & 
0.094 & 4.259 & 0.901 & 12.2 & 5.3 
\\
\midrule
16 depth bins & 
0.101 & 4.595 & 0.894 & 6.6 & 3.2 
\\
48 depth bins & 
0.095 & 4.330 & 0.900 & 8.9 & 4.8
\\
96 depth bins & 
0.092 & 4.181 & 0.903 & 12.5 & 5.4
\\
\midrule
32 attn. channels & 
0.104 & 4.761 & 0.885 & 8.7 & 3.7 
\\
48 attn. channels & 
0.098 & 4.332 & 0.894 & 9.6 & 4.3  
\\
96 attn. channels & 
0.093 & 4.207 & 0.899 & 12.5 & 5.5 
\\
\midrule
2 attn. layers & 
0.094 & 4.388 & 0.901 & 11.4 & 6.1
\\
4 attn. layers & 
0.093 & 4.321 & 0.901 & 13.3 & 6.2
\\
\midrule
\textbf{DepthFormer} & 
{0.090} & {4.149} & {0.905} & 15.2 & 6.4
\\
\bottomrule
\end{tabular}
}
\caption{
\textbf{Ablation analysis} of the various components of our architecture, including depth evaluation and GPU requirements. 
}
\label{table:ablation}
\end{table}



We also experimented with different variations of our architecture, obtained by modifying the number of attention layers $N$, attention feature channels $C$, and depth bins $D$. These show a clear overall trend that increasing cross-attention network complexity leads to improved results. This is further evidence that better feature matching is beneficial to depth estimation, but also shows that competitive results can still be obtained with simpler configurations. We leave further exploration of more complex architectures, as well as efficiency improvements~\cite{tay2020efficient,liu2021swin}, to future work.

\subsection{Cost Volume Generalization}

Our proposed architecture is \emph{modular}, in the sense that the cross-attention network can be separated from the joint single-frame decoding architecture. In this section we explore to which extent we can re-utilize cross-attention cost volumes between datasets, building on the well-studied intuition~\cite{MING202114,ruhkamp2021attention,sttr,monorec,deepv2d} that geometric features are more transferable than appearance-based ones. To this end, we design three experiments, considering the KITTI dataset as target and multiple other datasets as source. In \emph{hot swap}, we replace the cross-attention network trained on the target dataset with one trained on a source dataset, maintaining the same single-frame and pose networks, without further training. In \emph{fine-tune (mono}), we train the single-frame and pose networks from scratch, and use a frozen cross-attention network pre-trained on a source dataset. In \emph{fine-tune (all)} we follow the same setting, but also jointly optimize the pre-trained cross-attention network on the target dataset. To fully leverage synthetic data, the VKITTI2, PD, and TartanAir models are pre-trained with depth supervision (using a Smooth L1 loss) and use ground-truth relative poses. Real-world datasets (DDAD and Cityscapes) are pre-trained using the self-supervised loss described in Section \ref{sec:losses}. 

\captionsetup[table]{skip=6pt}

\begin{table}[t!]
\vspace{-2mm}
\small
\renewcommand{\arraystretch}{0.9}
\centering
{
\small
\setlength{\tabcolsep}{0.3em}
\begin{tabular}{l|l|ccc}
\toprule
\textbf{Dataset} & \textbf{Variation} & AbsRel$\downarrow$ & RMSE$\downarrow$ & $\delta < {1.25}$ $\uparrow$ 
\\
\midrule
\multirow{3}[2]{*}{\textbf{DDAD}} 
& Hot swap & 0.098 & 4.364 & 0.899 \\
& Fine-tune (mono) & 0.099 & 4.336 & 0.902 \\
& Fine-tune (all) & 0.091 & 4.187 & 0.904 \\
\midrule
\multirow{3}[2]{*}{\textbf{Cityscapes}}
& Hot swap & 0.097 & 4.339 & 0.897 \\
& Fine-tune (mono) & 0.096 & 4.291 & 0.899 \\
& Fine-tune (all) & 0.090 & 4.138 & 0.905 \\
\midrule
\multirow{3}[2]{*}{\textbf{VKITTI2}}
& Hot swap & 0.094 & 4.302 & 0.898 \\
& Fine-tune (mono) & 0.094 & 4.232 & 0.899 \\
& Fine-tune (all) & 0.091 & 4.192 & 0.904 \\
\midrule
\multirow{3}[2]{*}{\textbf{P. Domain}}
& Hot swap & 0.102 & 4.432 & 0.888 \\
& Fine-tune (mono) & 0.097 & 4.295 & 0.897 \\
& Fine-tune (all) & 0.090 & 4.110 & 0.904 \\
\midrule
\multirow{3}[2]{*}{\textbf{TartanAir}}
& Hot swap & 0.102 & 4.532 & 0.886 \\
& Fine-tune (mono) & 0.095 & 4.397 & 0.897 \\
& Fine-tune (all) & 0.091 & 4.187 & 0.905 \\
\midrule
\midrule
\textbf{KITTI} & 
\multicolumn{1}{|c|}{-----} & 0.090 & 4.149 & 0.905 
\\
\bottomrule
\end{tabular}
}
\caption{\textbf{Cross-attention cost volume generalization results} on the KITTI dataset, for different pre-trained source datasets. 
}
\label{table:kitti_generalization}
\end{table}

Results for these experiments are reported in Table \ref{table:kitti_generalization}. Interestingly, swapping the cross-attention network between datasets results in only a small degradation in performance, of around $5\%$. This indicates that the learned matching function is robust to distribution shifts between datasets. In fact, we achieved nearly identical results when only training the single-frame and pose networks from scratch, using a frozen cross-attention network pre-trained on a source dataset. However, because the cross-attention network is not optimized (i.e., it is kept frozen), training iterations are both faster (around $100\%$, from $7.3$ to $14.2$ FPS) and require less memory (around $20\%$, from $15.3$ to $12.4$ GB). Once convergence in this setting is achieved, we can reproduce the reported state-of-the-art results by fine-tuning all networks for only $5$ epochs, instead of the $50$ required when training the entire architecture from scratch. 




\section{Conclusion}

This paper proposes a novel attention-based cost volume generation procedure for multi-frame self-supervised monocular depth estimation. Our key contribution is a cross-attention module designed to refine feature matching between images, improving upon traditional appearance-based similarity metrics that are prone to ambiguity and local minima. We show that our cross-attention module leads to more robust matching, that is decoded into depth estimates and trained end-to-end using only a photometric objective. We establish a new state of the art on the KITTI and DDAD datasets, outperforming other single- and multi-frame self-supervised methods, and our results are even comparable to state-of-the-art single-frame supervised architectures. We also show that our learned cross-attention module is highly transferable, and can be used without fine-tuning across datasets to speed up convergence and decrease memory requirements at training time.


\appendix

\begin{table*}[!t]%
\renewcommand{\arraystretch}{0.96}
\small
\centering
\resizebox{0.45\linewidth}{!}{
\subfloat[\textbf{Single-frame depth network.} The target image $I_t$ is used as input, as well as the cross-attention cost volume $\mathcal{A}_{t \rightarrow c}$. Bold numbers indicate the 4 multi-scale output inverse depth maps, at increasing resolutions. Each sigmoid output is converted to depth using \emph{min} and \emph{max} ranges.]{
\begin{tabular}[b]{l|l|c|c|c}
\toprule
& \textbf{Layer Description} & \textbf{K} & \textbf{S} & \textbf{Output Dim.} \\ 
\toprule
\toprule
\multicolumn{5}{c}{\textbf{ResidualBlock (\#0)}} \\ 
\midrule
\#1 & Conv2d (\#0a) $\shortrightarrow$ BN $\shortrightarrow$ ReLU & K & 1 & \\
\#2 & Conv2d $\shortrightarrow$ BN $\shortrightarrow$ ReLU & K & S & \\
\#3 & Downsample(\#0) + \#2 $\shortrightarrow$ ReLU  & - & - &
\\
\toprule
\multicolumn{5}{c}{\textbf{UpsampleBlock (\#0, \#s)}} \\ 
\midrule
\#1 & Conv2d $\shortrightarrow$ BN $\shortrightarrow$ ReLU $\shortrightarrow$ Upsample & 3 & 1 & \\
\#2 & Conv2d (\#1 $\oplus$ \#s) $\shortrightarrow$ BN $\shortrightarrow$ ReLU  & 3 & 1 & \\
\toprule
\multicolumn{5}{c}{\textbf{InverseDepth (\#0)}} \\ 
\midrule
\#1 & Conv2d $\shortrightarrow$ Sigmoid & K & S & \\
\#2 & (\emph{max} - \emph{min}) $\odot$ \#1 + \emph{min} & - & - & \\
\toprule
\toprule
\#0a & Input RGB image & - & - & 3$\times$H$\times$W \\
\#0b & Input cost volume & - & - & 128$\times$H/4$\times$W/4 \\ 
\midrule
\multicolumn{5}{c}{\textbf{Encoder}} \\ \hline
\#1  & Conv2d $\shortrightarrow$ BN $\shortrightarrow$ ReLU   & 7 & 1 &  64$\times$H$\times$W \\
\#2  & Max. Pooling                 & 3 & 2 &  64$\times$H/2$\times$W/2 \\
\#3  & ResidualBlock (\#2) x2           & 3 & 1-2 &  64$\times$H/4$\times$W/4 \\
\#4  & \#3 $\oplus$ \#0b $\shortrightarrow$ Conv2d $\shortrightarrow$ BN $\shortrightarrow$ ReLU  & 3 & 1 & 64$\times$H/4$\times$W/4 \\
\#5  & ResidualBlock (\#4) x2           & 3 & 1-2 & 128$\times$H/8$\times$W/8 \\
\#6  & ResidualBlock (\#5) x2           & 3 & 1-2 & 256$\times$H/16$\times$W/16 \\
\#7  & ResidualBlock (\#6) x2           & 3 & 1-2 & 512$\times$H/32$\times$W/32 \\
\midrule
\multicolumn{5}{c}{\textbf{Decoder}} \\ 
\midrule
\#8 & UpsampleBlock (\#7,\#6)    & 3 & 1 & 256$\times$H/16$\times$W/16 \\
\#9 & UpsampleBlock (\#8,\#5)    & 3 & 1 & 128$\times$H/8$\times$W/8 \\
\textbf{\#10} & InverseDepth (\#8) & 3 & 1 & 1$\times$H/8$\times$W/8 \\
\#11 & UpsampleBlock (\#9,\#3)    & 3 & 1 & 64$\times$H/4$\times$W/4 \\
\textbf{\#12} & InverseDepth (\#11) & 3 & 1 & 1$\times$H/4$\times$W/4 \\
\#13 & UpsampleBlock (\#11,\#2)   & 3 & 1 & 32$\times$H/2$\times$W/2 \\
\textbf{\#14} & InverseDepth (\#13) & 3 & 1 & 1$\times$H/2$\times$W/2 \\
\#15 & UpsampleBlock (\#13,--)   & 3 & 1 & 32$\times$H$\times$W \\
\textbf{\#16} & InverseDepth (\#15) & 3 & 1 & 1$\times$H$\times$W \\
\bottomrule
\end{tabular}
}}
\resizebox{0.49\linewidth}{!}{
\subfloat[\textbf{Attention network.} It processes the target $I_t$ and context images $I_c$ independently, and the output is used to generate the cross-attention cost volume $\mathcal{A}_{t \rightarrow c}$, as described in Section 3.2.2, main paper.]{
\begin{tabular}[b]{l|l|c|c|c}
\toprule
& \textbf{Layer Description} & \textbf{K} & \textbf{S} & \textbf{Output Dim.} \\ 
\toprule
\toprule
\multicolumn{5}{c}{\textbf{ResidualBlock (\#0)}} \\ 
\midrule
\#1 & Conv2d $\shortrightarrow$ BN $\shortrightarrow$ ReLU & K & 1 & \\
\#2 & Conv2d $\shortrightarrow$ BN $\shortrightarrow$ ReLU & K & S & \\
\#3 & Downsample(\#0) + \#2 $\shortrightarrow$ ReLU  & - & - &
\\
\toprule
\multicolumn{5}{c}{\textbf{SpatialPyramidBlock (\#0, N)}} \\ 
\midrule
\#1  &  Avg. Pool & N & N & \\
\#2  &  Conv2d $\shortrightarrow$ BN $\shortrightarrow$ ReLU & K & S & \\
\toprule
\toprule
\#0 & Input RGB image & - & - & 6$\times$H$\times$W \\ 
\midrule
\#1  & Conv2d $\shortrightarrow$ BN $\shortrightarrow$ ReLU  & 3 & 2 & 16$\times$H/2$\times$W/2 \\
\#2  &  Conv2d $\shortrightarrow$ BN $\shortrightarrow$ ReLU  & 3 & 1 & 16$\times$H/2$\times$W/2 \\
\#3  &  Conv2d $\shortrightarrow$ BN $\shortrightarrow$ ReLU  & 3 & 1 & 32$\times$H/2$\times$W/2 \\
\#4  &  ResidualBlock (\#3)  & 3 & 2 & 64$\times$H/4$\times$W/4 \\
\#5  &  ResidualBlock (\#4)  & 3 & 2 & 128$\times$H/8$\times$W/8 \\
\#6  &  SpatialPyramidBlock (\#5,16)  & 1 & 1 & 32$\times$H/128$\times$W/128 \\
\#7  &  SpatialPyramidBlock (\#5,8)  & 1 & 1 & 32$\times$H/64$\times$W/64 \\
\#8  &  SpatialPyramidBlock (\#5,4)  & 1 & 1 & 32$\times$H/32$\times$W/32 \\
\#9  &  SpatialPyramidBlock (\#5,2)  & 1 & 1 & 32$\times$H/16$\times$W/16 \\
\#10  & Downsample(\#6 $\oplus$ \#7 $\oplus$ \#8 $\oplus$ \#9) & - & - & 128$\times$H/16$\times$W/16 \\
\#11 & DenseBlock (\#0 $\oplus$ \#10) & 1 & 1 & 128$\times$H$\times$W \\
\#12 & DenseBlock (\#4 $\oplus$ \#11) & 1 & 1 & 128$\times$H$\times$W \\
\#13 & DenseBlock (\#5 $\oplus$ \#12) & 1 & 1 & 128$\times$H$\times$W \\
\textbf{\#14} & DenseBlock (\#10 $\oplus$ \#13) & 1 & 1 & 128$\times$H$\times$W \\
\bottomrule
\end{tabular}
}} \\ \vspace{5mm}
\resizebox{0.49\linewidth}{!}{
\subfloat[\textbf{Pose network.} The target $I_t$ and context $I_c$ images are concatenated and used as input. The 6-dimensional output contains predicted relative translation $(x,y,z)$ and rotation $(roll,pitch,yaw)$ in Euler angles.]{
\begin{tabular}[b]{l|l|c|c|c}
\toprule
& \textbf{Layer Description} & \textbf{K} & \textbf{S} & \textbf{Output Dim.} \\ 
\toprule
\toprule
\multicolumn{5}{c}{\textbf{ResidualBlock (\#0)}} \\ 
\midrule
\#1 & Conv2d $\shortrightarrow$ BN $\shortrightarrow$ ReLU & K & 1 & \\
\#2 & Conv2d $\shortrightarrow$ BN $\shortrightarrow$ ReLU & K & S & \\
\#3 & Downsample(\#0) + \#2 $\shortrightarrow$ ReLU  & - & - &
\\
\toprule
\toprule
\#0 & Input 2 RGB images & - & - & 6$\times$H$\times$W \\ 
\midrule
\multicolumn{5}{c}{\textbf{Encoder}} \\ \hline
\#1  & Conv2d $\shortrightarrow$ BN $\shortrightarrow$ ReLU   & 7 & 1 &  64$\times$H$\times$W \\
\#2  & Max. Pooling                 & 3 & 2 &  64$\times$H/2$\times$W/2 \\
\#3  & ResidualBlock (\#2) x2           & 3 & 1-2 &  64$\times$H/4$\times$W/4 \\
\#4  & ResidualBlock (\#3) x2           & 3 & 1-2 & 128$\times$H/8$\times$W/8 \\
\#5  & ResidualBlock (\#4) x2           & 3 & 1-2 & 256$\times$H/16$\times$W/16 \\
\#6  & ResidualBlock (\#5) x2           & 3 & 1-2 & 512$\times$H/32$\times$W/32 \\
\midrule
\multicolumn{5}{c}{\textbf{Decoder}} \\ 
\midrule
\#7 & Conv2d $\shortrightarrow$ ReLU & 1 & 1 & 256$\times$H/32$\times$W/32 \\
\#8 & Conv2d $\shortrightarrow$ ReLU & 3 & 1 & 256 $\times$H/32$\times$W/32 \\
\#9 & Conv2d $\shortrightarrow$ ReLU & 3 & 1 & 256$\times$H/32$\times$W/32 \\
\#10 & Conv2d $\shortrightarrow$ ReLU & 1 & 1 & 6$\times$H/32$\times$W/32 \\
\textbf{\#11} & Global Avg. Pooling & - & - & 6$\times$1$\times$1 \\
\bottomrule
\end{tabular}
}}
\resizebox{0.49\linewidth}{!}{
\subfloat[\textbf{Context adjustment network.} Input high-response depth maps  $\hat{D}_H$ are normalized using Equation \ref{eq:norm}, and output depth maps $\tilde{D}_C$ are un-normalized using Equation \ref{eq:unnorm}.]{
\begin{tabular}[b]{l|l|c|c|c}
\toprule
& \textbf{Layer Description} & \textbf{K} & \textbf{S} & \textbf{Output Dim.} \\ 
\toprule
\toprule
\multicolumn{5}{c}{\textbf{ResidualBlock (\#0: N $\times$ H $\times$ W)}} \\ 
\midrule
\#1 & Conv2d $\shortrightarrow$ BN $\shortrightarrow$ ReLU & K & 1 & 3N $\times$H$\times$W \\
\#2 & Conv2d $\shortrightarrow$ BN $\shortrightarrow$ ReLU & K & S & N $\times$H$\times$W \\
\#3 & Downsample(\#0) + \#2 $\shortrightarrow$ ReLU & - & - & N $\times$H$\times$W \\
\toprule
\toprule
\#0a & Input RGB image & - & - & 3$\times$H$\times$W \\ 
\#0b & Input norm. depth map & - & - & 1$\times$H$\times$W \\ 
\midrule
\#1 & \#0a $\oplus$ \#0b & 3 & 1 & 4$\times$H$\times$W \\
\#2 & Conv2d $\shortrightarrow$ GN $\shortrightarrow$ ReLU & 3 & 1 & 16$\times$H$\times$W \\
\#3 & ResidualBlock($\oplus$ \#0b) x8 & 3 & 1 & 16$\times$H$\times$W \\
\textbf{\#4} & Conv2d + \#0b & 3 & 1 & 1$\times$H$\times$W \\
\bottomrule
\end{tabular}
}}
\caption{
\textbf{Network architectures used in our experiments.}  \emph{BN} stands for Batch Normalization \cite{ioffe2015batch}, \emph{Upsample} and \emph{Downsample} respectively increases and decreases spatial dimensions using bilinear interpolation to match the output resolution, \emph{ReLU} are Rectified Linear Units, \emph{Sigmoid} is the sigmoid activation function, and \emph{DenseBlock} are densely connected convolutional layers from \cite{huang2017densely}. The symbol $\oplus$ indicates feature concatenation, and $\odot$ indicates element-wise multiplication. 
}
\label{tab:networks}
\end{table*}


\captionsetup[table]{skip=6pt}

\begin{table*}[!t]
\small
\renewcommand{\arraystretch}{0.9}
\centering
{
\small
\setlength{\tabcolsep}{0.3em}
\begin{tabular}{l|c|c|cccc|ccc}
\toprule
\multirow{2}[2]{*}{\textbf{Method}} & 
\multirow{2}[2]{*}{\rotatebox{90}{\scriptsize{Superv.}}} &
\multirow{2}[2]{*}{\rotatebox{90}{\scriptsize{Multi-Fr.}}} & 
\multicolumn{4}{c|}{\textit{Lower is better}} &
\multicolumn{3}{c}{\textit{Higher is better}} 
\\
\cmidrule(lr){4-7} \cmidrule(lr){8-10}
& & &
AbsRel &
SqRel &
RMSE &
RMSE$_{log}$ &
$\delta < {1.25}$ &
$\delta < {1.25}^2$ &
$\delta < {1.25}^3$
\\
\midrule
Kuznietsov \emph{et al.} \cite{kuznietsov2017semi}
& D & 
& 0.113 & 0.741 & 4.621 & 0.189 & 0.862 & 0.960 & 0.986 
\\
Gan \emph{et al.} \cite{Gan2018MonocularDE}
& D &
& 0.098 & 0.666 & 3.933 & 0.173 & 0.890 & 0.964 & 0.985 
\\
Guizilini \emph{et al.} \cite{packnet-semisup}
& D &
& 0.072 & 0.340 & 3.265 & 0.116 & 0.934 & --- & --- 
\\
DORN \cite{fu2018deep} 
& D &
& 0.072 & 0.307 & \underline{2.727} & 0.120 & 0.932 & 0.984 & 0.994 
\\
Yin \emph{et al.} \cite{Yin2019enforcing}
& D &
& 0.072 & --- & 3.258 & 0.117 & 0.938 & 0.990 & \textbf{0.998} 
\\
PackNet-SFM \cite{packnet}
& M &
& 0.078 & 0.420 & 3.485 & 0.121 & 0.931 & 0.986 & 0.996 
\\
ManyDepth \cite{manydepth} 
& M & \checkmark
& 0.064 & 0.320 & 3.187 & 0.104 & 0.946 & 0.990 & 0.995 
\\
BTS \cite{lee2019big} 
& D &
& \underline{0.059} & \textbf{0.245} & {2.756} & {0.096} & \underline{0.956} & \textbf{0.993} & \textbf{0.998} 
\\
\midrule
\textbf{DepthFormer (MR)} 
& M & \checkmark
& \textbf{0.055} & {0.271} & 2.917 & \underline{0.095} & {0.955} & {0.991} & \textbf{0.998} 
\\
\textbf{DepthFormer (HR)} 
& M & \checkmark
& \textbf{0.055} & \underline{0.265} & \textbf{2.723} & \textbf{0.092} & \textbf{0.959} & \underline{0.992} & \textbf{0.998} 
\\
\bottomrule
\end{tabular}
}
\caption{
\textbf{Depth results} on the KITTI \emph{Eigen} test split \cite{eigen2014depth}, for distances up to 80m with the \emph{Garg} crop~\cite{garg2016unsupervised}, evaluated on the improved depth maps from \cite{gtkitti}. \emph{Superv.} indicates the source of supervision (M for monocular self-supervision and D for depth supervision); and \emph{Multi-Fr.} the use of multiple frames at test time. Monocular results are median-scaled at test time, to account for scale ambiguity. We report DepthFormer results in both half-resolution (MR, $640 \times 192$), and full resolution (HR, $1216 \times 352$), using the same training and architecture parameters (Section 4.2, main paper).
}
\label{table:kitti_depth_sup}
\end{table*}



\section{Network Details}

Below we describe each network used in our proposed DepthFormer architecture, and Table \ref{tab:networks} shows detailed diagrams for each of them. Note that our contributions do not require any network architecture in particular, and can be extended to incorporate recent developments for potential further improvements in performance\cite{packnet,shu2020featdepth,liu2021swin}. Open-source training and inference code, as well as pre-trained models, will be made available upon publication.

\subsection{Cross-Attention Network}

Similar to \cite{sttr}, we use an hourglass-shaped architecture as the encoder, modified with residual connections and spatial pyramid pooling modules~\cite{chang2018pyramid}. The decoder consists of transposed convolutions, dense-blocks~\cite{huang2017densely}, and a final convolution layer. The final feature map has the same spatial resolution as the input image, encoding both local and global contexts. This feature map is then downsampled to the cost volume resolution using bilinear interpolation.

\subsection{Single-Frame Depth Network}

We use a ResNet18 backbone~\cite{monodepth2} as the single-frame encoder, followed by a decoder that outputs multi-scale depth maps at four different resolutions: one-eighth, one-fourth, one-half, and the original input dimension. Following \cite{manydepth}, we concatenate the $H/4 \times W/4 \times 128$ encoded features with the $H/4 \times W/4 \times D$ multi-frame cost volume.  A bottleneck convolutional layer, with kernel size $3$, is then used to combine these two sources of features (single-frame and multi-frame) into a $H/4 \times W/4 \times 128$ feature map for further encoding and decoding (Figure 4, main paper).

\subsection{Context Adjustment Network}

The input to our context adjustment network is a $H/4 \times W/4 \times 4$ tensor created by concatenating the normalized high-response depth map $\tilde{D}_H$ and the target image $I_t$. Depth map normalization is done as such:
\begin{equation}
\small
\tilde{D}_H = \left( \hat{D}_H - mean(\hat{D}_H) \right) \Big/ std(\hat{D}_H)
\label{eq:norm}
\end{equation}
This normalized high-response depth map is refined through a series of residual blocks that expand the channel dimensions, before a ReLU activation restores it to the original shape. The high-response depth map is concatenated with the output of each residual block, and added to the final output using a long skip connection. This final output is then un-normalized using the original statistics, generating a context-adjusted predicted depth map $\hat{D}_C$:
\begin{equation}
\small
\hat{D}_{C} = \left(\tilde{D}_H + \theta_{C}(I_t,\tilde{D}_H)\right) std(\hat{D}_H) + mean(\hat{D}_H)
\label{eq:unnorm}
\end{equation}


\subsection{Pose Network}

Our pose network uses a ResNet18 backbone, modified to accommodate two input images by duplicating the convolutional weights of the first layer~\cite{monodepth2}. The bottleneck feature maps are further processed using a series of convolutional layers, with the last one outputting a $H/32 \times H/32 \times 6$ feature map. This feature map is then averaged over the spatial dimensions, generating a 6-dimensional vector containing the relative translation and rotation between frames, in Euler angles. Following \cite{manydepth}, we invert the order of input images when predicting backwards motion. 

\section{Comparison to Supervised Methods}

Our DepthFormer architecture was designed for self-supervised learning, in which training is conducted without explicit supervision from ground-truth depth maps. As mentioned in the main paper (Section 2.1), this is a very challenging setting, due to limitations of the photometric objective in the presence of dynamic objects, static frames, changes in luminosity, and so forth. Even so, our contributions in multi-frame feature matching lead to a depth estimation performance that surpasses even current state-of-the-art single-frame supervised depth estimation methods. These results are summarized in Table \ref{table:kitti_depth_sup}. More specifically, we achieve comparable performance to BTS~\cite{fu2018deep} when training and evaluating at half resolution ($640 \times 192$), and surpass it in almost all metrics when training and evaluating at the same full resolution ($1216 \times 352$). We believe the introduction of other self-supervised depth network architectures more suitable for high resolution processing~\cite{packnet} should lead to further improvements, however a more thorough exploration is left to future work. 

\section{Qualitative Examples}

\begin{figure*}[t!]
    \centering
    \subfloat{
    \includegraphics[width=0.24\textwidth,height=1.55cm]{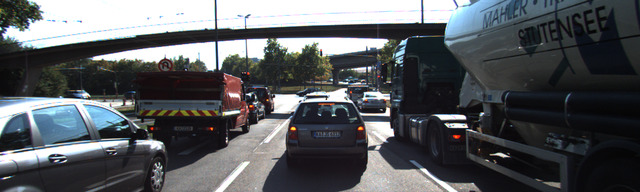}}
    \subfloat{
    \includegraphics[width=0.24\textwidth,height=1.55cm]{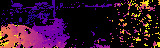}}
    \subfloat{
    \includegraphics[width=0.24\textwidth,height=1.55cm]{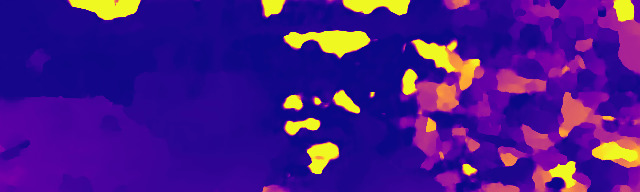}}
    \subfloat{
    \includegraphics[width=0.24\textwidth,height=1.55cm]{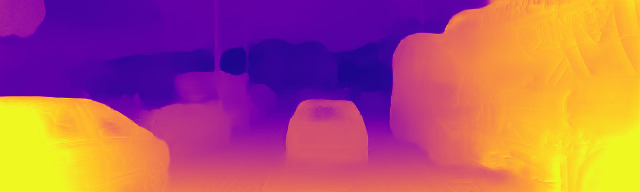}}
    \\
    \subfloat{
    \includegraphics[width=0.24\textwidth,height=1.55cm]{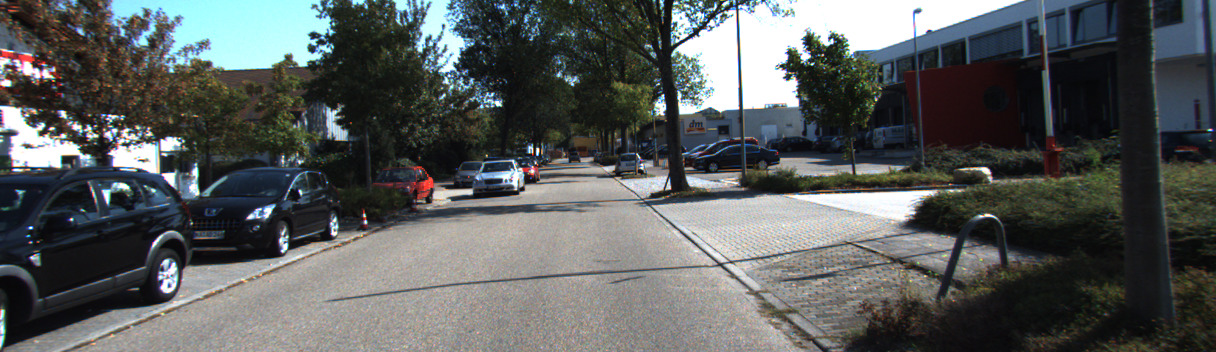}}
    \subfloat{
    \includegraphics[width=0.24\textwidth,height=1.55cm]{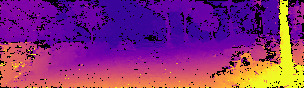}}
    \subfloat{
    \includegraphics[width=0.24\textwidth,height=1.55cm]{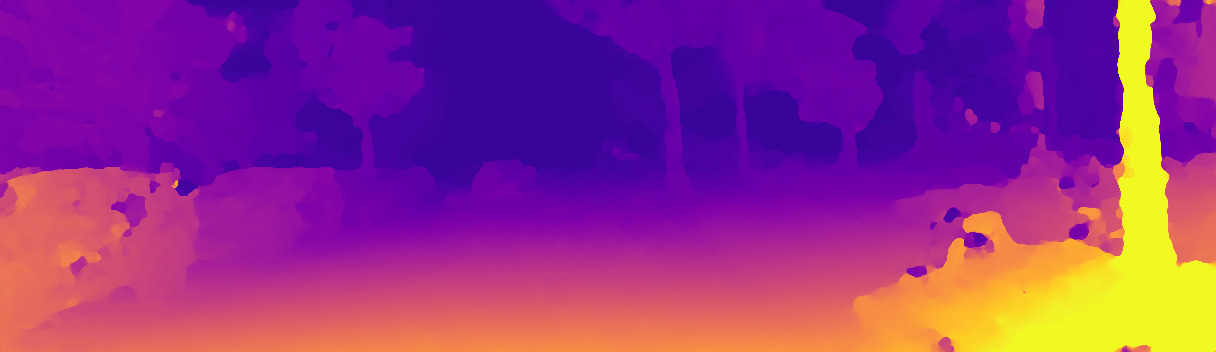}}
    \subfloat{
    \includegraphics[width=0.24\textwidth,height=1.55cm]{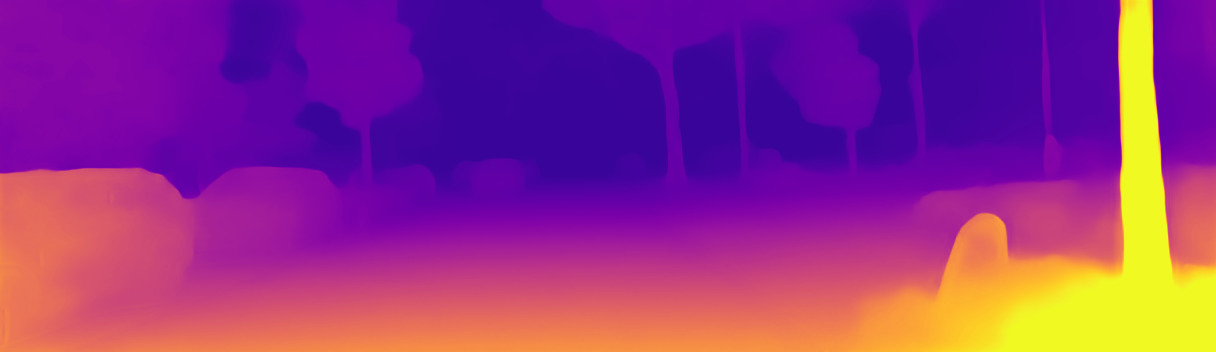}}
    \\
    \subfloat{
    \includegraphics[width=0.24\textwidth,height=1.55cm]{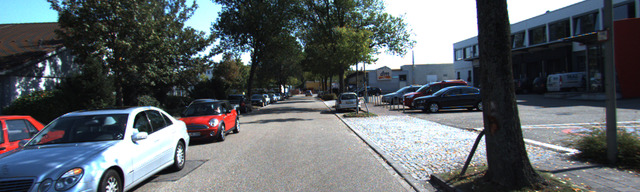}}
    \subfloat{
    \includegraphics[width=0.24\textwidth,height=1.55cm]{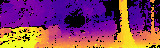}}
    \subfloat{
    \includegraphics[width=0.24\textwidth,height=1.55cm]{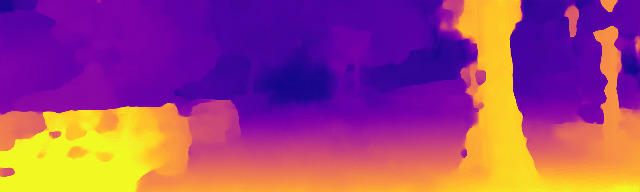}}
    \subfloat{
    \includegraphics[width=0.24\textwidth,height=1.55cm]{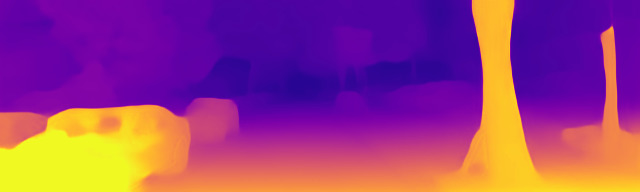}}
    \\
    \subfloat{
    \includegraphics[width=0.24\textwidth,height=1.55cm]{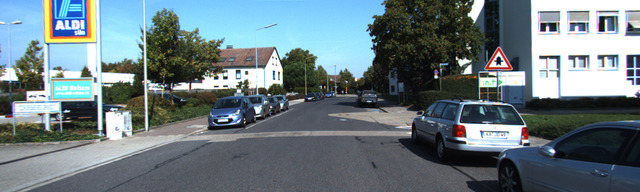}}
    \subfloat{
    \includegraphics[width=0.24\textwidth,height=1.55cm]{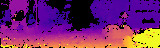}}
    \subfloat{
    \includegraphics[width=0.24\textwidth,height=1.55cm]{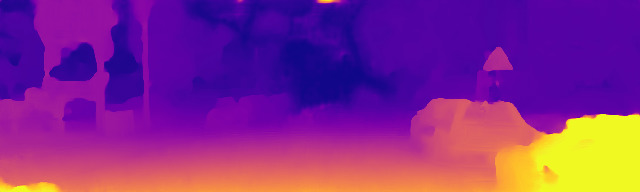}}
    \subfloat{
    \includegraphics[width=0.24\textwidth,height=1.55cm]{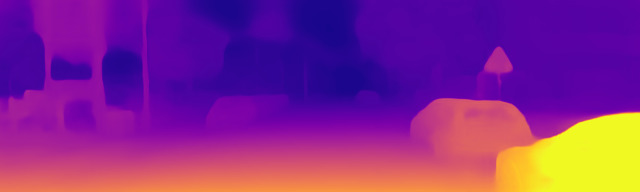}}
    \\
    \setcounter{subfigure}{0}
    \subfloat{
    \includegraphics[width=0.24\textwidth,height=1.55cm]{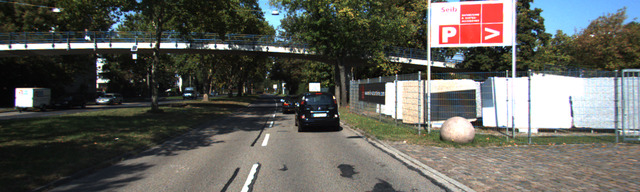}}
    \subfloat{
    \includegraphics[width=0.24\textwidth,height=1.55cm]{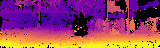}}
    \subfloat{
    \includegraphics[width=0.24\textwidth,height=1.55cm]{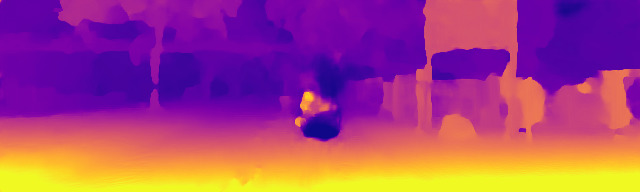}}
    \subfloat{
    \includegraphics[width=0.24\textwidth,height=1.55cm]{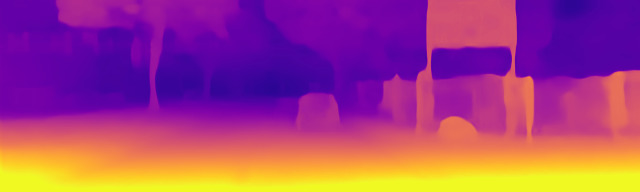}}
    \\
    \subfloat{
    \includegraphics[width=0.24\textwidth,height=1.55cm]{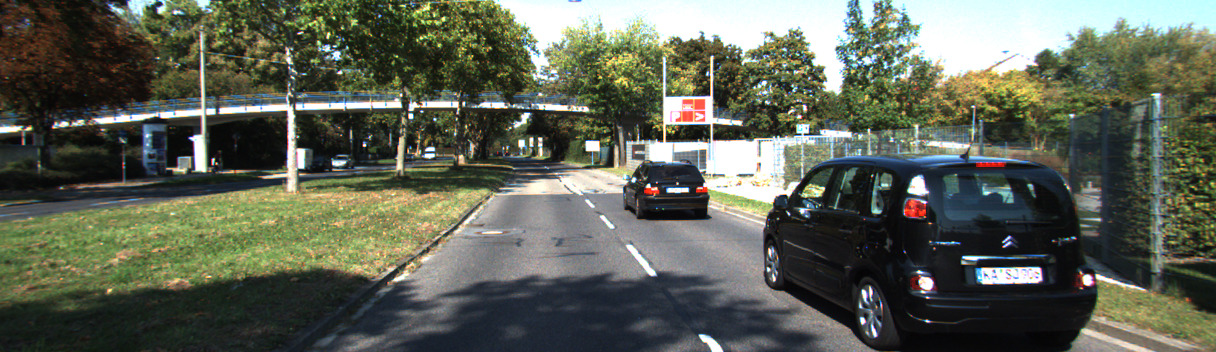}}
    \subfloat{
    \includegraphics[width=0.24\textwidth,height=1.55cm]{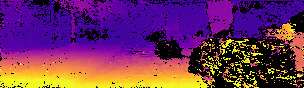}}
    \subfloat{
    \includegraphics[width=0.24\textwidth,height=1.55cm]{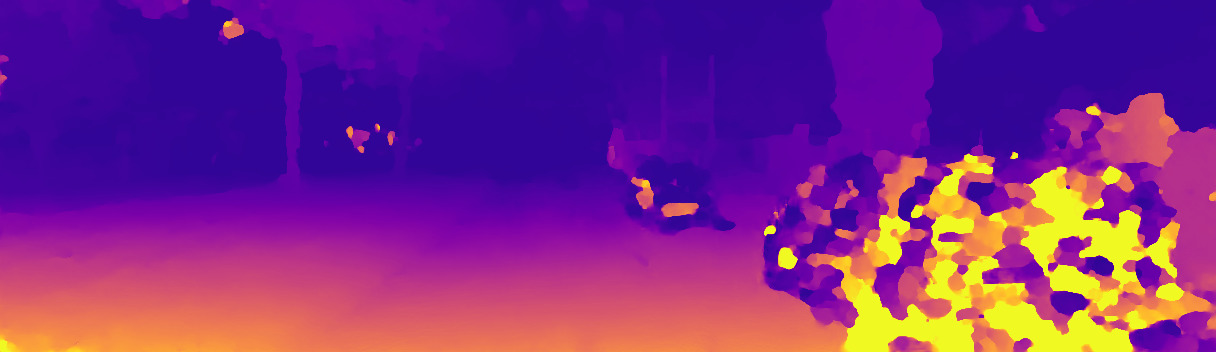}}
    \subfloat{
    \includegraphics[width=0.24\textwidth,height=1.55cm]{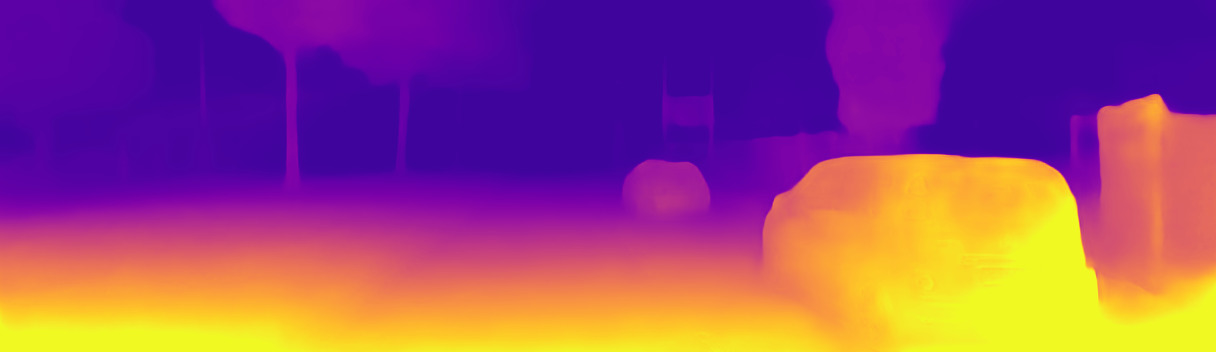}}
    \\    
    \subfloat{
    \includegraphics[width=0.24\textwidth,height=1.55cm]{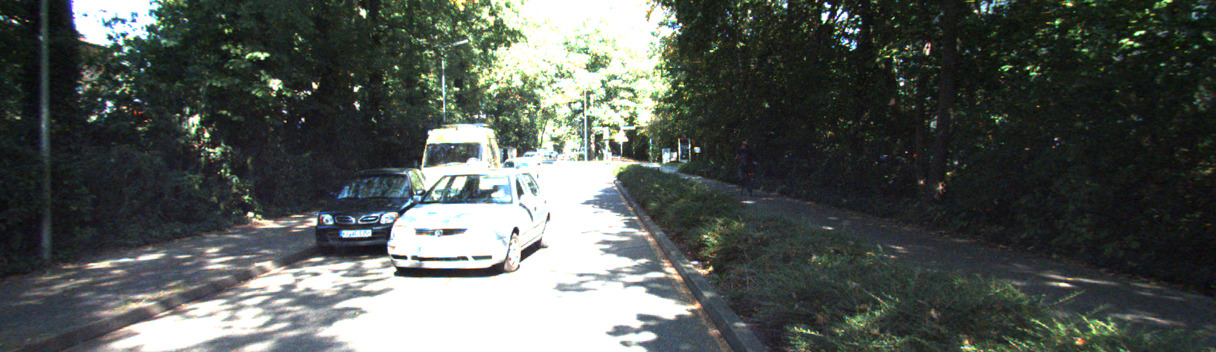}}
    \subfloat{
    \includegraphics[width=0.24\textwidth,height=1.55cm]{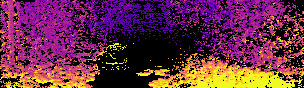}}
    \subfloat{
    \includegraphics[width=0.24\textwidth,height=1.55cm]{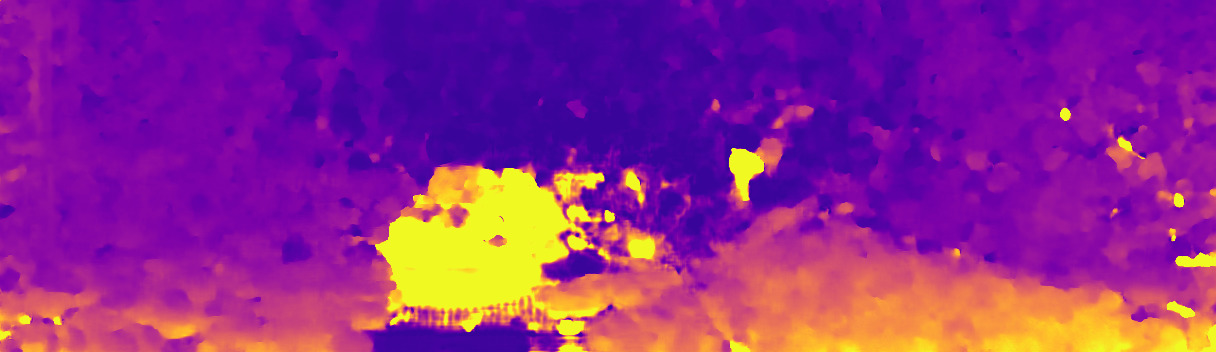}}
    \subfloat{
    \includegraphics[width=0.24\textwidth,height=1.55cm]{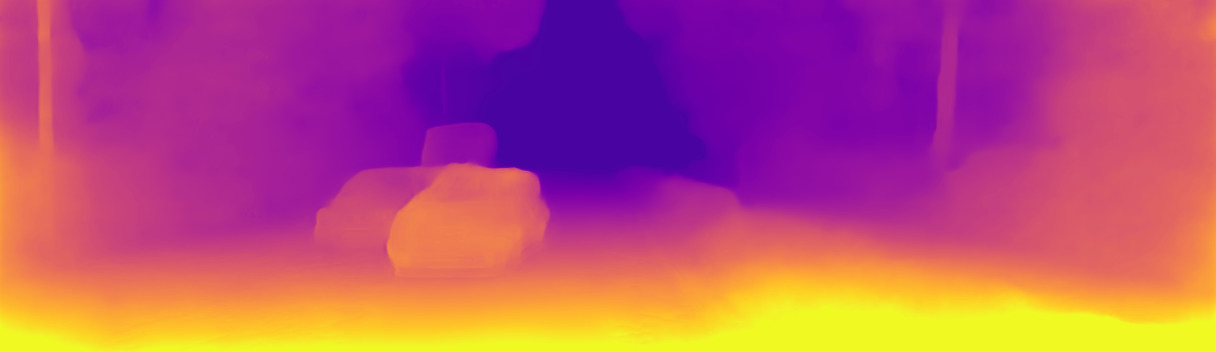}}
    \\    
    \setcounter{subfigure}{0}
    \subfloat{
    \includegraphics[width=0.24\textwidth,height=1.55cm]{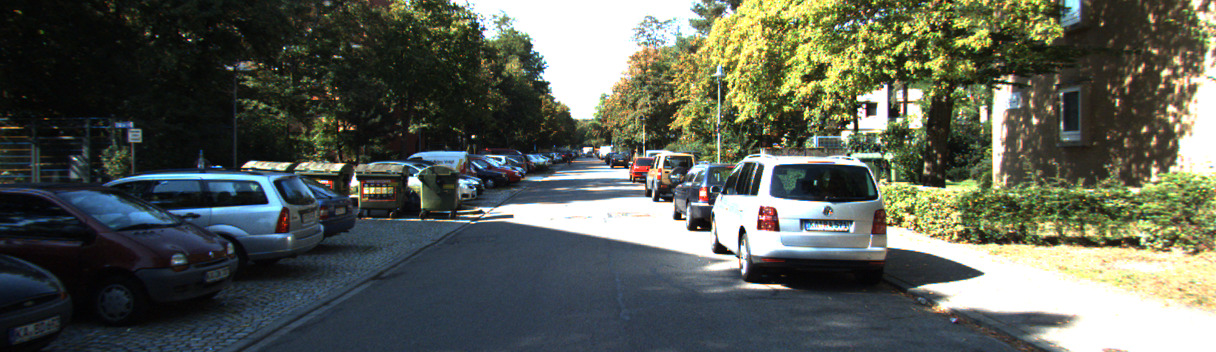}}
    \subfloat{
    \includegraphics[width=0.24\textwidth,height=1.55cm]{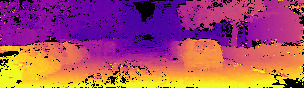}}
    \subfloat{
    \includegraphics[width=0.24\textwidth,height=1.55cm]{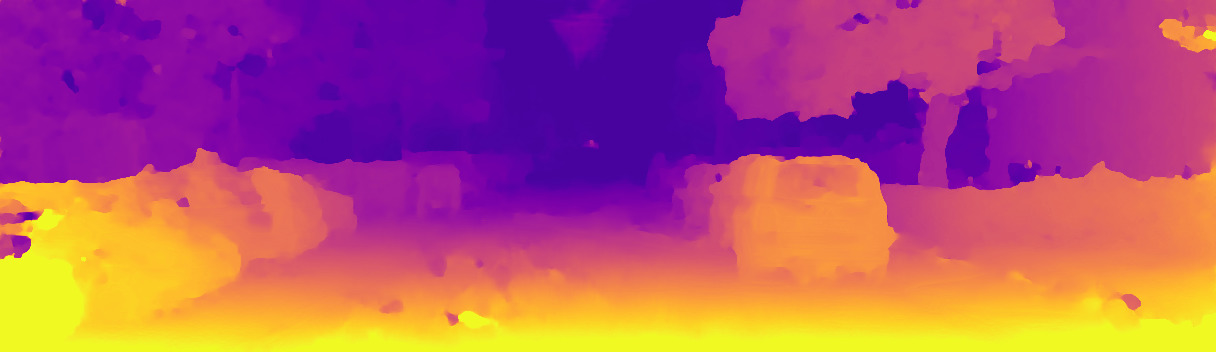}}
    \subfloat{
    \includegraphics[width=0.24\textwidth,height=1.55cm]{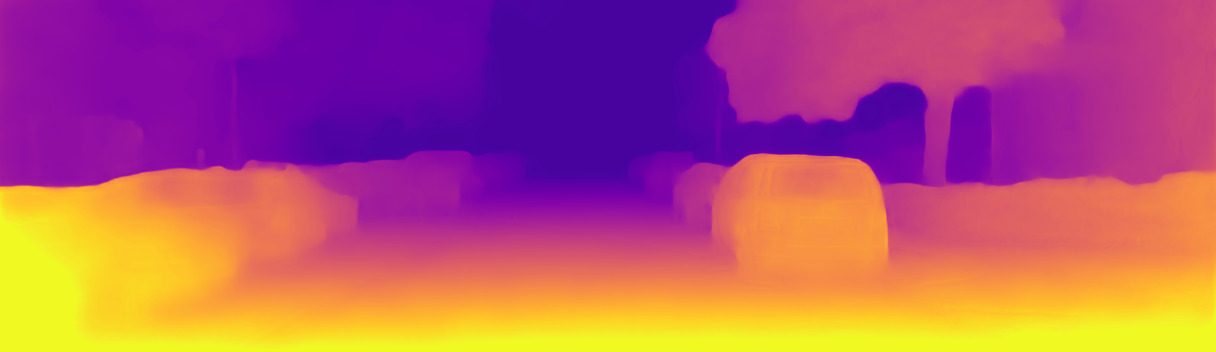}}
    \\
    \subfloat{
    \includegraphics[width=0.24\textwidth,height=1.55cm]{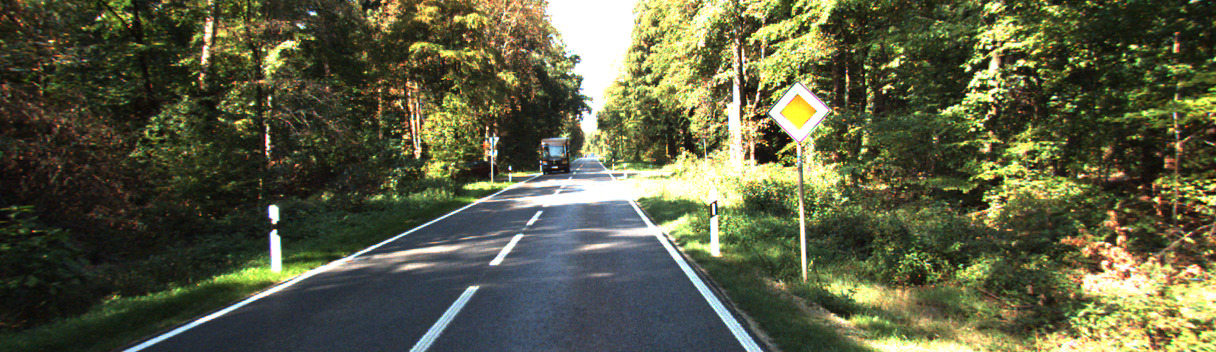}}
    \subfloat{
    \includegraphics[width=0.24\textwidth,height=1.55cm]{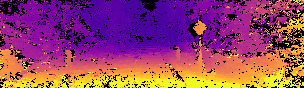}}
    \subfloat{
    \includegraphics[width=0.24\textwidth,height=1.55cm]{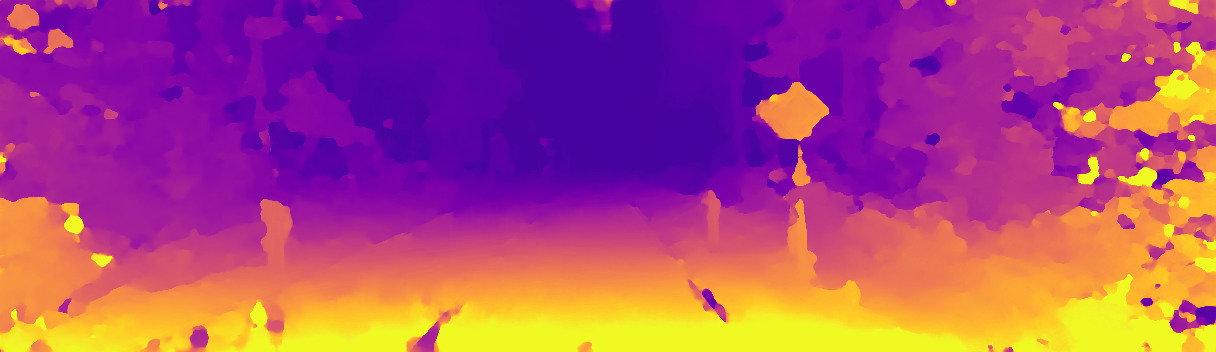}}
    \subfloat{
    \includegraphics[width=0.24\textwidth,height=1.55cm]{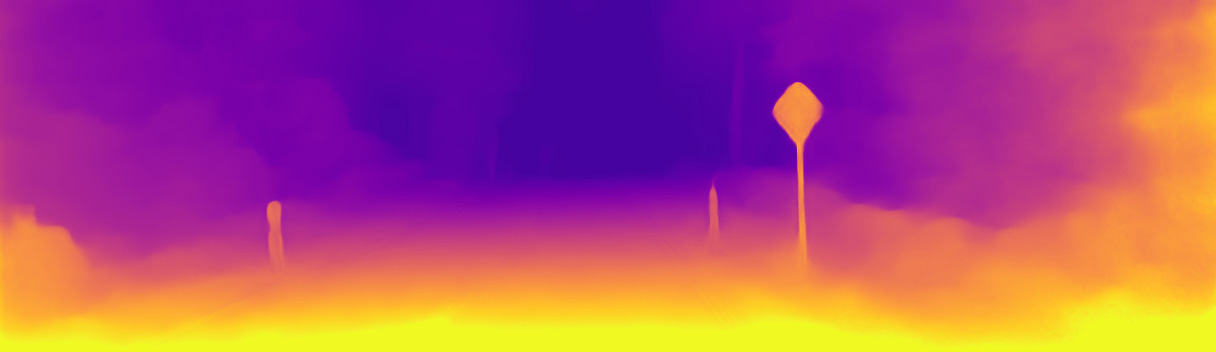}}
    \\
    \subfloat{
    \includegraphics[width=0.24\textwidth,height=1.55cm]{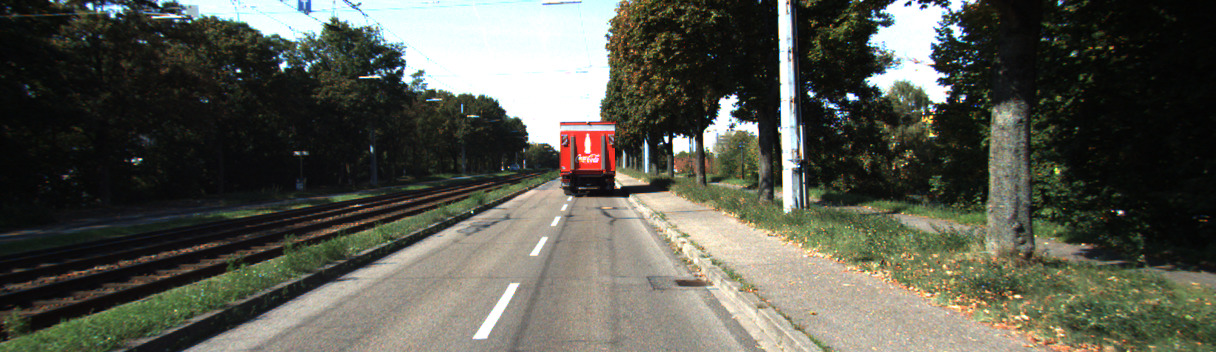}}
    \subfloat{
    \includegraphics[width=0.24\textwidth,height=1.55cm]{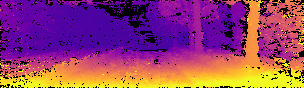}}
    \subfloat{
    \includegraphics[width=0.24\textwidth,height=1.55cm]{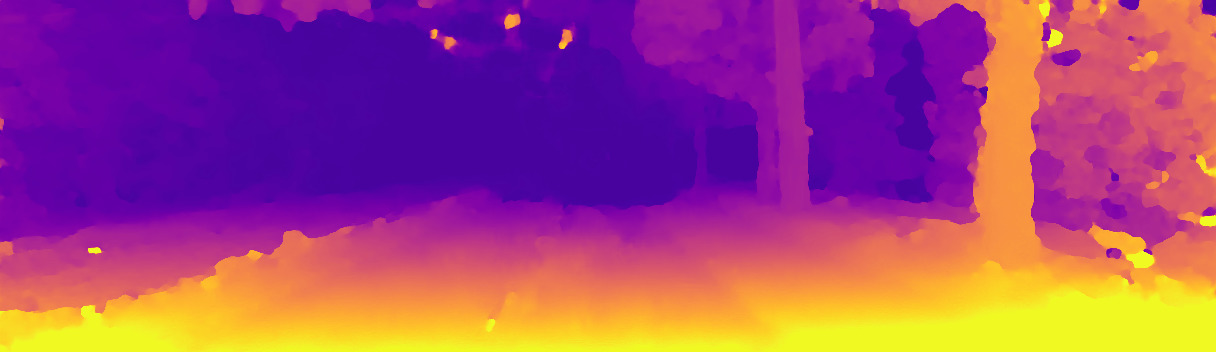}}
    \subfloat{
    \includegraphics[width=0.24\textwidth,height=1.55cm]{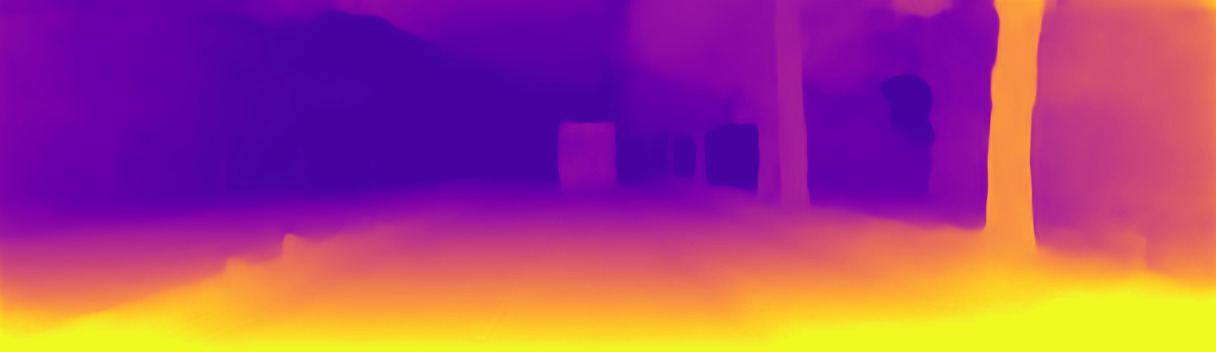}}
    \\       
    \subfloat{
    \includegraphics[width=0.24\textwidth,height=1.55cm]{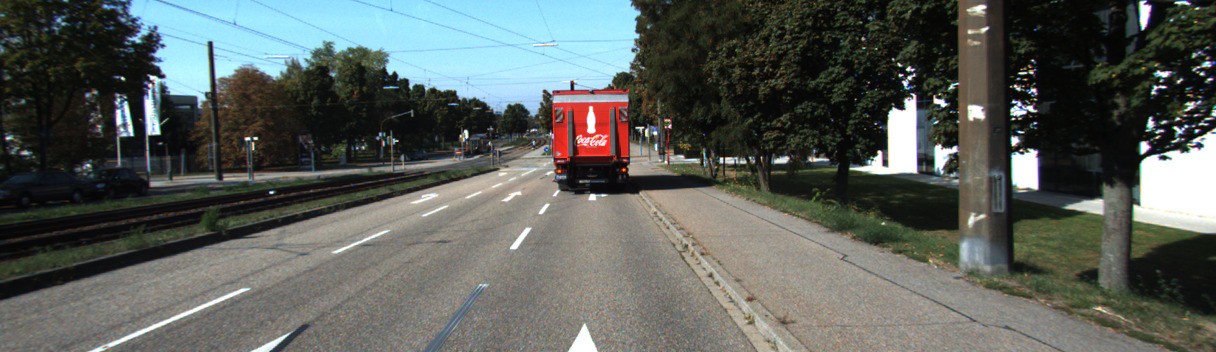}}
    \subfloat{
    \includegraphics[width=0.24\textwidth,height=1.55cm]{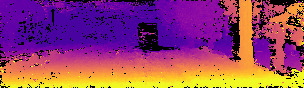}}
    \subfloat{
    \includegraphics[width=0.24\textwidth,height=1.55cm]{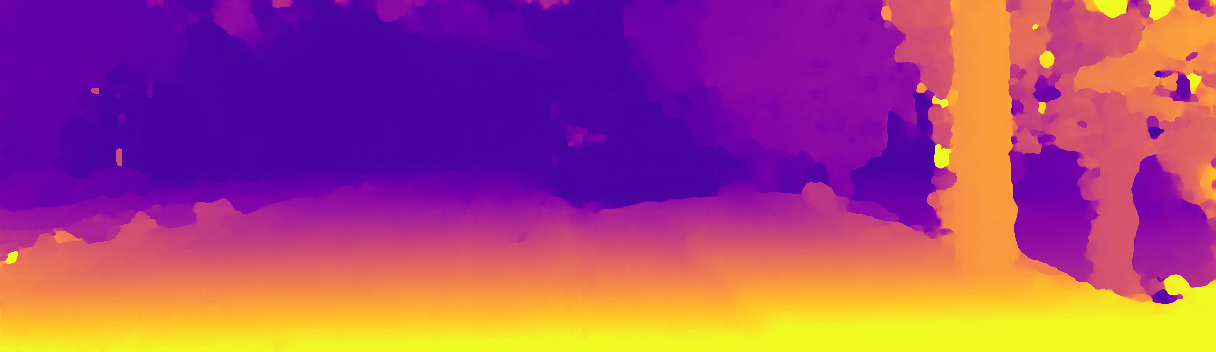}}
    \subfloat{
    \includegraphics[width=0.24\textwidth,height=1.55cm]{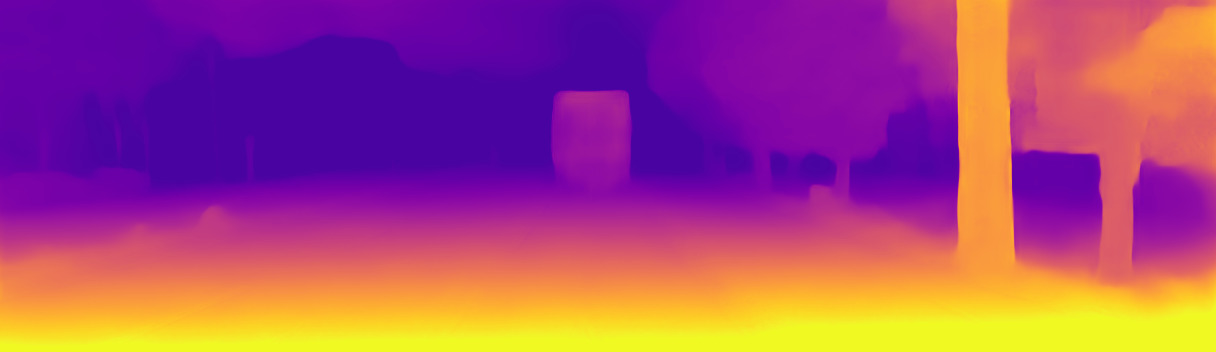}}
    \\       
    \subfloat{
    \includegraphics[width=0.24\textwidth,height=1.55cm]{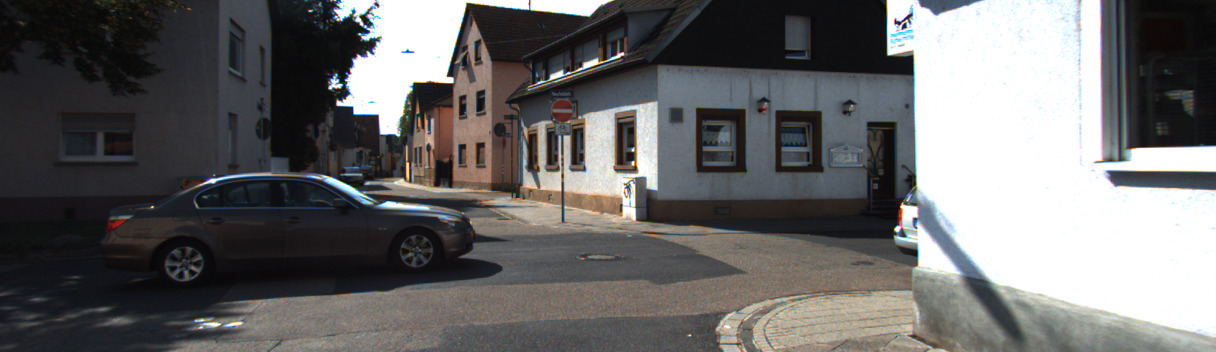}}
    \subfloat{
    \includegraphics[width=0.24\textwidth,height=1.55cm]{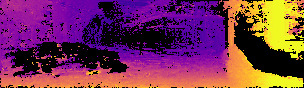}}
    \subfloat{
    \includegraphics[width=0.24\textwidth,height=1.55cm]{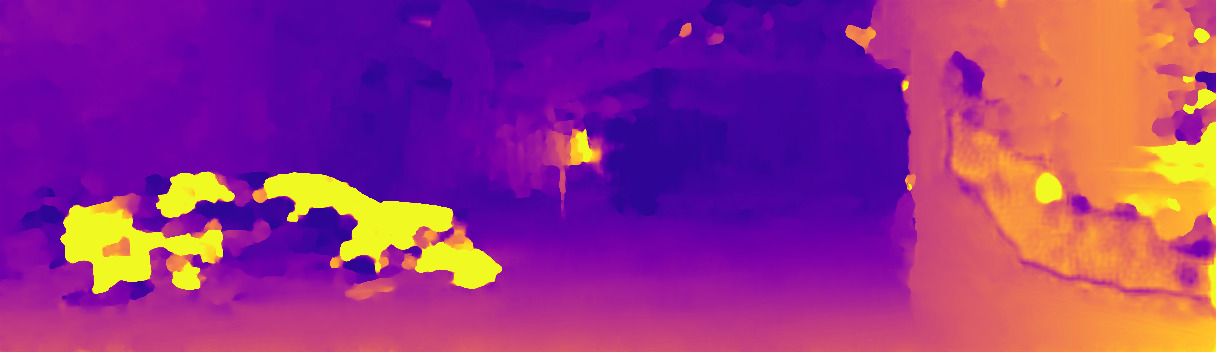}}
    \subfloat{
    \includegraphics[width=0.24\textwidth,height=1.55cm]{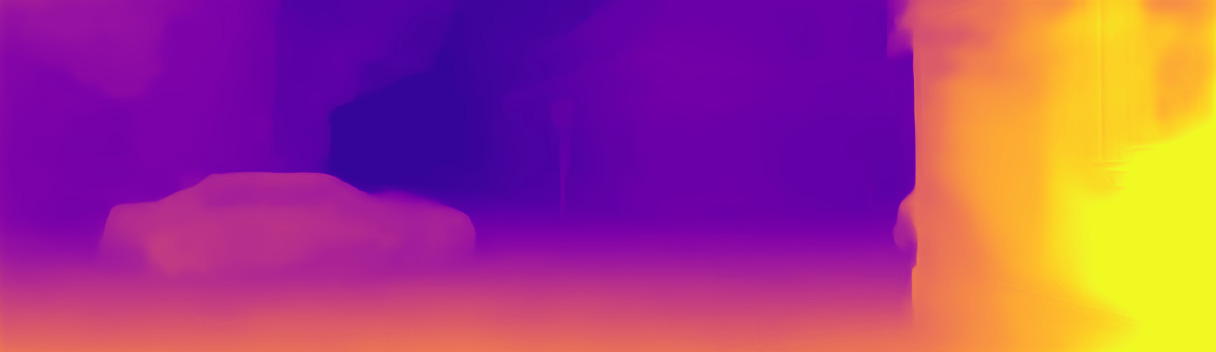}}
    \\
    \setcounter{subfigure}{0}
    \subfloat[Target image]{
    \includegraphics[width=0.24\textwidth]{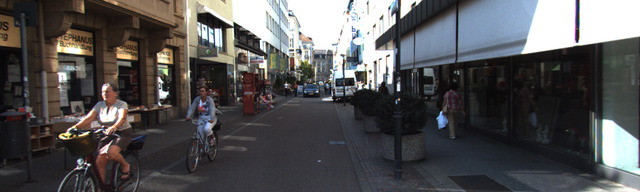}}
    \subfloat[High-response depth]{
    \includegraphics[width=0.24\textwidth]{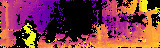}}
    \subfloat[Context-adjusted depth]{
    \includegraphics[width=0.24\textwidth]{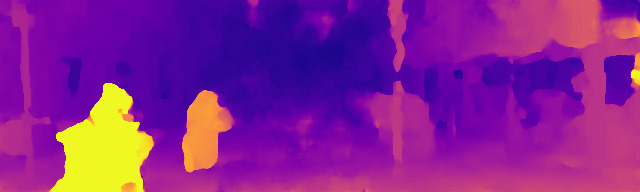}}
    \subfloat[Decoded depth]{
    \includegraphics[width=0.24\textwidth]{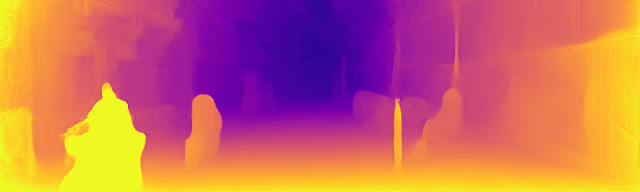}}
    \caption{\textbf{Qualitative depth estimation results} of our proposed DepthFormer architecture, on the KITTI dataset.}
    \label{fig:failure}
\end{figure*}

\begin{figure*}[t!]
    \centering
    \subfloat{
    \includegraphics[width=0.24\textwidth]{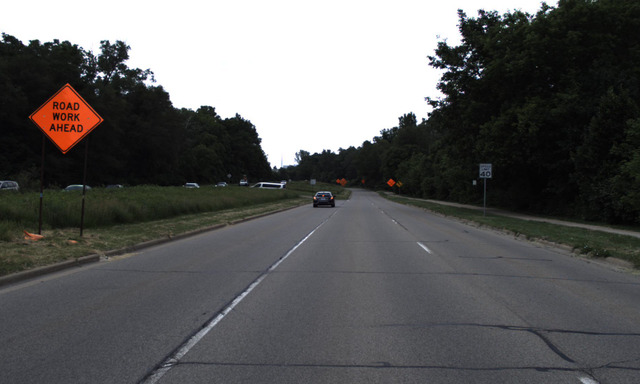}}
    \subfloat{
    \includegraphics[width=0.24\textwidth]{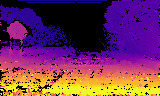}}
    \subfloat{
    \includegraphics[width=0.24\textwidth]{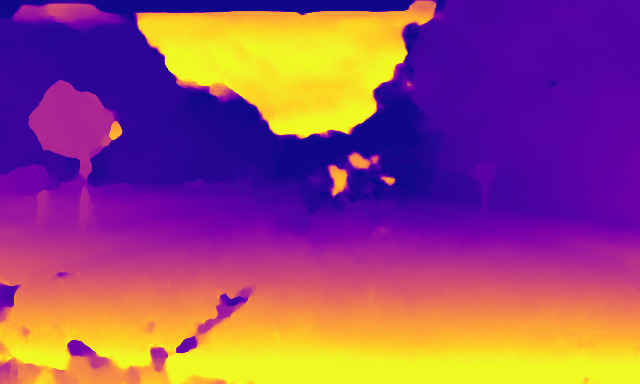}}
    \subfloat{
    \includegraphics[width=0.24\textwidth]{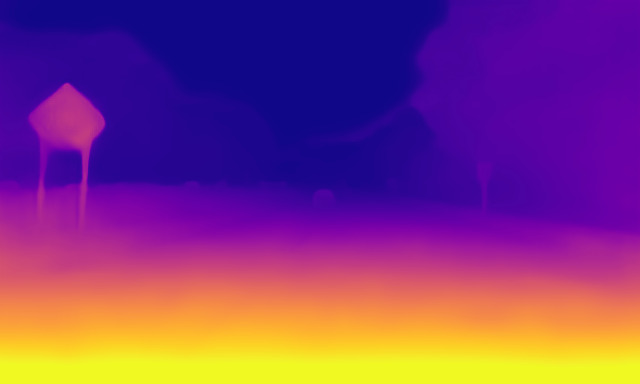}}
    \\
    \subfloat{
    \includegraphics[width=0.24\textwidth]{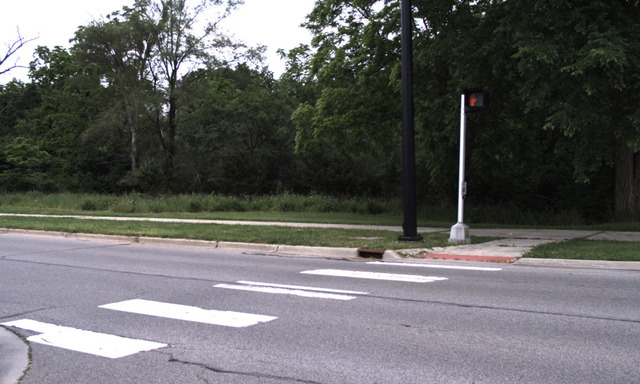}}
    \subfloat{
    \includegraphics[width=0.24\textwidth]{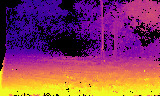}}
    \subfloat{
    \includegraphics[width=0.24\textwidth]{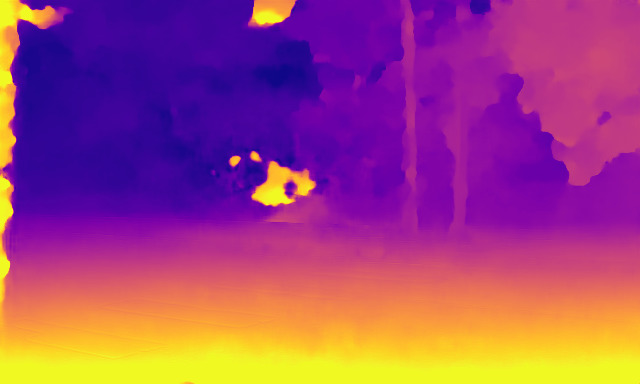}}
    \subfloat{
    \includegraphics[width=0.24\textwidth]{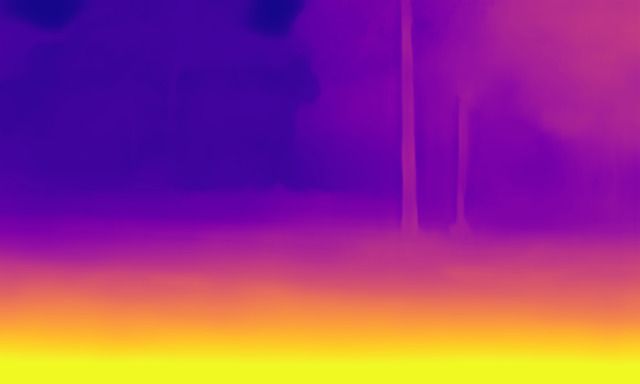}}
    \\    
    \subfloat{
    \includegraphics[width=0.24\textwidth]{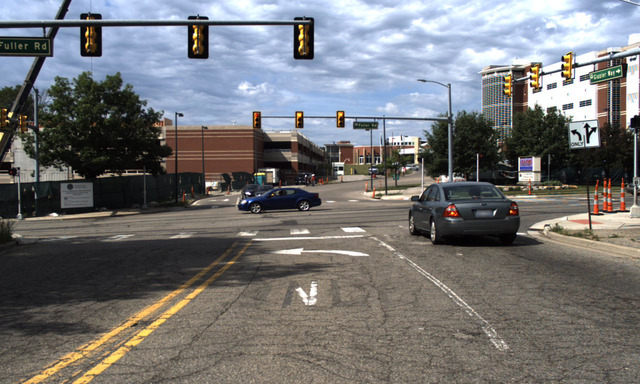}}
    \subfloat{
    \includegraphics[width=0.24\textwidth]{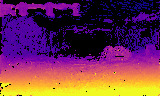}}
    \subfloat{
    \includegraphics[width=0.24\textwidth]{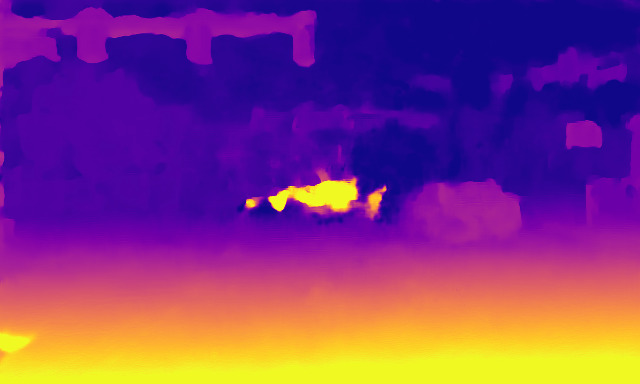}}
    \subfloat{
    \includegraphics[width=0.24\textwidth]{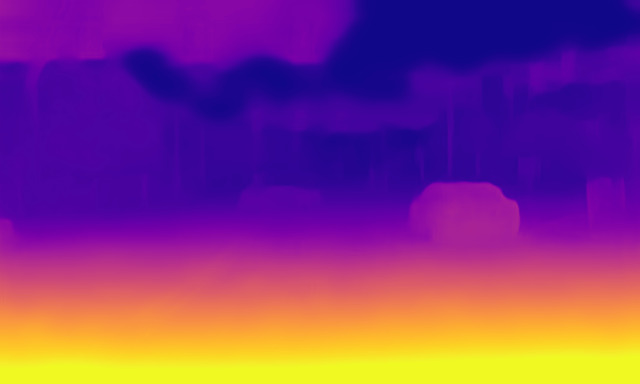}}
    \\    
    \subfloat{
    \includegraphics[width=0.24\textwidth]{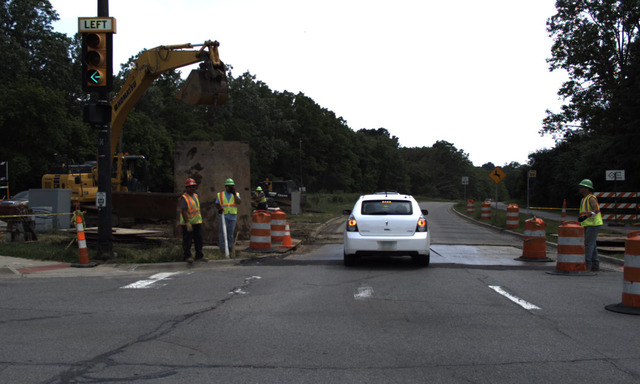}}
    \subfloat{
    \includegraphics[width=0.24\textwidth]{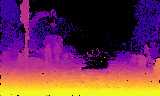}}
    \subfloat{
    \includegraphics[width=0.24\textwidth]{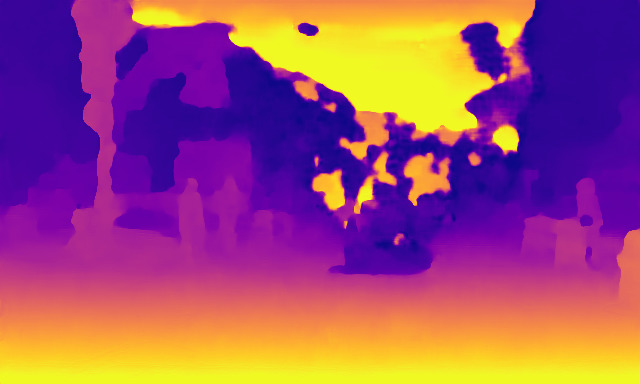}}
    \subfloat{
    \includegraphics[width=0.24\textwidth]{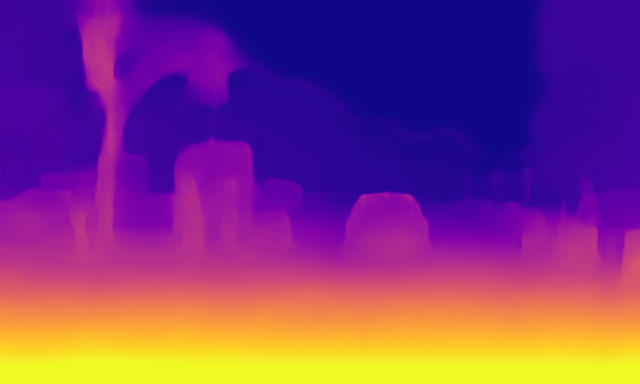}}
    \\    
    \subfloat{
    \includegraphics[width=0.24\textwidth]{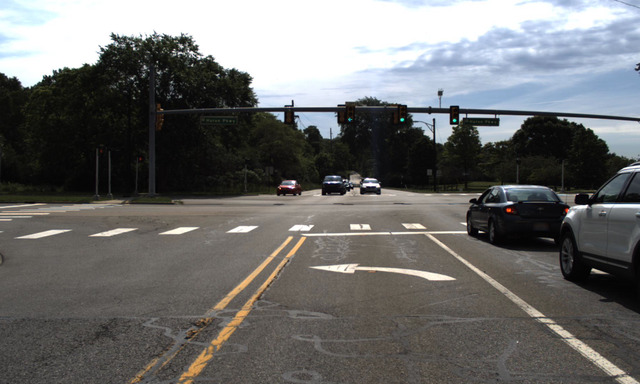}}
    \subfloat{
    \includegraphics[width=0.24\textwidth]{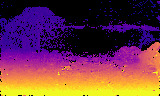}}
    \subfloat{
    \includegraphics[width=0.24\textwidth]{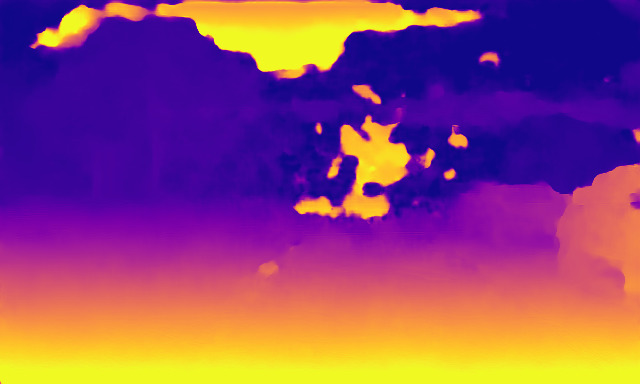}}
    \subfloat{
    \includegraphics[width=0.24\textwidth]{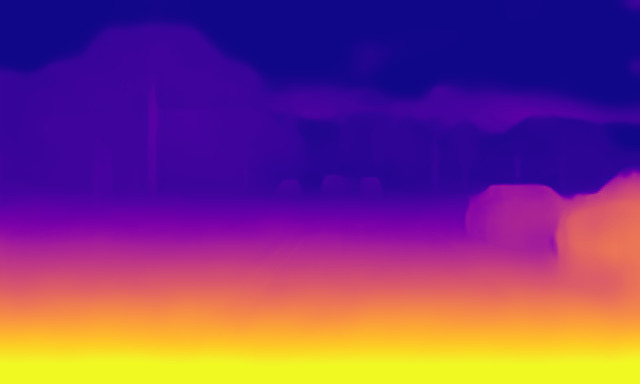}}
    \\  
    \setcounter{subfigure}{0}
    \subfloat{
    \includegraphics[width=0.24\textwidth]{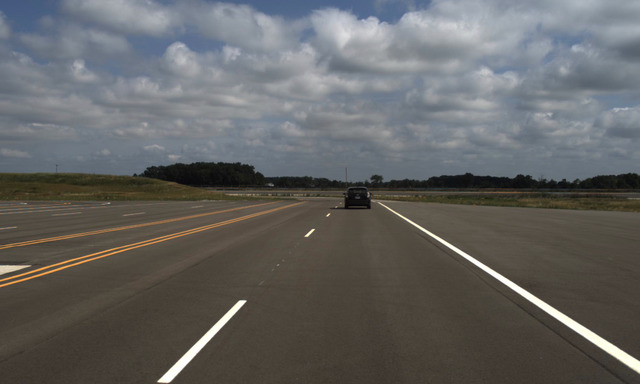}}
    \subfloat{
    \includegraphics[width=0.24\textwidth]{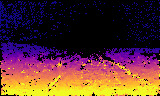}}
    \subfloat{
    \includegraphics[width=0.24\textwidth]{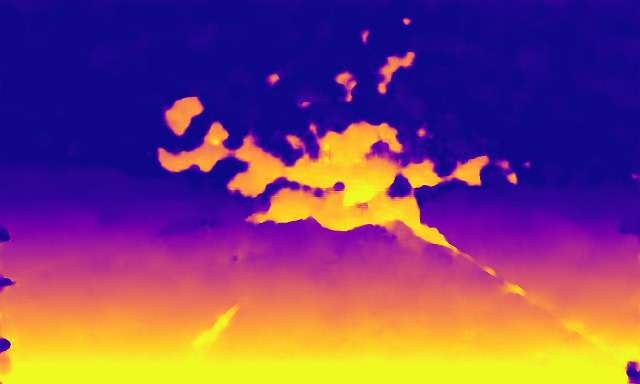}}
    \subfloat{
    \includegraphics[width=0.24\textwidth]{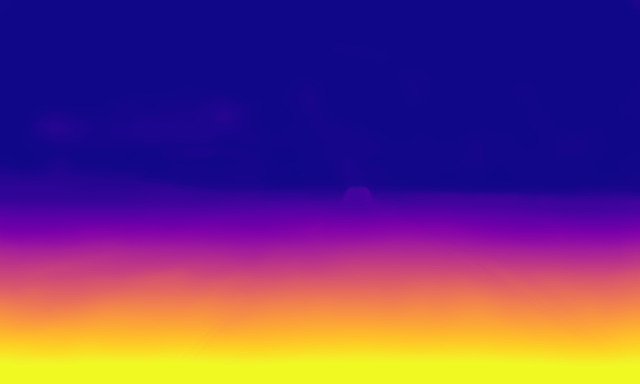}}
    \\    
    \subfloat{
    \includegraphics[width=0.24\textwidth]{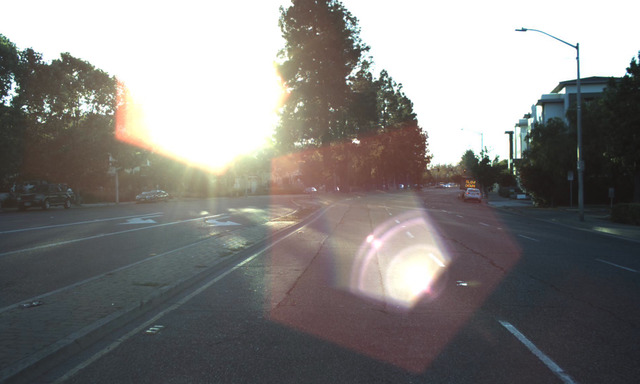}}
    \subfloat{
    \includegraphics[width=0.24\textwidth]{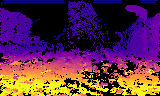}}
    \subfloat{
    \includegraphics[width=0.24\textwidth]{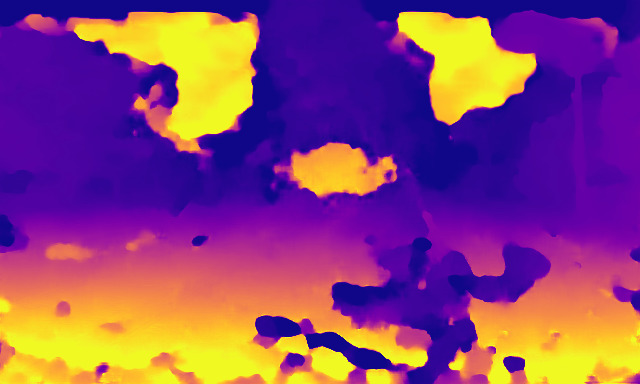}}
    \subfloat{
    \includegraphics[width=0.24\textwidth]{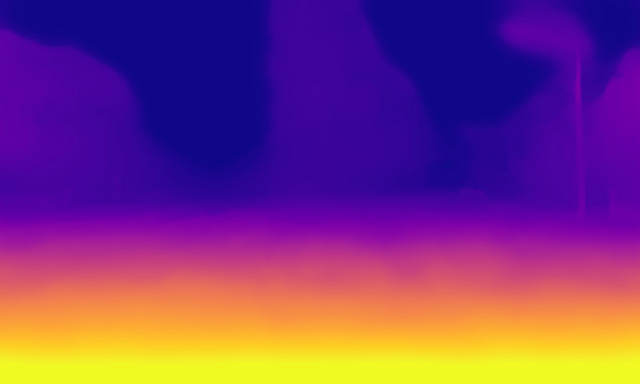}}
    \\            \setcounter{subfigure}{0}
    \subfloat[Target image]{
    \includegraphics[width=0.24\textwidth]{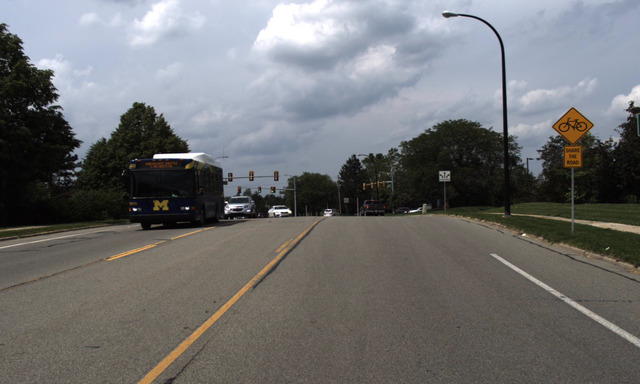}}
    \subfloat[High-response depth]{
    \includegraphics[width=0.24\textwidth]{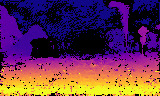}}
    \subfloat[Context-adjusted depth]{
    \includegraphics[width=0.24\textwidth]{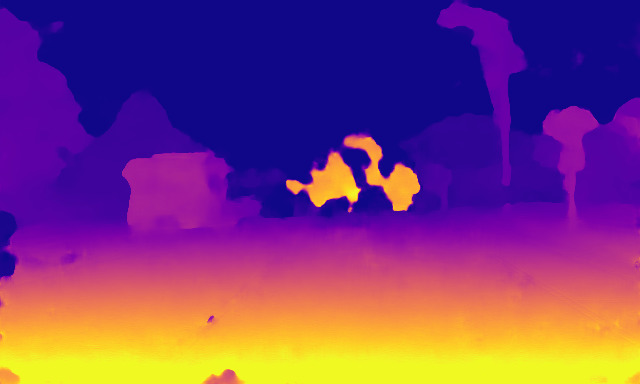}}
    \subfloat[Decoded depth]{
    \includegraphics[width=0.24\textwidth]{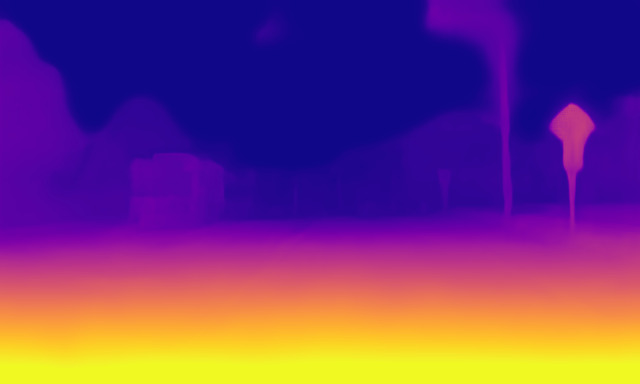}}
    \caption{\textbf{Qualitative depth estimation results} of our proposed DepthFormer architecture, on the DDAD dataset.}
    \label{fig:failure_ddad}
\end{figure*}

Some examples of predicted depth maps, including common failure cases due to lack of camera motion and dynamic objects, are shown in Figures \ref{fig:failure} and \ref{fig:failure_ddad} for the KITTI and DDAD datasets respectively. High-response depth maps (Section 3.3.1, main paper) are masked out using our proposed low-confidence threshold. These masked out regions usually include far-away objects towards the vanishing point, including the sky, and interestingly also occluded areas and dynamic objects. Context-adjusted depth maps (Section 3.3.2, main paper) are able to reason over these low-confidence areas by conditioning with information from the target image. However, they still fail in situations where multi-frame matching is inaccurate or ill-posed (e.g., lack of camera motion or dynamic objects). By introducing single-frame features for joint decoding (Section 3.3.3, main paper), we are able to also reason over these situations and achieve our reported state-of-the-art results. Quantitative evaluation of these intermediate depth maps is provided in Table 2 of the main paper. 

\section{Reconstructed Pointclouds}

\begin{figure*}[t!]
    \centering
    \subfloat{
        \includegraphics[width=0.49\textwidth,height=6.8cm]{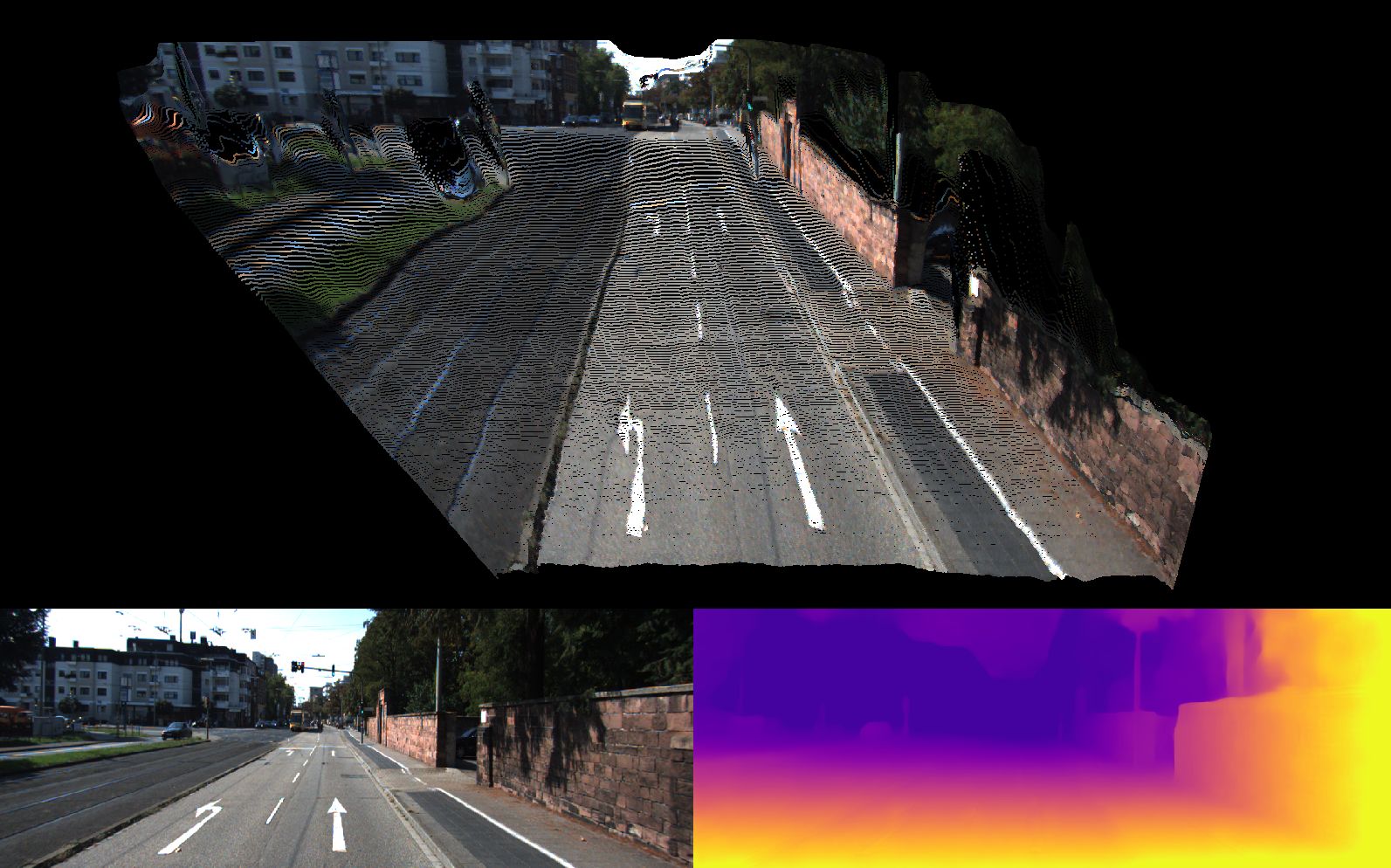}
    }
    \subfloat{
        \includegraphics[width=0.49\textwidth,height=6.8cm]{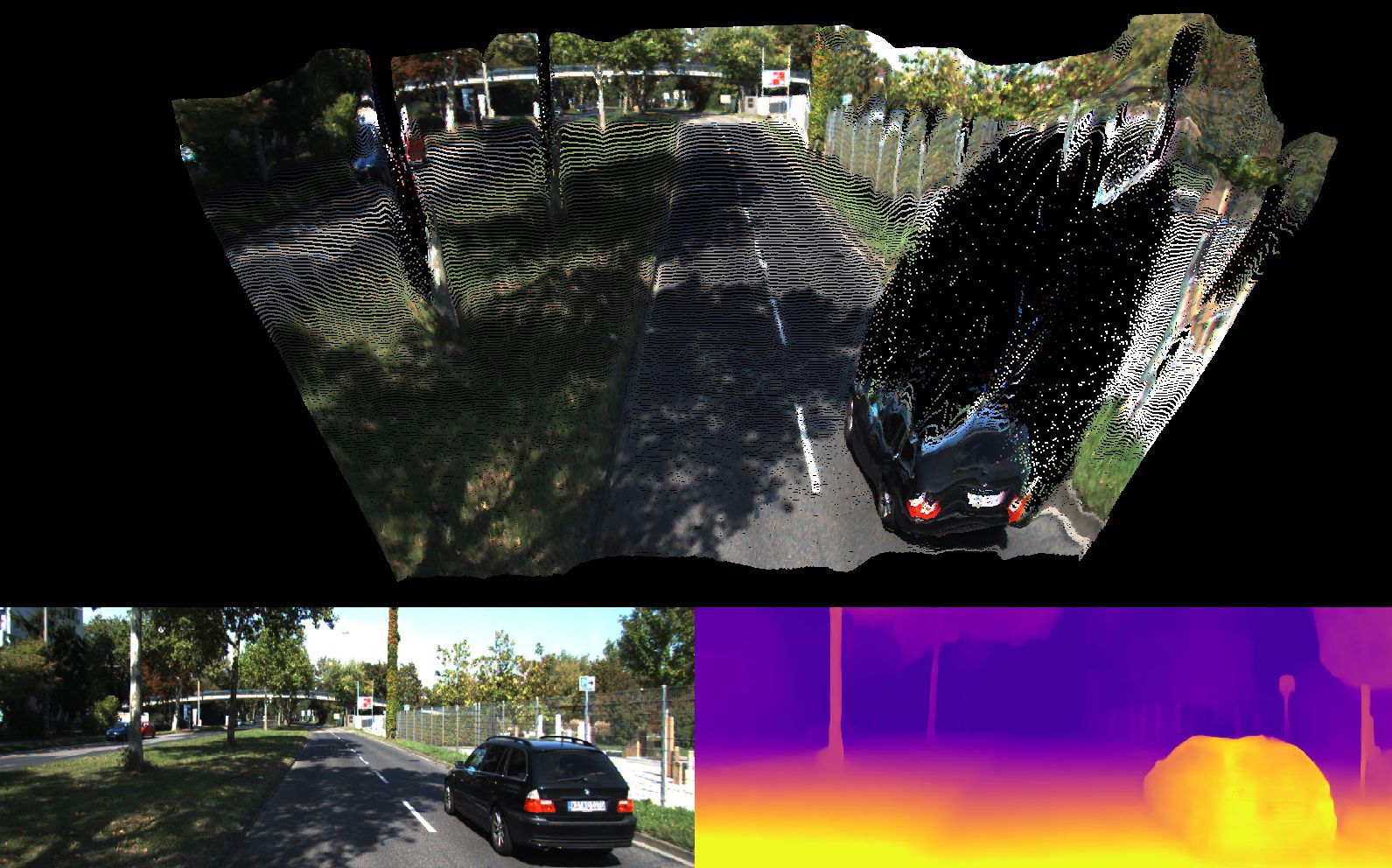}
    }
    \\ \vspace{2mm}
    \subfloat{
        \includegraphics[width=0.49\textwidth,height=6.8cm]{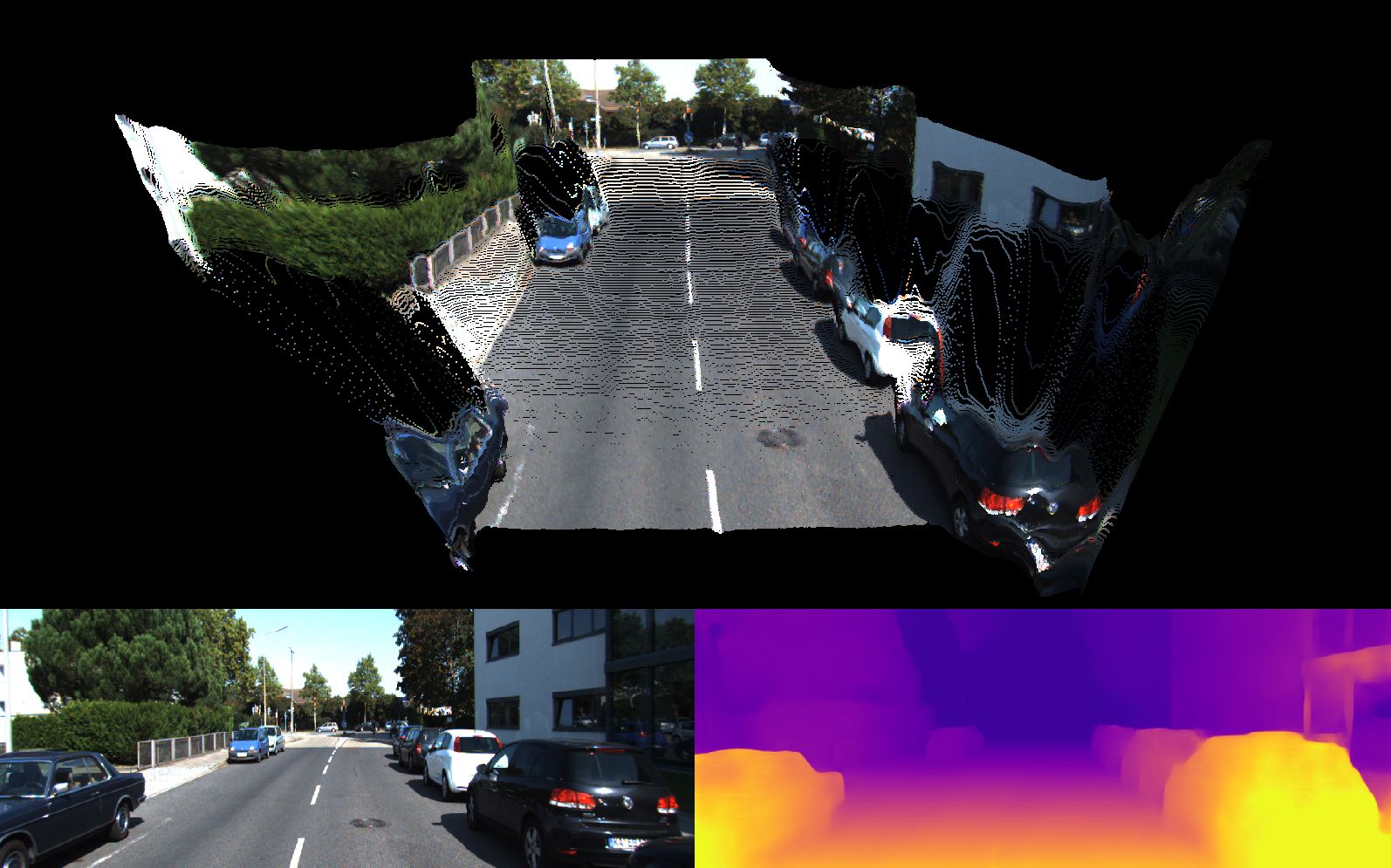}
    }
    \subfloat{
        \includegraphics[width=0.49\textwidth,height=6.8cm]{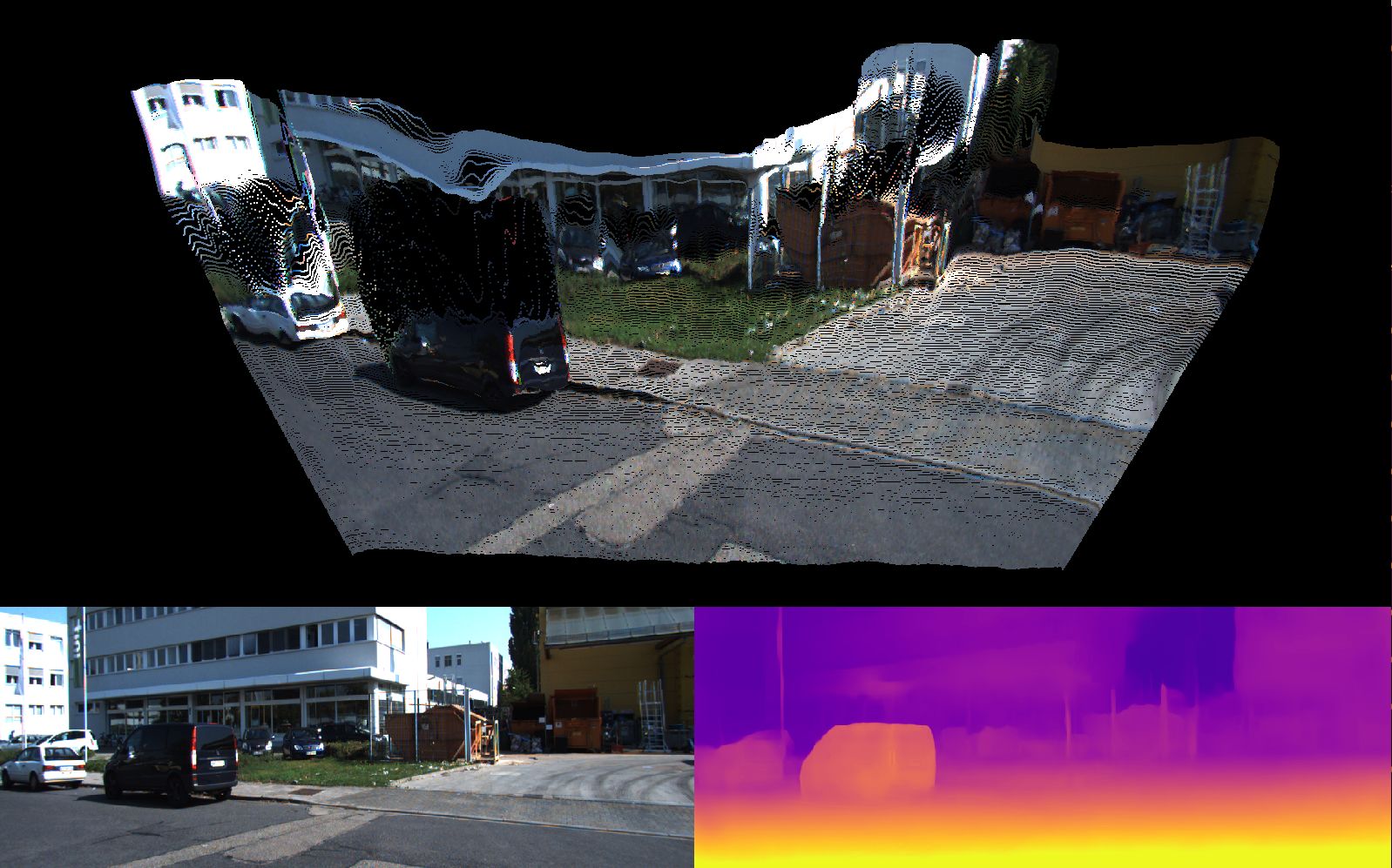}
    }
    \\ \vspace{2mm}
    \subfloat{
        \includegraphics[width=0.49\textwidth,height=6.8cm]{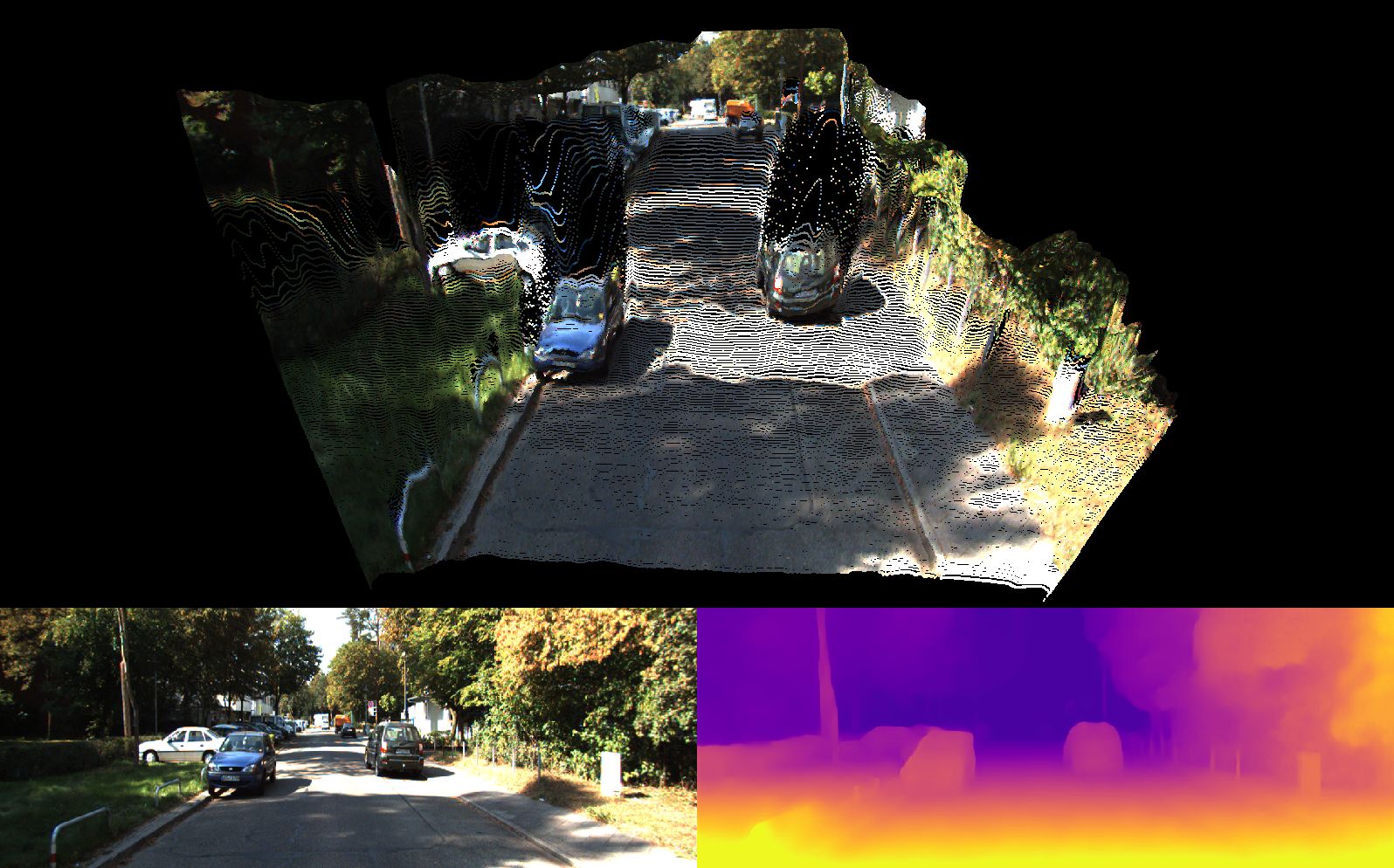}
    }
    \subfloat{
        \includegraphics[width=0.49\textwidth,height=6.8cm]{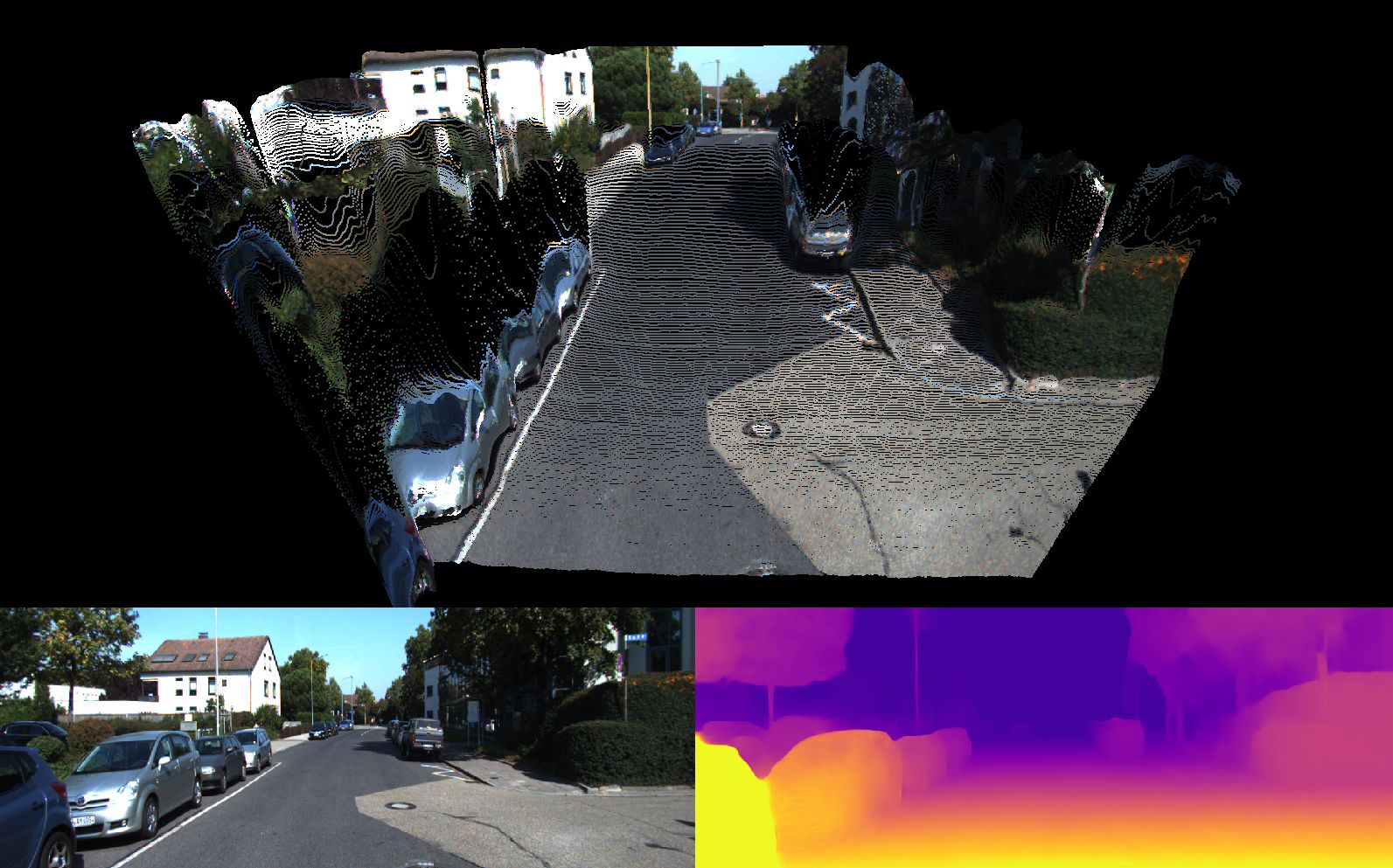}
    }    
    \caption{
        \textbf{Pointcloud reconstructions} obtained using our DepthFormer architecture, on the KITTI dataset. 
    }
    \label{fig:pointclouds}
\end{figure*}


\begin{figure*}[t!]
    \centering
    \subfloat{
        \includegraphics[width=0.45\textwidth,height=6.8cm]{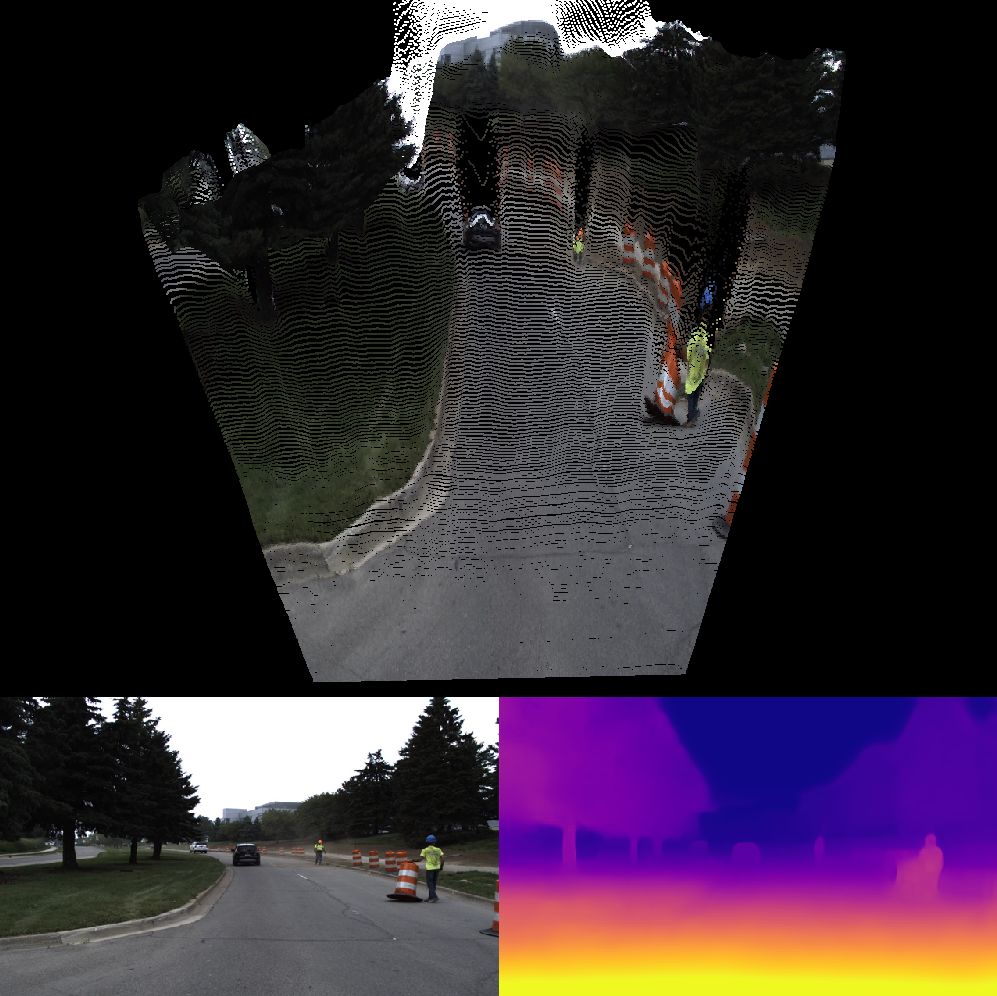}
    }
    \subfloat{
        \includegraphics[width=0.45\textwidth,height=6.8cm]{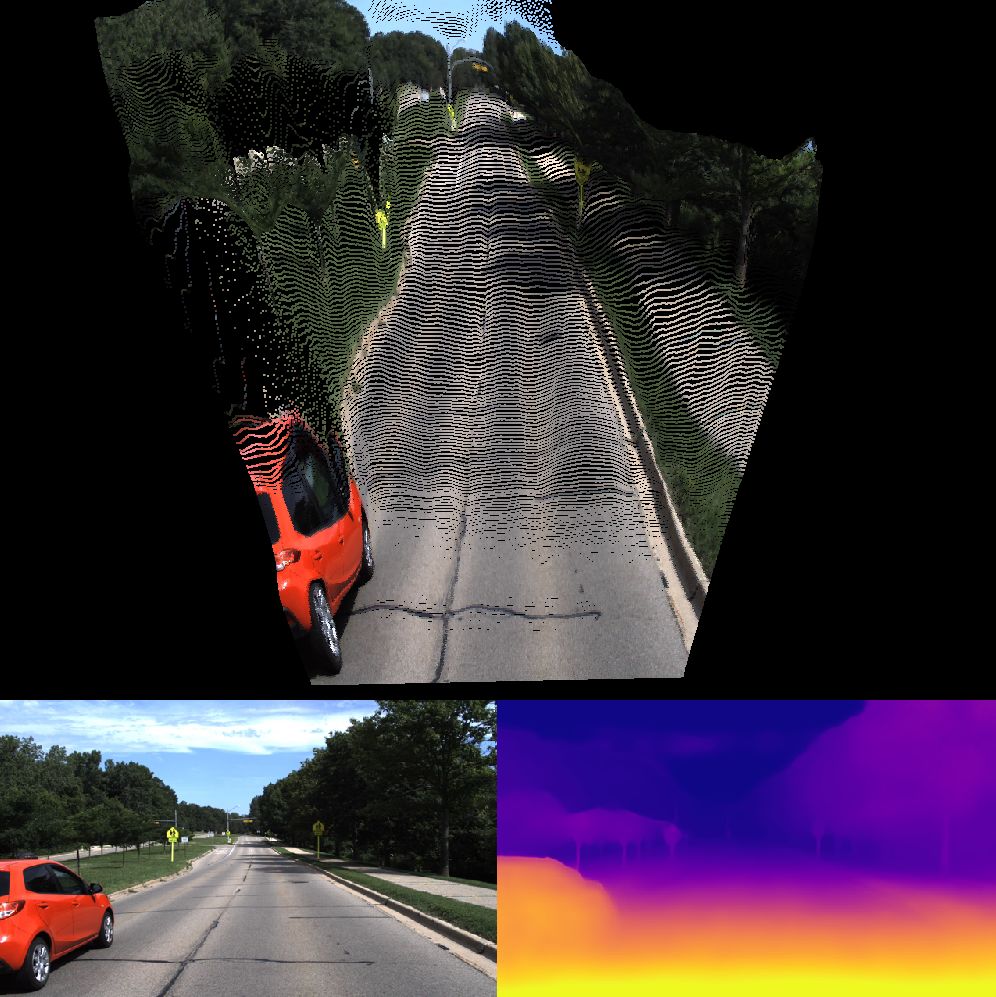}
    }
    \\ \vspace{2mm}
    \subfloat{
        \includegraphics[width=0.45\textwidth,height=6.8cm]{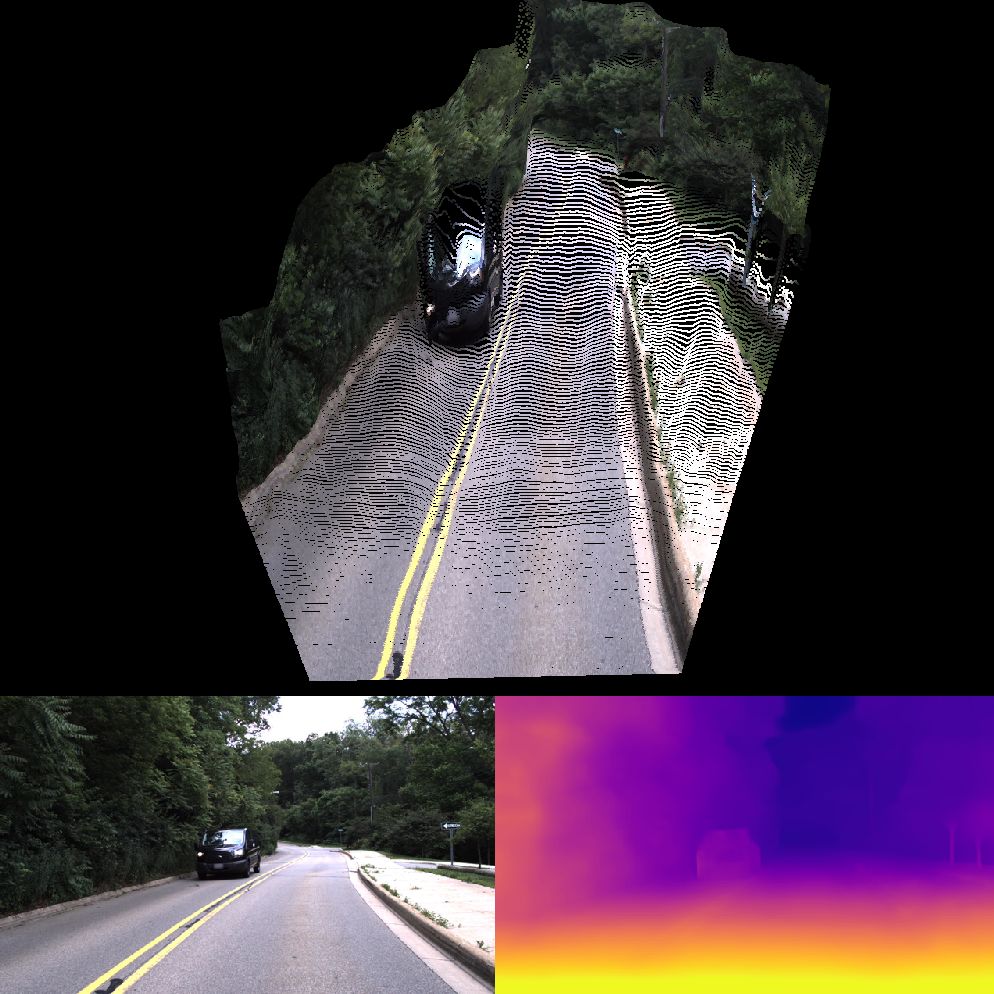}
    }
    \subfloat{
        \includegraphics[width=0.45\textwidth,height=6.8cm]{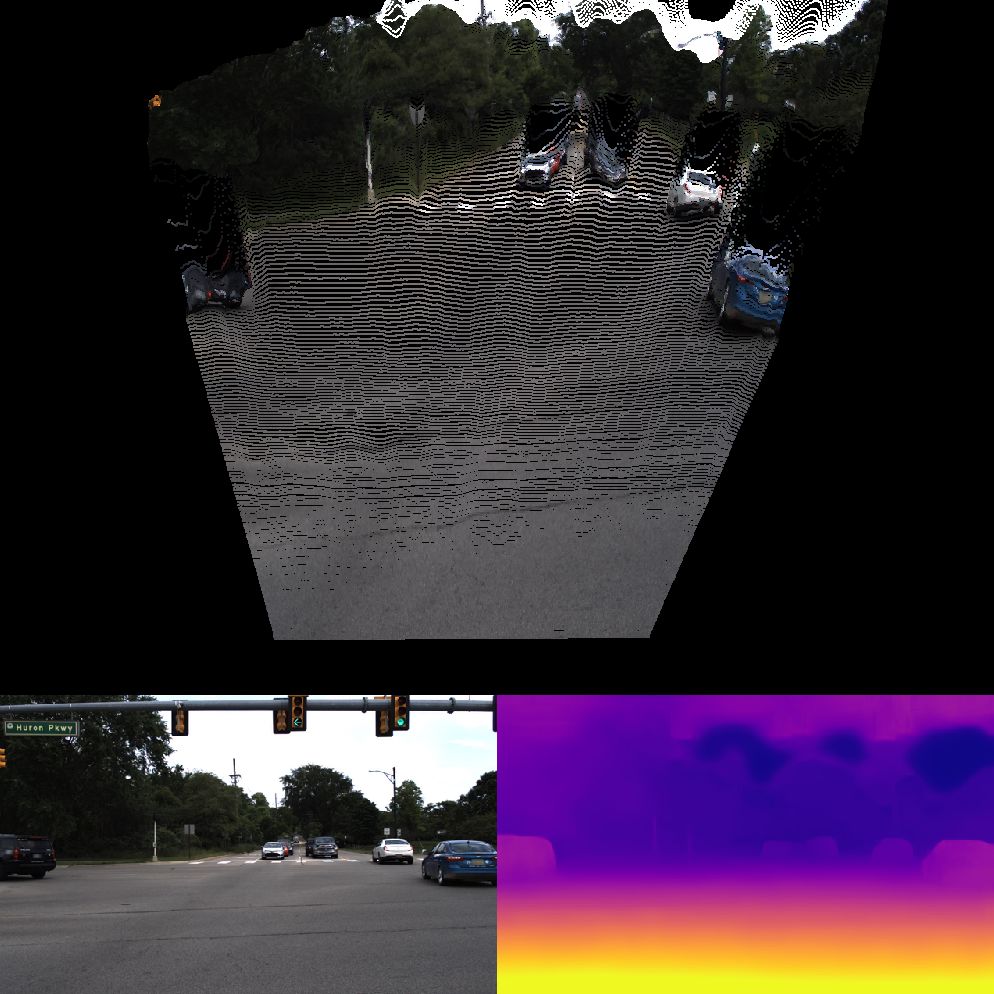}
    }
    \\ \vspace{2mm}
    \subfloat{
        \includegraphics[width=0.45\textwidth,height=6.8cm]{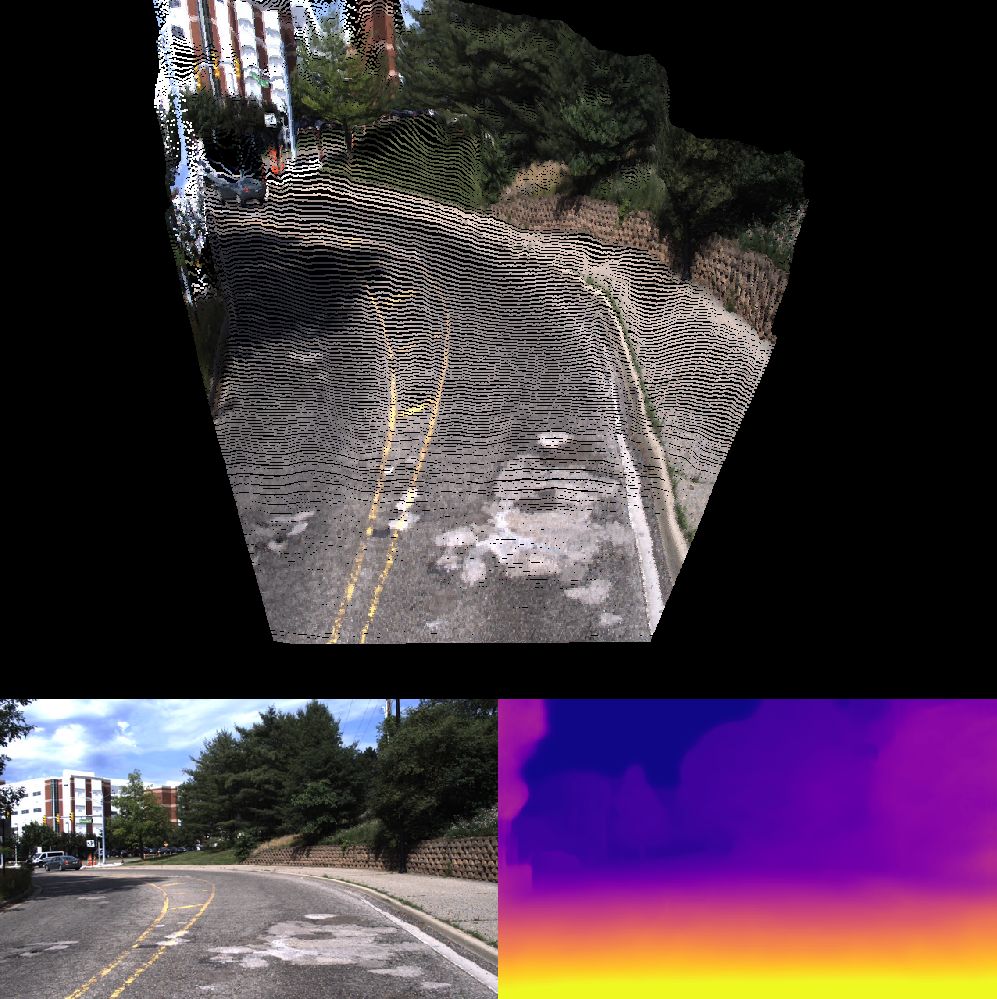}
    }
    \subfloat{
        \includegraphics[width=0.45\textwidth,height=6.8cm]{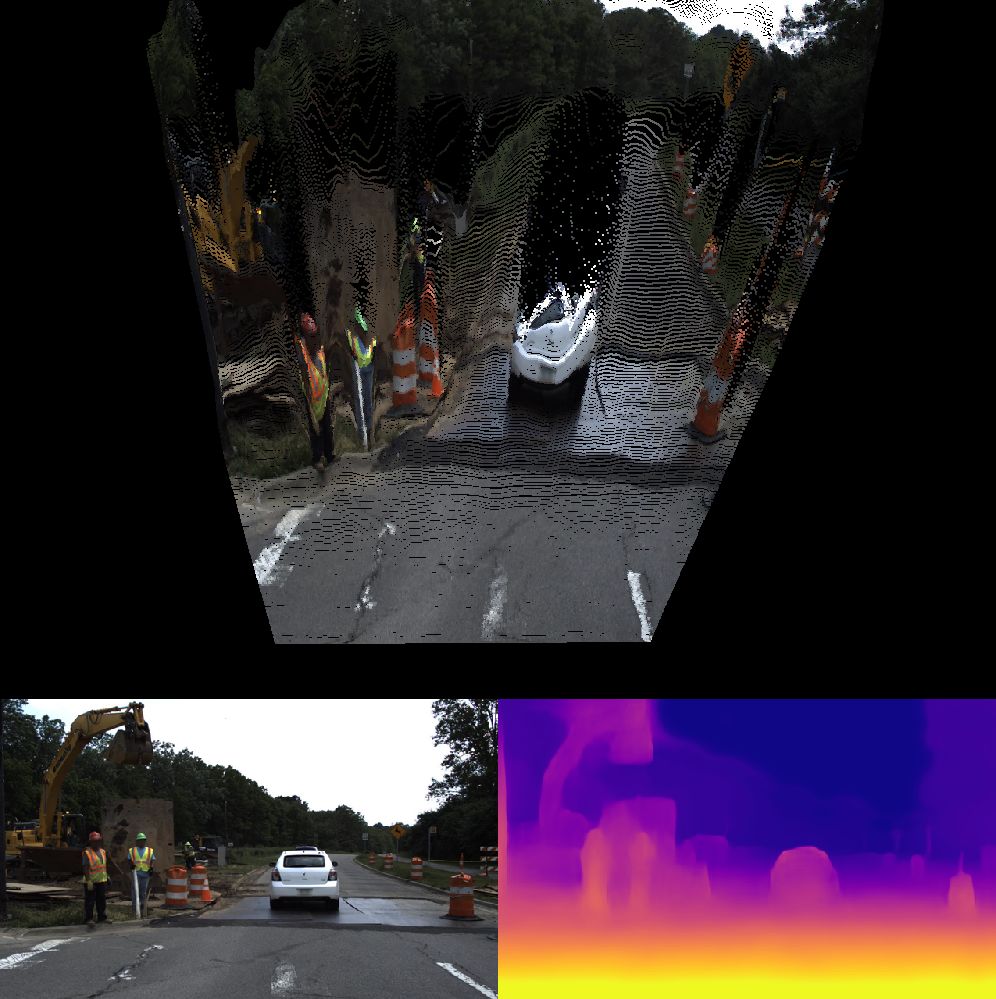}
    }    
    \caption{
        \textbf{Pointcloud reconstructions} obtained using our DepthFormer architecture, on the DDAD dataset. 
    }
    \label{fig:pointclouds_ddad}
\end{figure*}


We also show examples of reconstructed KITTI and DDAD pointclouds in Figures \ref{fig:pointclouds} and \ref{fig:pointclouds_ddad}. These pointclouds are obtained by unprojecting pixel colors to 3D space using known camera intrinsics, predicted depth maps, and predicted relative motion between frames. We reiterate here that no ground-truth is used at training or inference time, only videos. Even so, our architecture is able to reconstruct the observed environment, including low-texture regions, object boundaries, and dynamic objects to a high degree of accuracy, as shown in our quantitative evaluation (Table \ref{table:kitti_depth_sup}). For examples of pointcloud reconstruction over entire sequences, please refer to the supplementary video.





\section{Negative Impact}

Because our proposed method operates on a monocular self-supervised setting, it can process arbitrarily large amounts of unlabeled visual data without human intervention. However, more does not necessarily means better, and some amount of data curation is still desirable, to avoid the introduction of biases in trained models due to data imbalance. Another potential issue is privacy, and proper procedures should be taken when processing large quantities of data without supervision, to preserve individual anonymity. 

\section{Limitations}

Our proposed method increases robustness to some of the common challenges found in self-supervised monocular depth estimation, such as dynamic objects and static frames, by improving feature matching across frames. However, it does not explicitly address these issues, which would require 3D motion modeling in the form of scene flow~\cite{selfsceneflow} or tracking~\cite{zhou2020tracking}. Another common limitation of self-supervised monocular depth estimation is scale ambiguity, since models trained purely on image information cannot produce metrically-accurate predictions. Scale-aware results are necessary for downstream tasks that ingest our reconstructed pointclouds, such as 3D object detection~\cite{wang2021fcos3d}. Some works have addressed this limitation in the self-supervised setting by introducing weak velocity supervision~\cite{packnet} or additional geometric information such as camera height~\cite{wagstaff2021selfsupervised} or multi-camera extrinsics~\cite{guizilini2021surround}. Our proposed method does not address this issue, however it can directly benefit from these works to produce scale-aware estimates.

{\small
\bibliographystyle{ieee_fullname}
\bibliography{references}
}


\end{document}